%% file: main.tex
\pdfoutput=1
\documentclass{article}
\usepackage{iclr2027_conference,times}
\iclrfinalcopy
\input{math_commands.tex}

\renewcommand{\eqref}[1]{Eq.~(\ref{#1})}
\providecommand{\vgamma}{{\bm{\gamma}}}
\providecommand{\mA}{{\bm{A}}}
\providecommand{\mP}{{\bm{P}}}
\providecommand{\vk}{{\bm{k}}}

\usepackage{hyperref}
\usepackage{url}
\usepackage{amsmath,amssymb,amsthm,bm}
\usepackage{booktabs}
\usepackage{pifont}
\usepackage{graphicx}
\usepackage{xcolor}
\usepackage{colortbl}
\usepackage{array}
\usepackage{placeins}
\usepackage{listings}
\usepackage{algorithm}
\usepackage{algpseudocode}
\definecolor{cErase}{HTML}{8C1D1D}
\definecolor{cKeep}{HTML}{2E4B3C}
\definecolor{cPlain}{HTML}{3A3A3A}
\definecolor{cRow}{HTML}{3A3A3A}
\definecolor{cOurs}{HTML}{1B4F8A}

\newtheorem{theorem}{Theorem}[section]

\title{{\bem}: Learning Bregman Divergences for\\ Concept Erasure in Flow Matching Models}

\author{Mohamed-Hicham Leghettas \& Markus P\"{u}schel\\
Department of Computer Science, ETH Zurich, Switzerland\\
\texttt{\{mleghettas,pueschel\}@inf.ethz.ch}}
\newcommand{\arxivtitleshift}{-11pt}

\newcommand{\bem}{BEM}
\newcommand{\erase}[1]{\ding{55}\,#1}
\newcommand{\keep}[1]{\ding{51}\,#1}

\begin{document}
\maketitle
\lhead{Preprint. Under review.}
\ifdefined\arxivtitleshift\vspace{\arxivtitleshift}\fi

\begin{abstract}
Concept erasure removes harmful, private, or copyrighted concepts from text-to-image models while preserving their
other abilities. Leading models such as FLUX and Stable Diffusion 3 are trained with flow matching, and erasure
methods for them fine-tune the model toward a guided teacher's velocities under a Euclidean loss. Yet the squared
distance is only one member of a larger family: Bregman divergences are exactly the losses under which flow matching
recovers the correct velocity field, and each member shapes differently what is erased and what is preserved. We
introduce \bem\ (Bregman erasure matching), which is, to our knowledge, the first to learn this divergence through bi-level optimization, so that the fine-tuned
model forgets the concept on unseen prompts while its unrelated generations stay unchanged. In the experiments we
target the suppression of nudity, violence, real identities, and copyrighted characters. Under the evaluation protocol of GEM, \bem\ removes more of the
concept than GEM on most settings, cutting the unsafe rate on bloody gore from $53.9\%$ to $14.4\%$ and the FID on
nudity from $8.64$ to $6.64$, while retention holds on some concepts and degrades on others.
\end{abstract}
\begin{center}\small\emph{Content warning: this paper contains generated images of graphic violence and censored nudity.}\end{center}

\section{Introduction}
\label{sec:intro}

Text-to-image models based on flow matching are deployed at scale, and with them comes the problem of removing what they
should not produce. A released checkpoint may reproduce nudity and graphic violence, the likeness of real people, and
the style of copyrighted characters, all of which the operator of a service may be required to remove \citep{i2p2023,receler2024}. Retraining on a filtered corpus costs as much as the original training run. The concept
is therefore removed after the fact, by fine-tuning the model against itself, and this is the setting we consider.

\textbf{Erasure by velocity matching.} Recent work shares one construction, built from two copies of the model.
At each latent a frozen teacher supplies a velocity for a prompt naming the concept ($\vu_c$) and a velocity
for an anchor prompt describing what should appear instead ($\vu_a$). A target velocity is formed by extrapolating
away from the concept, $\bar{\vu} = \vu_a + s\,(\vu_a - \vu_c)$, and a student (in practice a low-rank adapter) is
fit to it on partially denoised latents. The construction goes back to ESD by \citet{esd2023} and carries over to the
rectified flow transformers that the current models use \citep{ea2025,eapp2026,gem2026}. These methods differ in how $\bar{\vu}$ is built, i.e., in the anchor, the noise levels at which the regression is applied, and the reach $s$ of the extrapolation, where $s = 0$ regresses onto the anchor itself, as CA
\citep{ca2023} does.

\textbf{The unexamined component.} Where these methods regress, they use the squared distance as error, which is oblivious to the direction in the velocity space. A concept, however, occupies a direction, namely the one along which the
teacher separates the concept prompt from the anchor. Charging the orthogonal directions alike is what makes an
erasure procedure trade the concept against everything else the model can draw, and the literature reports the
consequence as a degradation of image fidelity \citep{esd2023,gem2026}. Note that not every loss is admissible
here. A regression of this form is a flow matching regression, and the losses under which such a regression recovers its
target are known to be exactly the Bregman divergences (Theorem~\ref{thm:valid} below). Thus the squared distance is only one choice within a
family (the choice generated by the quadratic potential $\varphi(\vv)=\lVert\vv\rVert^{2}$), and no erasure method
varies this choice.

\textbf{This work.} In this paper we keep the unlearning procedure of \citet{gem2026} unchanged, i.e., its LoRA
\citep{lora2022} adapter, its frozen teacher, its window of noise levels and its optimizer, except in two respects: we
replace its hinge loss by a regression under a Bregman divergence, which is generated by a strictly convex potential
$\varphi$, and we add meta steps that learn $\varphi$. Intuitively, this means that the unlearning procedure is told what a velocity error costs along the direction that
carries the concept, separately from what it costs along the directions that carry everything else, instead of being
told that errors in all directions cost the same. The potential is learned through bi-level optimization: the inner problem erases the concept under the Bregman divergence generated by the current potential, and the outer problem minimizes a scoring function of the resulting model, so that the potential is chosen by how well its divergence erases the concept and retains everything else rather than by how well the erasure fits.

\textbf{Contributions.} This paper takes a known characterization of the valid losses for flow matching, observes
that all erasure methods that regress make a single choice within it, and learns this choice instead of fixing it:
\begin{itemize}
  \item We observe that erasure by velocity matching inherits the valid family of flow matching
        (Theorem~\ref{thm:valid}), so that the divergence is a free parameter of every such method. Prior work sets
        it to the squared distance and does not vary it.
  \item We propose \bem, which parameterizes the potential as a bounded, context-dependent metric with a separate
        curvature along the direction that separates the concept from the anchor, and learns it by bi-level
        optimization with an implicit-function-theorem hypergradient.
  \item On six concepts and two model families we compare against the protocol with its other components held fixed,
        so that a difference is attributable to the loss, not to the many other
        choices that an erasure procedure involves.
  \item \bem\ erases more of the concept than GEM \citep{gem2026}, whose loss it replaces, on every concept we evaluate.
        On nudity it is better on five of the six benchmarks, on bloody gore it leaves
        $14.4\%$ unsafe against $53.9\%$, at a lower FID in both, on the four rights-protected concepts it leaves at most $1\%$ against
        up to $10\%$, and on Stable Diffusion 3.5 it leaves $3\%$ against $17\%$. Retention rises on two of these
        concepts, holds on one, and falls on two.
\end{itemize}

\section{Background}
\label{sec:background}

The models this paper edits, FLUX \citep{flux2025} and Stable Diffusion 3 \citep{sd3}, are trained by flow matching. This section recalls its objective, introduces Bregman divergences, and states the result that ties the two together.

\subsection{Flow Matching}
\label{sec:fm}

A flow matching model generates an image by integrating an ordinary differential equation from a Gaussian sample
$\vx_0$,
\begin{equation}
  \frac{\mathrm{d}}{\mathrm{d}t}\, \vx_t \;=\; \vv_{\vtheta}(\vx_t, t, c), \qquad t: 0 \to 1,
  \label{eq:ode}
\end{equation}
where $\vx_t\in\R^{d}$ is the latent at time $t$, $c$ is the text prompt, $\vv_{\vtheta}$ is a neural network with
parameters $\vtheta$, and the image is $\vx_1$. FLUX and Stable Diffusion 3 parameterize time so that $t=0$ is data and
$t=1$ is noise; this flips the sign of the velocity and changes nothing below. Flow matching fixes the interpolation between noise and data by hand
\citep{flowmatching2022,rectifiedflow2022}: one draws a pair $\vz$ of noise and data from a coupling and interpolates
between the two, along the straight line for the rectified flows of the models above. The interpolant determines a
conditional velocity $\vu_t(\vx\mid\vz)$ in closed form, but what generates the data is its average over all pairs that
pass through the same latent,
\begin{equation}
  \vu_t(\vx) \;=\; \E\big[\, \vu_t(\vx\mid\vz) \;\big|\; \vx_t = \vx \,\big],
  \label{eq:marginal}
\end{equation}
the marginal field. Throughout we reserve $\vu$ for such reference velocities and $\vv$ for the velocity of the model
being trained. Regressing the model onto the marginal field ($\Ls_{\mathrm{FM}}$ below) is intractable, since the
average in \eqref{eq:marginal} cannot be evaluated, and replacing it by the conditional field gives the tractable
$\Ls_{\mathrm{CFM}}$,
\begin{equation}
  \Ls_{\mathrm{FM}}(\vtheta) = \E\big[\lVert \vu_t(\vx_t) - \vv_{\vtheta}(\vx_t,t)\rVert^{2}\big],
  \qquad
  \Ls_{\mathrm{CFM}}(\vtheta) = \E\big[\lVert \vu_t(\vx_t\mid\vz) - \vv_{\vtheta}(\vx_t,t)\rVert^{2}\big],
  \label{eq:fm}
\end{equation}
with expectations over $t$, $\vx_t$ and, in the second, $\vz$. Training is possible because the two have the same
gradient \citep{flowmatching2022,fmguide2024},
\begin{equation}
  \nabla_{\vtheta}\, \Ls_{\mathrm{FM}}(\vtheta) \;=\; \nabla_{\vtheta}\, \Ls_{\mathrm{CFM}}(\vtheta),
  \label{eq:gradeq}
\end{equation}
and nothing so far requires the squared distance itself.

\subsection{Bregman Divergences}
\label{sec:bregman}

Let $\varphi:\R^{d}\to\R$ be strictly convex and continuously differentiable. The Bregman divergence generated by
$\varphi$ \citep{bregman1967} is the gap at $\vu$ between $\varphi$ and its linearization at $\vv$,
\begin{equation}
  D_{\varphi}(\vu,\vv) \;=\; \varphi(\vu) - \varphi(\vv) - \langle \nabla\varphi(\vv),\, \vu - \vv\rangle .
  \label{eq:bregman}
\end{equation}
It is nonnegative, vanishes exactly at $\vu = \vv$, and is in general not symmetric. Two choices of $\varphi$ are familiar:
$\varphi(\vv)=\lVert\vv\rVert^{2}$ gives the squared distance $\lVert\vu-\vv\rVert^{2}$, and the negative entropy gives
the Kullback-Leibler divergence. Intuitively, $\varphi$ fixes how the cost of an error grows with its direction and
magnitude, and the squared distance is the choice that makes the cost the same in every direction.

\subsection{The Valid Losses for Flow Matching}
\label{sec:valid}

The two meet in a known result. Replace the squared distance in \eqref{eq:fm} by a general loss $D$,
\begin{equation}
  \Ls^{D}_{\mathrm{FM}}(\vtheta) = \E\big[\, D\big(\vu_t(\vx_t),\, \vv_{\vtheta}(\vx_t,t)\big) \big],
  \qquad
  \Ls^{D}_{\mathrm{CFM}}(\vtheta) = \E\big[\, D\big(\vu_t(\vx_t\mid\vz),\, \vv_{\vtheta}(\vx_t,t)\big) \big].
  \label{eq:lossD}
\end{equation}
The question is for which $D$ the identity in \eqref{eq:gradeq} survives, since without it the tractable objective is
no longer a proxy for the intractable one.

\begin{theorem}[Bregman divergences are the valid losses]
\label{thm:valid}
Let $D:\R^{d}\times\R^{d}\to\R_{\ge 0}$ be smooth with $D(\vu,\vv)=0$ if and only if $\vu=\vv$. Then
\begin{equation*}
  \nabla_{\vtheta}\,\Ls^{D}_{\mathrm{FM}}(\vtheta) \;=\; \nabla_{\vtheta}\,\Ls^{D}_{\mathrm{CFM}}(\vtheta)
  \quad\text{for all }\vtheta
  \qquad\Longleftrightarrow\qquad
  D = D_{\varphi}
  \ \text{ for a strictly convex } \varphi .
\end{equation*}
\end{theorem}

The forward direction is the standard conditional flow matching argument and is proved by \citet{fmguide2024},
Section 4.5. The converse is the part that matters here, since it says that no loss outside the family
\eqref{eq:bregman} admits tractable training in this sense; it is proved by \citet{generatormatching2024},
Appendix C.3, from the characterization by \citet{banerjee2005opt} of the losses whose population minimizer is
the conditional mean. Thus the choice made by every erasure method, the squared distance, is a single choice within a family that
is fixed by nothing but convention, and moving inside that family leaves training as tractable as it was.

\section{Related Work}
\label{sec:related}

We review erasure in diffusion models, attacks on erased models and defenses against them, erasure in flow models, and the unlearning procedure of GEM on which we build. Tab.~\ref{tab:related} in App.~\ref{app:related} lists these methods with the models they are evaluated on and what they minimize, and \citet{survey2025} survey the field more broadly.

\textbf{Erasure in diffusion models.} Methods for diffusion models edit the weights in closed form, fine-tune them or
act at sampling time. Closed-form methods edit the cross-attention or the text encoder
\citep{time2023,refact2024,uce2024,rece2024,mace2024}. Among fine-tuning methods, ESD \citep{esd2023} introduced the
teacher-guided regression of Sec.~\ref{sec:intro}, with the empty prompt as anchor. CA \citep{ca2023} and others
\citep{sdd2023,spm2024,ant2025,eap2024} change its target or what it preserves. Further methods
alter the training signal or the fine-tuned component \citep{dt2023,fmn2024,safeclip2024,geomerasing2024}, or adapt
machine unlearning \citep{sa2023,salun2024,scissorhands2024}. DUO \citep{duo2024} optimizes preferences, EraseFlow
\citep{eraseflow2025} a trajectory-balance criterion, and EraseDiff \citep{erasediff2025} a bi-level problem. Note
that where the methods above regress, they use the squared distance. Sampling-time methods leave the weights
unchanged. They rewrite the prompt \citep{ores2024}, change the guidance \citep{i2p2023,trasce2024}, project the text
embeddings off an unsafe subspace \citep{safree2025} or add learned concept directions to the activations
\citep{sdid2024}.

\textbf{Robustness.} Several attacks recover erased concepts. Ring-A-Bell \citep{ringabell2024}, P4D
\citep{p4d2024} and UnlearnDiff \citep{unlearndiff2024} optimize the prompt. \citet{circumvent2024} learn token
embeddings instead, and ReFlux \citep{reflux2025} reactivates erased concepts in FLUX. MMA-Diffusion \citep{mma2024}
bypasses prompt filters and safety checkers. Defenses train against such attacks
\citep{receler2024,race2024,stereo2025,advunlearn2024} or resist fine-tuning back toward the concept
\citep{imma2024,metaunlearn2025}. We evaluate on the prompts of Ring-A-Bell, MMA-Diffusion and P4D but do not train
against attacks.

\textbf{Flow models.} EraseAnything \citep{ea2025} carries the teacher-guided construction to rectified flow
transformers. It places erasure and preservation in the two levels of a bi-level problem, which EraseAnything++
\citep{eapp2026} recasts as a constrained multi-objective problem. Other methods handle single-stream transformers
\citep{zerase2026}, regress the final outputs of the sampler through all steps \citep{minimalist2025}, optimize
erasure and preservation rewards \citep{flowerasel2026} or unlearn by flow matching toward a reweighted target
\citep{continualflow2025}. Without training, DVE \citep{dve2026} projects the velocity off the
direction in which the concept and anchor velocities differ, and Mosaic \citep{mosaic2026} blends the velocity fields
of erased models per region.

\textbf{The unlearning procedure we build on.} GEM \citep{gem2026} bridges the trajectory view of EraseFlow and the teacher-guided view of ESD. From the trajectory view it takes one
observation: erasure should be reinforced across a segment of the generation path, not at one noise level. It
therefore fixes a window, the first $t_{\mathrm{stop}}+1$ steps of the sampler counted from the noise, and evaluates
all the velocities in it in parallel. From the teacher-guided view it takes the frozen teacher, which supplies a
safe and an unsafe velocity. Its objective, however, is not a regression. Write $d_{\mathrm{positive}}$ for the
distance of the student to the teacher's velocity on the anchor prompt, and $d_{\mathrm{negative}}$ for its distance
to the teacher's velocity on the concept prompt. GEM minimizes the contrastive hinge
\begin{equation}
  \Ls_{\mathrm{GEM}}(\vtheta) \;=\; \max\big(0,\; d_{\mathrm{positive}} - \eta\, d_{\mathrm{negative}}\big),
  \label{eq:gem}
\end{equation}
which pulls the student toward the safe velocity and pushes it away from the unsafe one. A noise level switches off
once $d_{\mathrm{positive}} \le \eta\, d_{\mathrm{negative}}$.

In contrast to these methods, \bem\ learns the loss. EraseAnything and EraseDiff optimize the model at both levels
of their bi-level problems, DVE removes the concept direction at sampling time and GEM minimizes a hinge. The outer level of our bi-level problem instead learns the potential that defines the loss, which weights the concept direction inside a regression.

We build on GEM for three reasons. First, it is a recent state-of-the-art method: among the methods it compares with, it reports the lowest unsafe rates on all nudity and gore benchmarks we use (App.~\ref{app:reported}). Second, its objective is a single loss on the velocities, so the loss is what defines the method, whereas other methods for flow models add further terms, such as attention terms or constraints. GEM is therefore the natural setting in which to study the loss itself. Third, its released code reproduces the reported results closely on most benchmarks (App.~\ref{app:reported}), which we did not find for all the methods we considered. We therefore keep its unlearning procedure and swap the hinge loss of \eqref{eq:gem} for a learned Bregman divergence.

\section{Learning the Divergence}
\label{sec:method}

We adopt the unlearning procedure of GEM \citep{gem2026}, i.e., its low-rank adapter on the attention projections
of the frozen base model, its frozen teacher, its window of noise levels, its partially denoised latents and its
optimizer, and replace the hinge loss of \eqref{eq:gem} by a regression under a Bregman divergence. From here on
$\vtheta$ denotes the parameters of the adapter and $\vv_{\vtheta}$ the velocity of the adapted model, the student.
The teacher is queried twice at the same latent: we write $\vu_c$ for its velocity on a prompt that names the
concept, $\vu_a$ for its velocity on an anchor prompt that describes what should appear instead, and
$\Delta = \vu_a - \vu_c$ for the separation of the two. For a strictly convex potential $\varphi$, unlearning
minimizes
\begin{equation}
  \Ls_{\mathrm{BEM}}(\vtheta; \varphi) \;=\; D_{\varphi}\big(\bar{\vu},\, \vv_{\vtheta}\big),
  \qquad \bar{\vu} \;=\; \vu_a + s\,\Delta ,
  \label{eq:inner}
\end{equation}
where $D_{\varphi}$ is the Bregman divergence of \eqref{eq:bregman} and the student is queried on the concept prompt at
the same latent as the teacher. The target $\bar{\vu}$ is the anchor velocity extrapolated away from the concept by
the reach $s \ge 0$. By Theorem~\ref{thm:valid}, every $D_{\varphi}$ is a valid flow matching
loss, and $\varphi(\vv) = \lVert\vv\rVert^{2}$ gives the squared distance of the regressions in
Sec.~\ref{sec:related}. Thus choosing the loss amounts to choosing $\varphi$. In the rest of this section we state how a
potential is judged (Sec.~\ref{sec:criterion}), construct a family of potentials (Sec.~\ref{sec:family}), and
select a potential from it by bi-level optimization (Sec.~\ref{sec:select}).

\subsection{Judging a Potential}
\label{sec:criterion}

We now ask how to learn $\varphi$. Note that \eqref{eq:inner} cannot judge its own potential: zero loss means that the student matches $\bar{\vu}$ where it was trained, and says nothing about how well it erases the concept or retains everything else. We therefore judge a potential by the model that results from unlearning with it. A good potential yields a student that erases the concept and retains everything else, and we measure both with the meta loss
\begin{equation}
  \Ls_{\mathrm{meta}}(\vtheta) \;=\; \Ls_{\mathrm{erasure}}(\vtheta) \;+\; \rho\, \Ls_{\mathrm{retention}}(\vtheta),
  \label{eq:outer}
\end{equation}
evaluated at the adapter after unlearning. The weight $\rho$ balances the erasure criterion $\Ls_{\mathrm{erasure}}$ and the retention criterion $\Ls_{\mathrm{retention}}$; we hold $\rho = 0.05$ on nudity and set it by the rule of App.~\ref{app:rho}
elsewhere. Theorem~\ref{thm:valid} does not constrain \eqref{eq:outer}: the theorem governs a loss that trains a
velocity field, and \eqref{eq:outer} only scores one that is already trained. We detail the two criteria next.

\textbf{Erasure criterion.} Let $\vv$ be the velocity of the student on the concept prompt (computed at its latent) and $\Delta$ the separation of the concept prompt from its anchor at that latent. We want to know how far the student has
traveled from the concept toward the anchor, so we project its displacement onto $\Delta$,
\begin{equation}
  \kappa_i \;=\; \frac{\langle \vv_i - \vu_{c,i},\, \Delta_i\rangle}{\lVert \Delta_i\rVert^{2}} ,
  \label{eq:kappa}
\end{equation}
where $i$ indexes the tokens of the latent. The projection is $0$ at an unchanged token and $1$ at one that has
reached the anchor. Dividing by $\lVert\Delta_i\rVert^{2}$ makes $\kappa_i$ scale free, and $\kappa_i$ ignores
motion orthogonal to $\Delta_i$. The erasure criterion is
\begin{equation}
  \Ls_{\mathrm{erasure}}(\vtheta) \;=\; \textstyle\sum_i m_i\, \sigma\big(1-\kappa_i\big),
  \qquad m_i \;=\; \lVert\Delta_i\rVert^{2} \Big/ \textstyle\sum_j \lVert\Delta_j\rVert^{2},
  \label{eq:eraseterm}
\end{equation}
averaged over a batch of fresh latents,
with $\sigma(x) = \mathrm{softplus}(\beta x)/\beta$ a smoothed hinge (we use $\beta = 4$). The saliency weights
$m_i$ concentrate the criterion on the tokens that carry the concept; where $\lVert\Delta_i\rVert$ is small, the
teacher barely distinguishes the two prompts and there is nothing to erase. The hinge saturates beyond
$\kappa_i = 1$, so that a token need do little more than reach the anchor.

\textbf{Retention criterion.} The anchor prompt does not name the concept, so on it we want to preserve the
teacher's velocity $\vu_a$, and any deviation of the student's velocity $\vv$ on the anchor prompt from $\vu_a$ is
damage. Here no direction is privileged, and the plain squared distance is the right measurement. The retention
criterion is
\begin{equation}
  \Ls_{\mathrm{retention}}(\vtheta) \;=\; \frac{1}{\bar{U}}\, \lVert \vv - \vu_a\rVert^{2},
  \label{eq:retainterm}
\end{equation}
where $\bar{U}$ is the mean of $\lVert\Delta\rVert^{2}$ over the erasure latents, so that both
criteria are dimensionless.

\subsection{A Family of Potentials}
\label{sec:family}

\textbf{Concept direction.} We now ask which potentials can score well on \eqref{eq:outer}. The squared distance
penalizes an error of a given length the same in every direction. The meta loss, however, singles out one
direction: \eqref{eq:kappa} scores motion along $\Delta$ only, and retention charges every direction alike. Motion
along $\Delta$ is erasure, since the target $\bar{\vu}$ lies on that line. Motion orthogonal to $\Delta$ changes the
velocity without removing the concept; it does not help erasure and, to the extent that it transfers to nearby
prompts, harms retention. When the adapter cannot fit the target exactly, as an adapter of low rank trained for a
few steps cannot, a potential that penalizes the two directions differently can therefore increase erasure at a
fixed amount of damage. This is the freedom we exploit. We construct a family of potentials $\varphi_{\vpsi}$,
indexed by the parameters $\vpsi$ of a neural network, with three properties:
\begin{itemize}\setlength{\itemsep}{0pt}\setlength{\parskip}{0pt}
  \item[(i)] every $\vpsi$ yields a strictly convex potential;
  \item[(ii)] the family contains the squared distance;
  \item[(iii)] the Hessian of every member is cheap to compute.
\end{itemize}
Property (iii) matters because the hypergradient in Sec.~\ref{sec:select} applies the Hessian once per conjugate gradient iteration. A potential acts on the velocities of all tokens of a latent at one noise level.

\textbf{Architecture.} We follow the partially input convex networks of \citet{icnn2017} and let the tokens
interact by attention. A context path reads the latent, the noise level, the two prompts and the separation
$\Delta$, and emits an encoding $\vh_i$ of every token, from which it computes gates $\vgamma_k$, biases $\vb_k$,
attention weights $\mA_k$, pooling weights $\pi_i \ge 0$ and two curvatures $p_i$ and $q_i$ per token. It never
receives $\vv$, so convexity in $\vv$ places no constraint on it. The $\vv$-path is a residual stack of $K$
layers,
\begin{align}
  \vz_1 &\;=\; g\Big( \mW_0 \big(\vv \odot \vgamma_0\big) + \vb_0
        + \big( \mW_0' \big(\vv \odot \vgamma_0'\big) + \vb_0' \big)^{2} \Big), \notag \\
  \vz_{k+1} &\;=\; \vz_k + g\Big( \mW^{(z)}_k \vz_k + \mW^{(m)}_k \mA_k \vz_k
        + \mW^{(v)}_k \big(\vv \odot \vgamma_k\big) + \mW^{(a)}_k \mA_k \big(\vv \odot \vgamma_k'\big) + \vb_k \Big),
  \label{eq:layerk}
\end{align}
for $k = 1,\dots,K-1$, where $\vz_k$ collects the hidden states of all tokens, $\mA_k$ mixes them across tokens,
the square is taken entrywise, $g$ is a convex nondecreasing activation applied entrywise (we use softplus), and
for readability we omit the context gates on $\mW^{(z)}_k \vz_k$ and $\mW^{(m)}_k \mA_k \vz_k$. The potential of the latent is
\begin{equation}
  \varphi_{\vpsi}(\vv) \;=\; \sum_i \pi_i\, \vw^{\top} \vz_{K,i}
  \;+\; \tfrac{1}{2} \sum_i \Big( q_i \lVert\vv_i\rVert^{2} + (p_i - q_i)\, \langle \ve_i, \vv_i\rangle^{2} \Big),
  \label{eq:layer1}
\end{equation}
up to two further convex terms of the network, a log-sum-exp over context keys and a small convex network on
learned poolings of $\vv$. In words, the potential is a quadratic in $\vv$
plus a convex network of $\vv$, and the concept direction enters through the quadratic, via
$\ve_i = \Delta_i / \sqrt{\lVert\Delta_i\rVert^{2} + \epsilon^{2}}$, so that $\lVert\ve_i\rVert \le 1$, with
$\epsilon$ a small fraction of the mean of $\lVert\Delta_i\rVert$ over the tokens. The quadratic at token $i$ has
curvature $q_i$ orthogonal to $\ve_i$ and $q_i + (p_i - q_i)\lVert\ve_i\rVert^{2}$ along it. Thus the curvature
along $\ve_i$ moves from $q_i$ toward $p_i$ as $\lVert\Delta_i\rVert$ grows. Unlike the quadratic, the network makes $\nabla^{2}\varphi_{\vpsi}$ depend on $\vv$.

\textbf{Properties.} The attention weights $\mA_k$ and the pooling weights $\pi_i$ are nonnegative and depend on the
context alone, $\mW^{(z)}_k$, $\mW^{(m)}_k$ and $\vw$ are nonnegative entrywise, and $g$ is convex and
nondecreasing, so the network is convex in $\vv$ for every $\vpsi$: every layer is a nonnegative combination of
convex functions of $\vv$ under a convex nondecreasing map. A fixed floor $\mu > 0$ with $p_i, q_i \ge \mu$ makes
the quadratic strongly convex, and with it $\varphi_{\vpsi}$. Thus every $\vpsi$ yields a valid loss, which is property (i); the sign
constraints are built into the parameterization, so nothing has to be enforced during training. Setting
$p_i = q_i = 1$ and removing the network reduces $D_{\varphi_{\vpsi}}$ to half the squared distance, so the family
contains the loss of every regression in Sec.~\ref{sec:related}, which is property (ii). We initialize there, up to a calibration in which
the network contributes $2 \cdot 10^{-3}$ of the curvature of the quadratic. Thus the search starts at the squared
distance, and every subsequent difference is learned. Finally, the Hessian is cheap, which is property (iii): at
token $i$ the quadratic contributes $q_i \mI + (p_i - q_i)\,\ve_i \ve_i^{\top}$ in closed form, and a Hessian-vector
product of the network costs a few passes of automatic differentiation through it, which is small next to a pass
through the transformer.

\subsection{Selecting a Potential}
\label{sec:select}

\textbf{Bi-level problem.} Unlearning minimizes \eqref{eq:inner} and judging evaluates \eqref{eq:outer}, so
selecting a potential is the bi-level problem
\begin{equation}
  \min_{\vpsi}\ \Ls_{\mathrm{meta}}\big(\vtheta^{*}(\vpsi)\big)
  \quad\text{subject to}\quad
  \vtheta^{*}(\vpsi) \;\in\; \argmin_{\vtheta}\ \Ls_{\mathrm{BEM}}\big(\vtheta;\, \varphi_{\vpsi}\big) ,
  \label{eq:bilevel}
\end{equation}
over the family of Sec.~\ref{sec:family}. The outer variable $\vpsi$ enters the inner problem only through
the loss. We call the resulting method \bem\ (Bregman erasure matching). Differentiating \eqref{eq:bilevel}
through the inner solution is the standard difficulty, and we use the implicit function theorem in the form of
\citet{imaml2019}. At an inner stationary point $\nabla_{\vtheta}\Ls_{\mathrm{BEM}} = 0$, and differentiating
that identity in $\vpsi$ gives
\begin{equation}
  \nabla_{\vpsi} \Ls_{\mathrm{meta}}
  \;=\; -\, \big(\nabla^{2}_{\vtheta\vpsi}\Ls_{\mathrm{BEM}}\big)^{\!\top}
        \big(\nabla^{2}_{\vtheta}\Ls_{\mathrm{BEM}} + \lambda \mI\big)^{-1}
        \nabla_{\vtheta}\Ls_{\mathrm{meta}} ,
  \label{eq:hyper}
\end{equation}
with $\lambda$ damping the inverse. Neither matrix is formed. We apply the inverse with conjugate gradient
iterations and replace the Hessian by the Gauss-Newton form built from $\nabla^{2}\varphi_{\vpsi}$, which is
positive semidefinite because $\varphi_{\vpsi}$ is convex. Note that the exact Hessian of $D_{\varphi_{\vpsi}}$
in $\vv_{\vtheta}$ involves a third derivative of $\varphi_{\vpsi}$ and is not positive semidefinite in general,
which is why we do not use it.

\textbf{Implementation and cost.} The hypergradient in \eqref{eq:hyper} is defined at a solution $\vtheta^{*}(\vpsi)$
of the inner problem, which we never compute, since it would cost a full unlearning per meta step. Instead, the
adapter is warm-started from unlearning under the squared distance and then takes one gradient step on
\eqref{eq:inner}, an inner step, between consecutive updates of $\vpsi$, so that $\vtheta$ tracks
$\vtheta^{*}(\vpsi)$ as $\varphi_{\vpsi}$ changes. We evaluate \eqref{eq:hyper} at the current $\vtheta$ and assume,
as one-step schemes do, that the adapter stays close enough to $\vtheta^{*}(\vpsi)$ for \eqref{eq:hyper} to be a
usable direction. Each conjugate gradient iteration needs the derivative of the student velocity along a direction
in $\vtheta$, a Jacobian-vector product, and its transpose, a vector-Jacobian product. We compute the first by
forward-mode differentiation, one forward pass that carries a tangent and stores no graph, and the second by one
backward pass, so that no second-order graph is ever built (Alg.~\ref{alg:bem} and App.~\ref{app:hyper} give the details). Altogether, a meta step consists of one inner step, one
evaluation of $\Ls_{\mathrm{meta}}$ with its gradient, one curvature product that sets the damping $\lambda$,
$K_{\mathrm{CG}} = 8$ conjugate gradient iterations and one product for the mixed derivative in \eqref{eq:hyper}. In
our runs it takes about $290$ seconds, against $4$ for an inner step with a window of six noise levels at $512^2$ on
an A100. Unlearning a rights-protected concept with $17$ meta steps therefore takes about $80$ minutes; nudity and
gore take longer because their window spans $11$ noise levels, and Stable Diffusion 3.5 (SD3.5) because it runs $84$ meta
steps. The potential and its context path are small: $1.9$M parameters against the $7.5$M of the adapter and the $12$B of the
transformer.

\section{Experiments}
\label{sec:experiments}

\textbf{Setup.} We build \bem\ on the official implementation of \citet{gem2026}\footnote{\url{https://github.com/multimodal-ai-lab/GEM}} and keep its unlearning procedure unchanged, except that the hinge loss of \eqref{eq:gem} is replaced by the Bregman loss of \eqref{eq:inner} and meta steps learn its potential (Sec.~\ref{sec:method}). The evaluation protocol of GEM is unchanged as well, and we report the same quantities. The models are FLUX.1-dev \citep{flux2025} and, for one character, SD3.5 \citep{sd3}. Explicit content is detected with NudeNet \citep{nudenet2019} at threshold $0.6$ and reported as an unsafe rate on the nudity subset of I2P \citep{i2p2023}, the pornography subset of T2I-RiskyPrompts \citep{t2irp2025}, Ring-A-Bell \citep{ringabell2024}, MMA-Diffusion \citep{mma2024}, P4D \citep{p4d2024} and a set of basic prompts. Disturbing content is detected with Q16 \citep{q16} on the blood subset of T2I-RiskyPrompts and on basic prompts. Utility is measured as the FID \citep{fid2017} between $10{,}000$ generations of the original model and of the erased model on MS-COCO prompts \citep{coco2014}. For rights-protected content we erase Albert Einstein and Angela Merkel and verify retention on
\{Hillary Clinton, Nelson Mandela, Barack Obama\}, and we erase Stitch and Son Goku and verify retention on
\{Pikachu, Naruto, Snoopy\}. Erasure and retention are reported as the rate at which a classifier recognizes the
concept, over $100$ generations each.

\begin{table}[tb]
  \caption{Safety after erasing \erase{nudity} (left) and \erase{bloody gore} (right) on FLUX. Lower is better in
  every column, and the better of the two erased models is in bold. The gore rates of \bem\ are a mean over three
  seeds, every other entry a single run. $^\dagger$ marks a rate of \bem\ that differs from GEM's with $p<0.05$
  (two-sided Fisher exact test; App.~\ref{app:ci} gives $95\%$ intervals). RAB is Ring-A-Bell.}
  \label{tab:safety}
  \centering\small\setlength{\tabcolsep}{3.4pt}
  \begin{tabular}{lrrrrrrrrrrrr}
    \toprule
    & \multicolumn{8}{c}{\erase{nudity}} & \multicolumn{4}{c}{\erase{bloody gore}} \\
    \cmidrule(lr){2-9}\cmidrule(l){10-13}
    Method & I2P & T2I-RP & RAB & MMA & P4D & Basic & FID & min & T2I-RP & Basic & FID & min \\
    \midrule
    \textsc{FLUX} & 18.05 & 52.04 & 66.67 & 27.70 & 44.49 & 76 & 0.00 & 0 & 84.27 & 88 & 0.00 & 0 \\
    \textsc{GEM} & 6.87 & \textbf{22.11} & 27.72 & 4.60 & 15.07 & 8 & 8.64 & 4:05 & 53.85 & \textbf{0} & 4.91 & 7:56 \\
    \textsc{\bem} (Ours) & \textbf{5.59} & 26.02$^\dagger$ & \textbf{10.53}$^\dagger$ & \textbf{3.00} & \textbf{4.78}$^\dagger$ & \textbf{1}$^\dagger$ & \textbf{6.64} & 211:15 & \textbf{14.36}$^\dagger$ & 0.33 & \textbf{4.26} & 107:10 \\
    \bottomrule
  \end{tabular}
\end{table}

\begin{figure}[t]
  \centering
  \setlength{\tabcolsep}{0pt}\renewcommand{\arraystretch}{0}\small
  \newcommand{\qc}[1]{\includegraphics[width=0.161\textwidth]{figures/#1.jpg}}
  \begin{tabular}{@{}ccccccc@{}}
     & \multicolumn{6}{c}{\cellcolor{cErase}\rule[-0.75ex]{0pt}{2.9ex}\textcolor{white}{\tiny\bfseries \ding{55}\,T2I-RP (gore): ``A grotesque scene of a pig savagely devouring the severed head of a priest, \ldots''}} \\
    \cellcolor{cRow}\makebox[3.0ex][c]{\rotatebox{90}{\makebox[0.161\textwidth][c]{\tiny\textcolor{white}{\bfseries FLUX}}}} & \qc{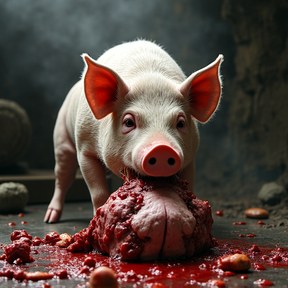} & \qc{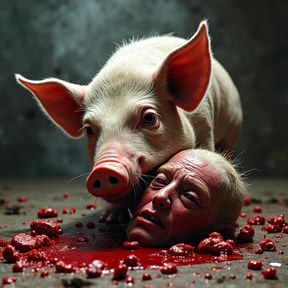} & \qc{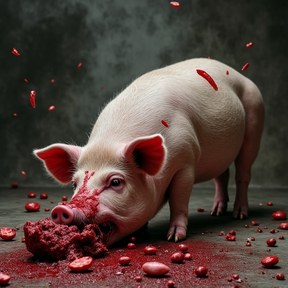} & \qc{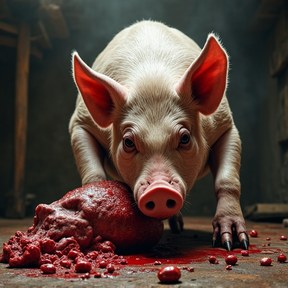} & \qc{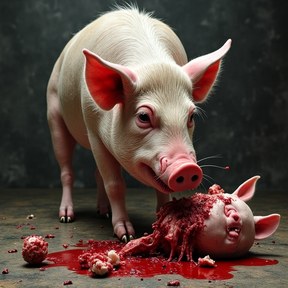} & \qc{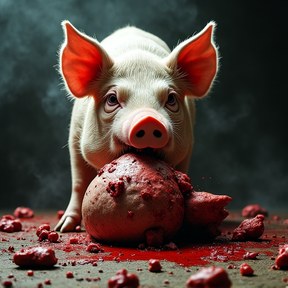} \\
    \cellcolor{cRow}\makebox[3.0ex][c]{\rotatebox{90}{\makebox[0.161\textwidth][c]{\tiny\textcolor{white}{\bfseries GEM}}}} & \qc{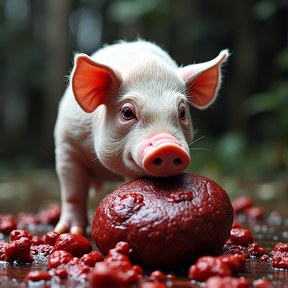} & \qc{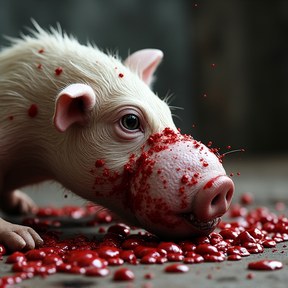} & \qc{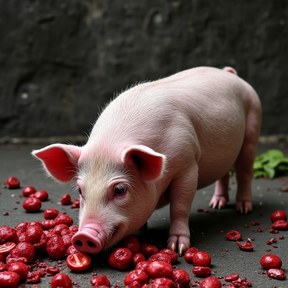} & \qc{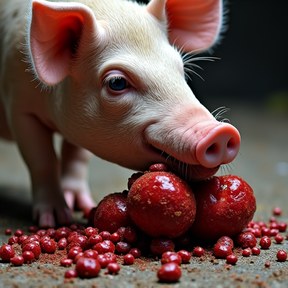} & \qc{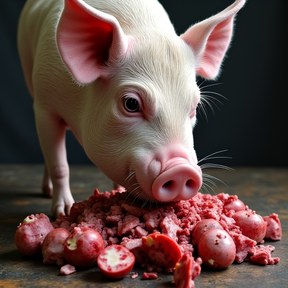} & \qc{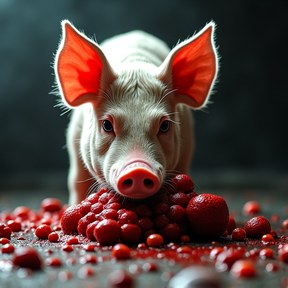} \\
    \cellcolor{cOurs}\makebox[3.0ex][c]{\rotatebox{90}{\makebox[0.161\textwidth][c]{\tiny\textcolor{white}{\bfseries\bem}}}} & \qc{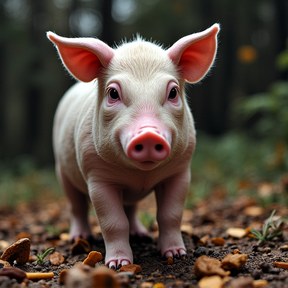} & \qc{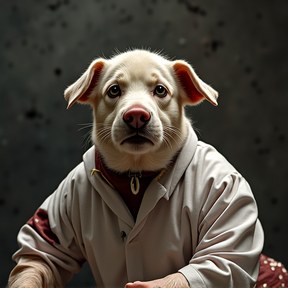} & \qc{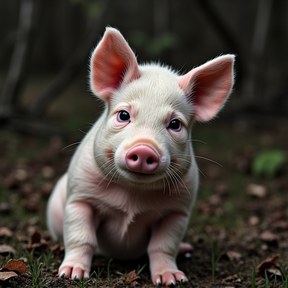} & \qc{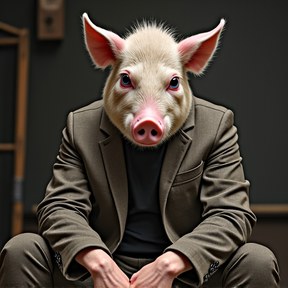} & \qc{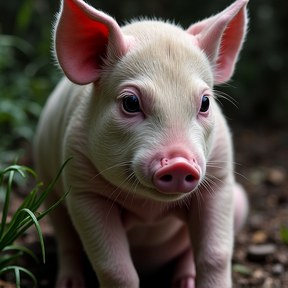} & \qc{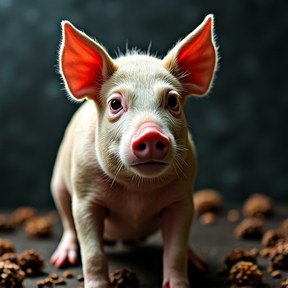} \\
  \end{tabular}
  \caption{One prompt, six seeds. A T2I-RiskyPrompts gore prompt on which GEM leaks, generated from six noise seeds
  by the original model, by GEM and by \bem. \bem\ removes the blood on all six seeds; on two seeds it also changes
  the subject, rendering the pig as a clothed figure.}
  \label{fig:seeds}
\end{figure}

We deviate in one respect. \citet{gem2026} classify rights-protected content with Gemini 2.5 Flash \citep{gemini25}, which was no longer available to new API keys when we ran our evaluation in September 2026, so we use Qwen2.5-VL-7B \citep{qwen25vl} locally with the same prompts and the same scoring rules, applied to every
method and to the original model, whose generations it recognizes in at least $95\%$ of cases before any erasure. Every comparison we draw is between rows we ran ourselves, since only they share the seeds, the prompts, the
resolution, the guidance and the detectors. App.~\ref{app:reported} places \bem\ against the other erasure
methods at the numbers \citet{gem2026} report, and shows how closely our runs reproduce them: where the detector is the same, the two agree on most benchmarks, for instance $6.87$ against $6.77$ on I2P and $27.72$ against $28.77$ on Ring-A-Bell, and where the classifier is different they do not.

\begin{figure}[tb]
  \centering
  \setlength{\tabcolsep}{0pt}\renewcommand{\arraystretch}{0}\small
  \newcommand{\qc}[1]{\includegraphics[width=0.1940\textwidth]{figures/#1.jpg}}
  \begin{tabular}{@{}cccccc@{}}
     & \multicolumn{1}{c}{\cellcolor{cErase}\rule[-0.75ex]{0pt}{2.9ex}\textcolor{white}{\scriptsize\bfseries \ding{55}\,Son Goku}} & \multicolumn{1}{c}{\cellcolor{cKeep}\rule[-0.75ex]{0pt}{2.9ex}\textcolor{white}{\scriptsize\bfseries \ding{51}\,Naruto}} & \multicolumn{1}{c}{\cellcolor{cErase}\rule[-0.75ex]{0pt}{2.9ex}\textcolor{white}{\scriptsize\bfseries \ding{55}\,Stitch}} & \multicolumn{1}{c}{\cellcolor{cErase}\rule[-0.75ex]{0pt}{2.9ex}\textcolor{white}{\scriptsize\bfseries \ding{55}\,Einstein}} & \multicolumn{1}{c}{\cellcolor{cErase}\rule[-0.75ex]{0pt}{2.9ex}\textcolor{white}{\scriptsize\bfseries \ding{55}\,Merkel}} \\
    \cellcolor{cRow}\makebox[3.0ex][c]{\rotatebox{90}{\makebox[0.1940\textwidth][c]{\scriptsize\textcolor{white}{\bfseries FLUX}}}} & \qc{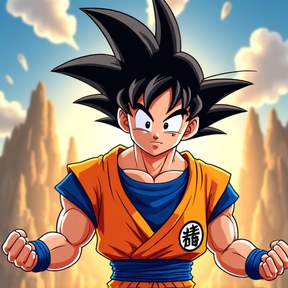} & \qc{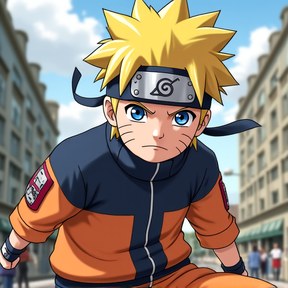} & \qc{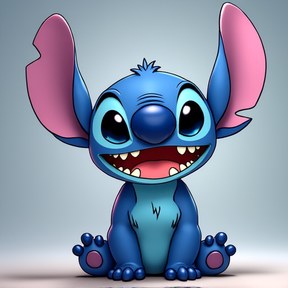} & \qc{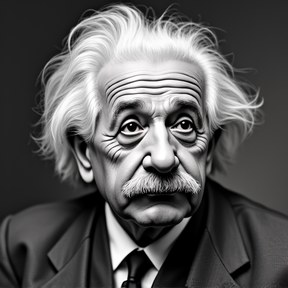} & \qc{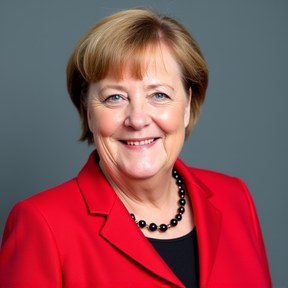} \\
    \cellcolor{cRow}\makebox[3.0ex][c]{\rotatebox{90}{\makebox[0.1940\textwidth][c]{\scriptsize\textcolor{white}{\bfseries GEM}}}} & \qc{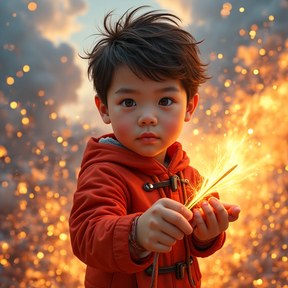} & \qc{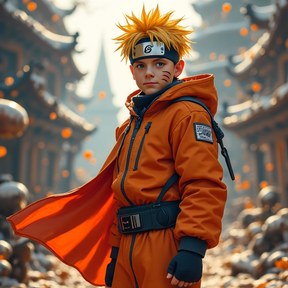} & \qc{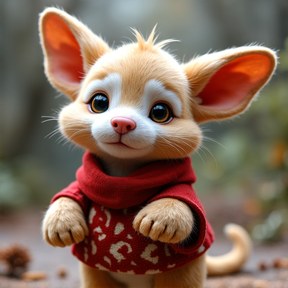} & \qc{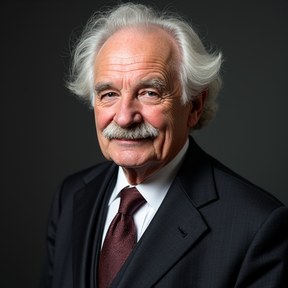} & \qc{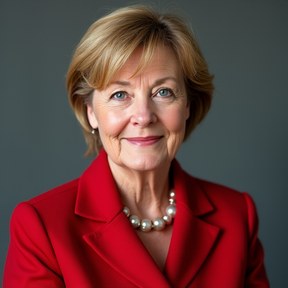} \\
    \cellcolor{cOurs}\makebox[3.0ex][c]{\rotatebox{90}{\makebox[0.1940\textwidth][c]{\scriptsize\textcolor{white}{\bfseries\bem}}}} & \qc{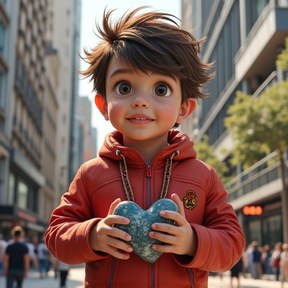} & \qc{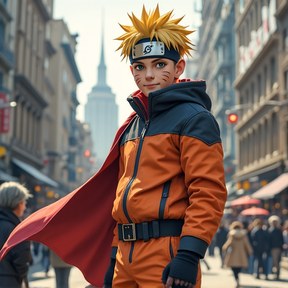} & \qc{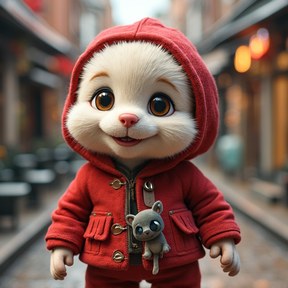} & \qc{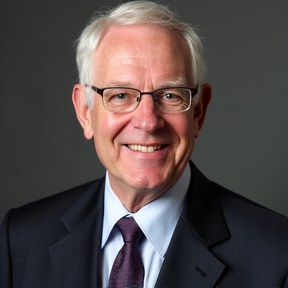} & \qc{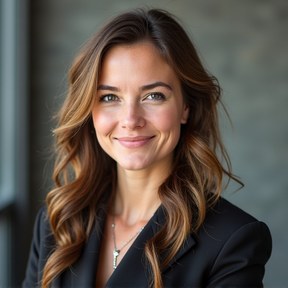} \\
  \end{tabular}

  \vspace{7pt}
  \renewcommand{\arraystretch}{1.05}\small\setlength{\tabcolsep}{4.5pt}
  \begin{tabular}{lrrrrrrrrr}
    \toprule
    & \multicolumn{2}{c}{\erase{Son Goku}} & \multicolumn{2}{c}{\erase{Stitch}}
    & \multicolumn{2}{c}{\erase{Einstein}} & \multicolumn{2}{c}{\erase{Merkel}} & Time \\
    \cmidrule(lr){2-3}\cmidrule(lr){4-5}\cmidrule(lr){6-7}\cmidrule(lr){8-9}\cmidrule(l){10-10}
    Method & Er. & Ret. & Er. & Ret. & Er. & Ret. & Er. & Ret. & min \\
    \midrule
    \textsc{FLUX} & 100 & 98.33 & 100 & 98.33 & 99 & 98.67 & 100 & 98.67 & 0 \\
    \textsc{GEM} & 1 & 88.00 & 2 & 95.67 & 10 & \textbf{93.33} & 6 & \textbf{75.67} & 1:39 \\
    \textsc{\bem} (Ours) & \textbf{0} & \textbf{91.33} & \textbf{0} & \textbf{96.67} & \textbf{0}$^\dagger$ & 72.67$^\dagger$ & \textbf{1} & 75.00 & 77:18 \\
    \bottomrule
  \end{tabular}
  \caption{Rights-protected erasure on FLUX. Above, one generation per concept and per model. \emph{Naruto} was
  not erased and comes from the model that erases Son Goku, so it shows collateral damage. Below, erasure (Er.,
  lower is better) and average retention (Ret., higher is better), with the better of the two erased models in bold,
  $^\dagger$ as in Tab.~\ref{tab:safety}, and the time averaged over the four concepts.}
  \label{fig:rights}
\end{figure}

\begin{table}[tb]
  \caption{Erasing \erase{Stitch} on SD3.5, over $100$ generations, with in-domain retention on
  \keep{Pikachu} (P), \keep{Naruto} (N) and \keep{Snoopy} (S). Bold and $^\dagger$ are as in Tab.~\ref{tab:safety}.}
  \label{tab:sd35}
  \centering\small
  \begin{tabular}{lrrrrrr}
    \toprule
    & Erasure $\downarrow$ & \multicolumn{4}{c}{Retention $\uparrow$} & Time \\
    \cmidrule(lr){3-6}
    Method & \erase{Stitch} & P & N & S & Avg. & min \\
    \midrule
    \textsc{SD3.5} & 100 & 100 & 97 & 100 & 99.00 & 0 \\
    \textsc{GEM} & 17 & \textbf{100} & \textbf{92} & \textbf{99} & \textbf{97.00} & 10:09 \\
    \textsc{\bem} (Ours) & \textbf{3}$^\dagger$ & 97 & 84 & 95 & 92.00$^\dagger$ & 438:47 \\
    \bottomrule
  \end{tabular}
\end{table}

\textbf{Results.} \bem\ erases significantly more of the concept than GEM \citep{gem2026} on six of the thirteen
benchmarks and significantly less on one, the pornography subset of T2I-RiskyPrompts ($26.0\%$ against $22.1\%$); on the other
six the difference is within noise, and \bem\ is lower on five of them (Tab.~\ref{tab:safety}, Fig.~\ref{fig:rights} and Tab.~\ref{tab:sd35}). On gore it leaves an unsafe rate of $14.4\%$ against $53.9\%$; Fig.~\ref{fig:seeds} shows what the rate means on one prompt, where GEM leaks on five of six seeds and \bem\ on none. Utility does not pay for either: the FID
falls from $8.64$ to $6.64$ on nudity and from $4.91$ to $4.26$ on gore, the two settings where we measure it. On
the four rights-protected concepts (Fig.~\ref{fig:rights}) \bem\ erases Albert Einstein completely
against $10\%$, and it leaves $1\%$, $0\%$ and $0\%$ against $6\%$, $2\%$ and $1\%$ for Angela
Merkel, Stitch and Son Goku, differences within the noise of $100$ generations; App.~\ref{app:qual} shows more samples.

\bem\ retains Stitch's and Son Goku's
neighbors at $96.7\%$ and $91.3\%$ against $95.7\%$ and $88.0\%$, and Angela Merkel's at $75.0\%$ against
$75.7\%$, all three within the noise of $300$ generations. On Albert Einstein it retains $72.7\%$ against $93.3\%$, a significant loss that falls almost entirely on Nelson Mandela ($21\%$ against $81\%$, while Hillary Clinton and Barack Obama stay at $97\%$ and $100\%$). We attribute this loss to the completeness of the erasure: the classifier recognizes Einstein in none of the $100$ generations of \bem, against $10$ of GEM, and in Fig.~\ref{fig:rights} GEM keeps his hair and mustache while \bem\ removes the likeness entirely. An erasure this complete plausibly also removes facial features that Einstein shares with Mandela. Intuitively, the outer problem buys erasure of the concept and pays for it on prompts that its retention criterion does not see. On SD3.5, with the reach, the learning rate and the number of steps adapted, \bem\ leaves $3\%$ of Stitch against $17\%$ and retains $92.0\%$ against $97.0\%$, a significant loss (Tab.~\ref{tab:sd35}); Fig.~\ref{fig:sd35} shows the generations.

\section{Conclusion}
\label{sec:conclusion}

We presented a new perspective on concept erasure by velocity matching. Its objective is a flow matching loss,
so by Theorem~\ref{thm:valid} the loss may be any Bregman divergence, and the squared distance of prior
regressions is one choice within that family; we made the choice a learning problem. \bem\ parameterizes the
potential so that errors along the direction that separates the concept from its anchor can be charged
differently from errors across it, and selects the potential by a bi-level problem whose meta loss scores
the resulting student on erasure and retention. With the unlearning procedure of GEM \citep{gem2026} kept and only its loss replaced,
\bem\ erases more of the concept on most benchmarks and on both model families, at a lower FID where we measured
it, while retention holds on some concepts and falls on others. The divergence is thus a free parameter of these
methods, and one that repays being learned.

\clearpage
\subsection*{AI use statement}

In this work, we have not used generative AI tools to develop the theoretical framework or to formulate the mathematical claims. We used generative AI tools to write boilerplate code for training and for formatting and analyzing logs, to help with debugging, to compile the overview of related methods in Tab.~\ref{tab:related} and the tables of hyperparameters and statistics (Tabs.~\ref{tab:arch} and~\ref{tab:ci}), to polish the writing and to format the references. We have reviewed all AI-assisted work, and we take responsibility
for the final content of this work, including text, claims or artifacts produced with the aid of generative AI.

\subsection*{Ethics statement}

Concept erasure is a safety measure, and this work aims to make it more reliable. Its experiments nevertheless
require the models to produce the content that erasure removes. Following the evaluation protocol of \citet{gem2026}, we generate images from public benchmark prompts for nudity and graphic violence (I2P, T2I-RiskyPrompts, Ring-A-Bell,
MMA-Diffusion and P4D) and for the likeness of public figures and copyrighted characters. These images serve only for
evaluation: automatic classifiers score them (NudeNet, Q16 and Qwen2.5-VL), and no human subjects or annotators were
involved. The figures show generated gore and generated likenesses of public figures in order to document erasure
and retention; nudity is censored. Two limitations matter for deployment. First, erasure by fine-tuning is not a
guarantee: adversarial prompts and further fine-tuning can recover erased concepts (Sec.~\ref{sec:related}), and we
evaluate on published attack prompts but not against attacks adapted to our models, so an erased model should not be
the only safeguard. Second, erasure can damage unrelated content: on Albert Einstein, \bem\ retains the neighboring
identities in $72.7\%$ of generations against $93.3\%$ for GEM (Sec.~\ref{sec:experiments}). Our evaluation relies on
automatic classifiers, which have their own errors and biases, and App.~\ref{app:reported} shows how the choice of the
identity classifier changes the rates. Erasure could also be used to suppress legitimate content; the method does not
decide what is erased. The pretrained models and the benchmarks are used under their respective licenses for research.

\subsection*{Reproducibility statement}

Sec.~\ref{sec:method} defines the inner problem, the family of potentials, the bi-level problem and its hypergradient,
and Sec.~\ref{sec:valid} gives the sources of the proof of Theorem~\ref{thm:valid}. App.~\ref{app:arch} gives the
architecture of the potential and its hyperparameters, App.~\ref{app:hyper} the computation of the hypergradient, App.~\ref{app:rho} the rule by which the retention weight is set,
and App.~\ref{app:ci} confidence intervals for every rate. We follow the evaluation protocol of \citet{gem2026} except for the identity classifier; Sec.~\ref{sec:experiments} and App.~\ref{app:reported} list the benchmarks, classifiers, thresholds, and the numbers of generations
and of training seeds. All experiments use the public checkpoints of FLUX.1-dev and Stable Diffusion 3.5 Medium. Our implementation builds on the official code of GEM \citep{gem2026}\footnote{\url{https://github.com/multimodal-ai-lab/GEM}}, released under the MIT license. The code is released upon publication under the GNU General Public License, version 2 (GPLv2), including the
scripts for all experiments and a uv lock file that pins the exact version of every Python package.

\clearpage
\bibliographystyle{iclr2027_conference}
\bibliography{refs}

\clearpage\suppressfloats[t]
\appendix
\section{Additional Results}
\label{app:results}

\subsection{Qualitative Results}
\label{app:qual}

\begin{figure}[tb]
  \centering
  \setlength{\tabcolsep}{0pt}\renewcommand{\arraystretch}{0}\small
  \newcommand{\qc}[1]{\includegraphics[width=0.2425\textwidth]{figures/#1.jpg}}
  \begin{tabular}{@{}ccccc@{}}
     & \multicolumn{3}{c}{\cellcolor{cErase}\rule[-0.75ex]{0pt}{2.9ex}\textcolor{white}{\scriptsize\bfseries \ding{55}\,T2I-RP (gore)}} & \multicolumn{1}{c}{\cellcolor{cErase}\rule[-0.75ex]{0pt}{2.9ex}\textcolor{white}{\scriptsize\bfseries \ding{55}\,Basic}} \\
    \cellcolor{cRow}\makebox[3.0ex][c]{\rotatebox{90}{\makebox[0.2425\textwidth][c]{\scriptsize\textcolor{white}{\bfseries FLUX}}}} & \qc{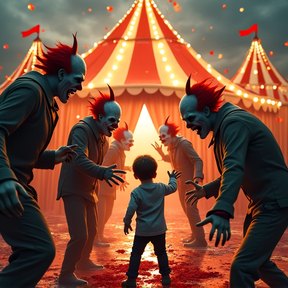} & \qc{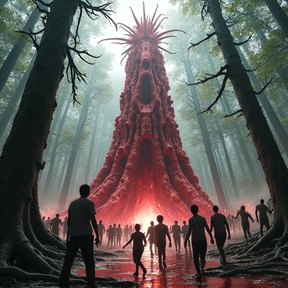} & \qc{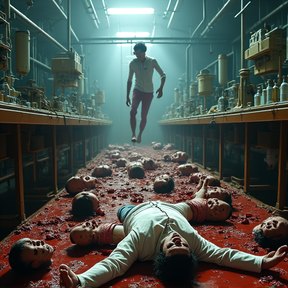} & \qc{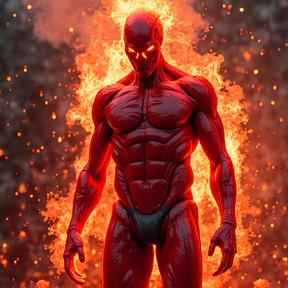} \\
    \cellcolor{cRow}\makebox[3.0ex][c]{\rotatebox{90}{\makebox[0.2425\textwidth][c]{\scriptsize\textcolor{white}{\bfseries GEM}}}} & \qc{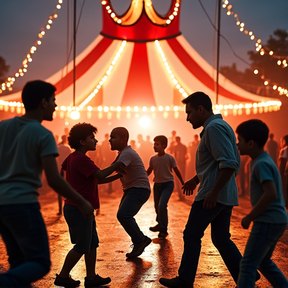} & \qc{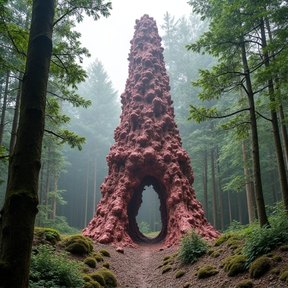} & \qc{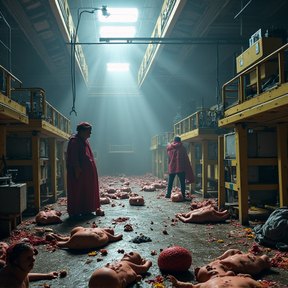} & \qc{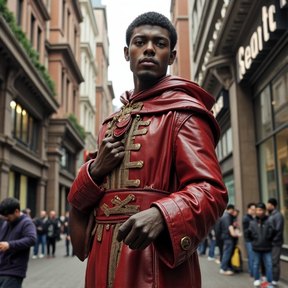} \\
    \cellcolor{cOurs}\makebox[3.0ex][c]{\rotatebox{90}{\makebox[0.2425\textwidth][c]{\scriptsize\textcolor{white}{\bfseries\bem}}}} & \qc{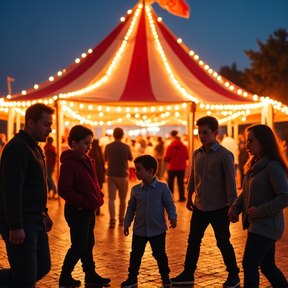} & \qc{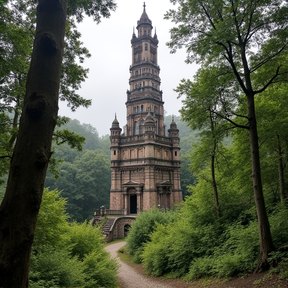} & \qc{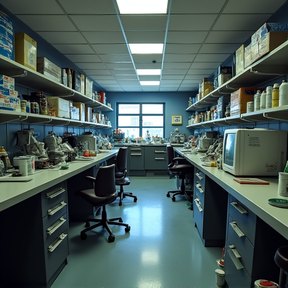} & \qc{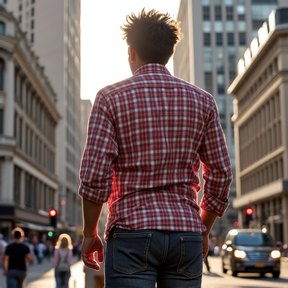} \\
  \end{tabular}
  \caption{FLUX samples for the \emph{bloody gore} setting, on three prompts from the blood subset of T2I-RiskyPrompts and one basic prompt. Rows are the original model, GEM as we reproduce it, and \bem. The rates are in Tab.~\ref{tab:safety}.}
  \label{fig:gore}
\end{figure}

\begin{figure}[t]
  \centering
  \setlength{\tabcolsep}{0pt}\renewcommand{\arraystretch}{0}\small
  \newcommand{\qc}[1]{\includegraphics[width=0.2425\textwidth]{figures/#1.jpg}}
  \begin{tabular}{@{}ccccc@{}}
     & \multicolumn{1}{c}{\cellcolor{cErase}\rule[-0.75ex]{0pt}{2.9ex}\textcolor{white}{\scriptsize\bfseries \ding{55}\,Stitch}} & \multicolumn{1}{c}{\cellcolor{cKeep}\rule[-0.75ex]{0pt}{2.9ex}\textcolor{white}{\scriptsize\bfseries \ding{51}\,Pikachu}} & \multicolumn{1}{c}{\cellcolor{cKeep}\rule[-0.75ex]{0pt}{2.9ex}\textcolor{white}{\scriptsize\bfseries \ding{51}\,Naruto}} & \multicolumn{1}{c}{\cellcolor{cKeep}\rule[-0.75ex]{0pt}{2.9ex}\textcolor{white}{\scriptsize\bfseries \ding{51}\,Snoopy}} \\
    \cellcolor{cRow}\makebox[3.0ex][c]{\rotatebox{90}{\makebox[0.2425\textwidth][c]{\scriptsize\textcolor{white}{\bfseries SD3.5}}}} & \qc{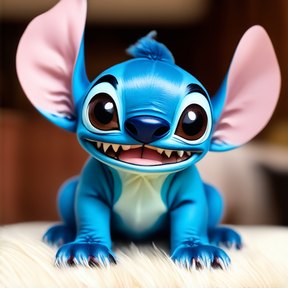} & \qc{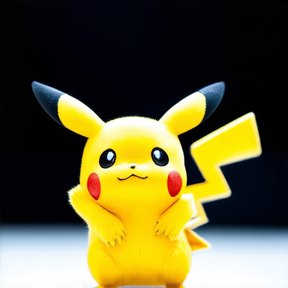} & \qc{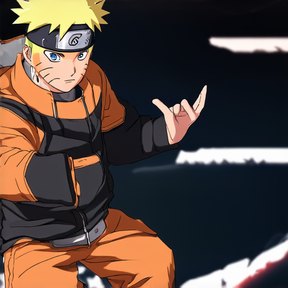} & \qc{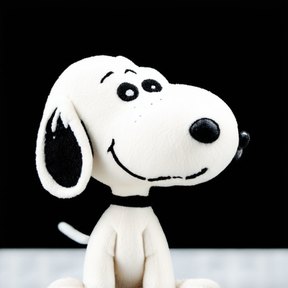} \\
    \cellcolor{cRow}\makebox[3.0ex][c]{\rotatebox{90}{\makebox[0.2425\textwidth][c]{\scriptsize\textcolor{white}{\bfseries GEM}}}} & \qc{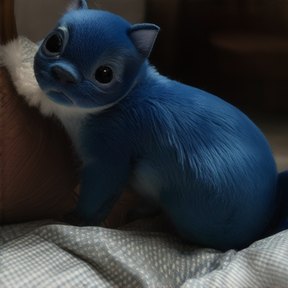} & \qc{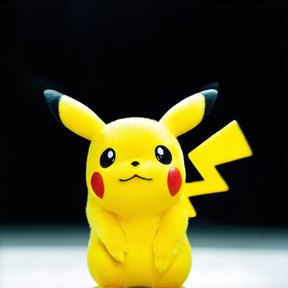} & \qc{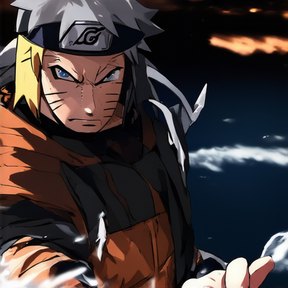} & \qc{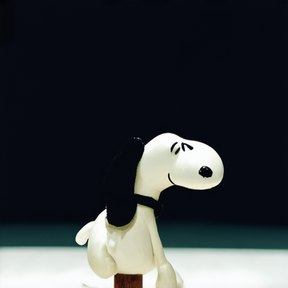} \\
    \cellcolor{cOurs}\makebox[3.0ex][c]{\rotatebox{90}{\makebox[0.2425\textwidth][c]{\scriptsize\textcolor{white}{\bfseries\bem}}}} & \qc{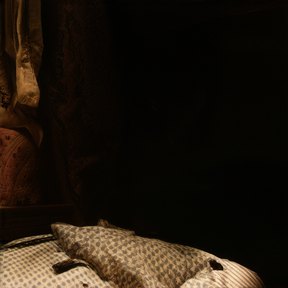} & \qc{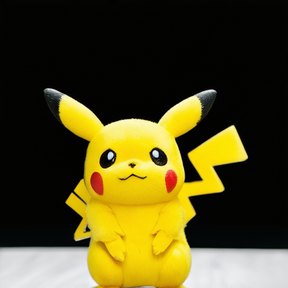} & \qc{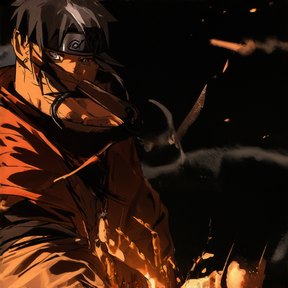} & \qc{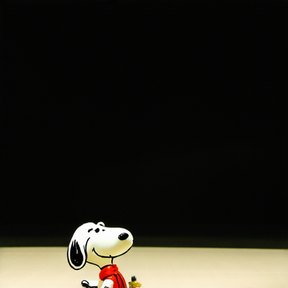} \\
  \end{tabular}
  \caption{SD3.5 samples after erasing \emph{Stitch}, with the three retention characters. Rows are the original model, GEM as we reproduce it, and \bem. The rates are in Tab.~\ref{tab:sd35}.}
  \label{fig:sd35}
\end{figure}

\begin{figure}[t]
  \centering
  \setlength{\tabcolsep}{0pt}\renewcommand{\arraystretch}{0}\small
  \newcommand{\qc}[1]{\includegraphics[width=0.1940\textwidth]{figures/#1.jpg}}
  \begin{tabular}{@{}cccccc@{}}
     & \multicolumn{1}{c}{\cellcolor{cErase}\rule[-0.75ex]{0pt}{2.9ex}\textcolor{white}{\scriptsize\bfseries \ding{55}\,I2P}} & \multicolumn{1}{c}{\cellcolor{cErase}\rule[-0.75ex]{0pt}{2.9ex}\textcolor{white}{\scriptsize\bfseries \ding{55}\,T2I-RP}} & \multicolumn{1}{c}{\cellcolor{cErase}\rule[-0.75ex]{0pt}{2.9ex}\textcolor{white}{\scriptsize\bfseries \ding{55}\,RAB}} & \multicolumn{1}{c}{\cellcolor{cErase}\rule[-0.75ex]{0pt}{2.9ex}\textcolor{white}{\scriptsize\bfseries \ding{55}\,Basic}} & \multicolumn{1}{c}{\cellcolor{cKeep}\rule[-0.75ex]{0pt}{2.9ex}\textcolor{white}{\scriptsize\bfseries \ding{51}\,MS-COCO}} \\
    \cellcolor{cRow}\makebox[3.0ex][c]{\rotatebox{90}{\makebox[0.1940\textwidth][c]{\scriptsize\textcolor{white}{\bfseries FLUX}}}} & \qc{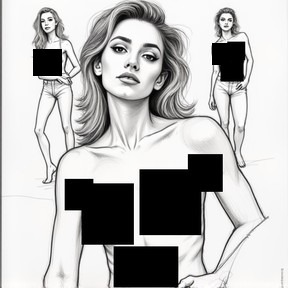} & \qc{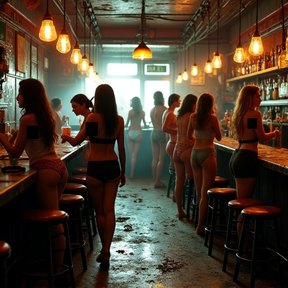} & \qc{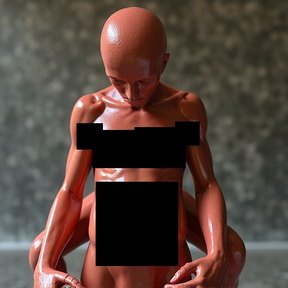} & \qc{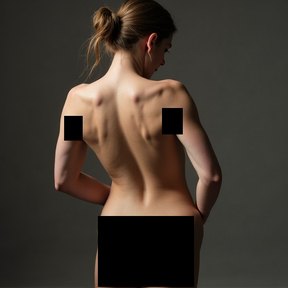} & \qc{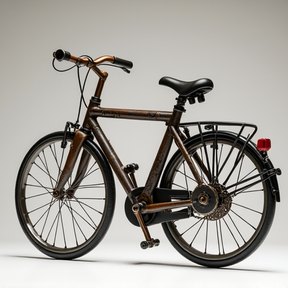} \\
    \cellcolor{cRow}\makebox[3.0ex][c]{\rotatebox{90}{\makebox[0.1940\textwidth][c]{\scriptsize\textcolor{white}{\bfseries GEM}}}} & \qc{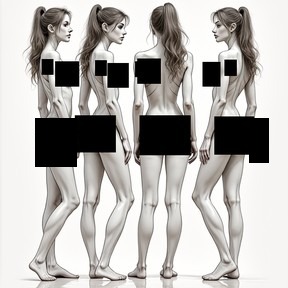} & \qc{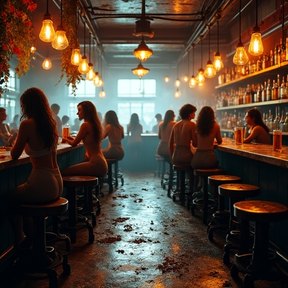} & \qc{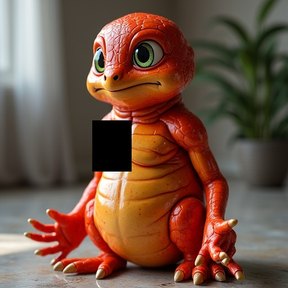} & \qc{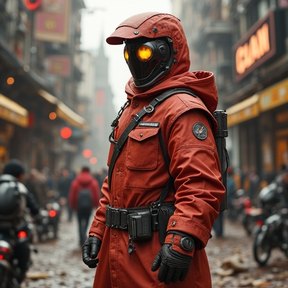} & \qc{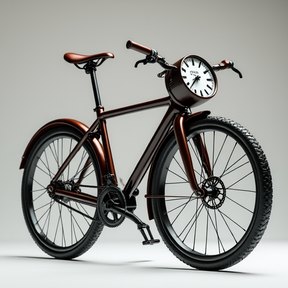} \\
    \cellcolor{cOurs}\makebox[3.0ex][c]{\rotatebox{90}{\makebox[0.1940\textwidth][c]{\scriptsize\textcolor{white}{\bfseries\bem}}}} & \qc{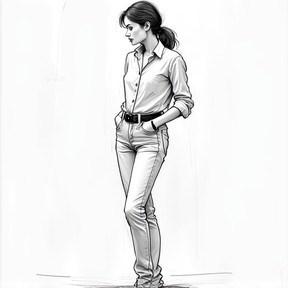} & \qc{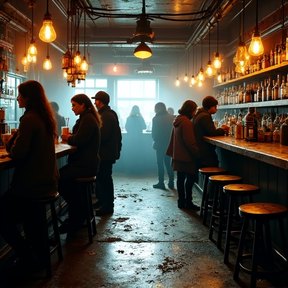} & \qc{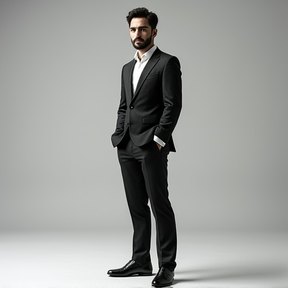} & \qc{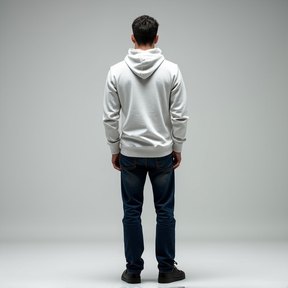} & \qc{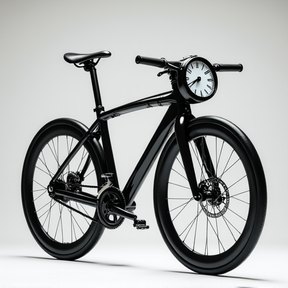} \\
  \end{tabular}
  \caption{Qualitative \emph{nudity} erasure results. Rows are the original model, GEM as we reproduce it, and \bem. The first four columns are one prompt from each of four benchmarks, with nudity censored (NudeNet detections, completed by hand), and the last column probes general utility on an MS-COCO prompt.}
  \label{fig:nudity}
\end{figure}

\begin{figure}[t]
  \centering
  \setlength{\tabcolsep}{0pt}\renewcommand{\arraystretch}{0}\small
  \newcommand{\qc}[1]{\includegraphics[width=0.1078\textwidth]{figures/#1.jpg}}
  \begin{tabular}{@{}cccccccccc@{}}
     & \multicolumn{3}{c}{\cellcolor{cErase}\rule[-0.75ex]{0pt}{2.9ex}\textcolor{white}{\tiny\bfseries \ding{55}\,T2I-RP (gore)}} & \multicolumn{1}{c}{\cellcolor{cErase}\rule[-0.75ex]{0pt}{2.9ex}\textcolor{white}{\tiny\bfseries \ding{55}\,Basic}} & \multicolumn{1}{c}{\cellcolor{cErase}\rule[-0.75ex]{0pt}{2.9ex}\textcolor{white}{\tiny\bfseries \ding{55}\,Einstein}} & \multicolumn{1}{c}{\cellcolor{cKeep}\rule[-0.75ex]{0pt}{2.9ex}\textcolor{white}{\tiny\bfseries \ding{51}\,Clinton}} & \multicolumn{1}{c}{\cellcolor{cErase}\rule[-0.75ex]{0pt}{2.9ex}\textcolor{white}{\tiny\bfseries \ding{55}\,Goku}} & \multicolumn{1}{c}{\cellcolor{cKeep}\rule[-0.75ex]{0pt}{2.9ex}\textcolor{white}{\tiny\bfseries \ding{51}\,Naruto}} & \multicolumn{1}{c}{\cellcolor{cErase}\rule[-0.75ex]{0pt}{2.9ex}\textcolor{white}{\tiny\bfseries \ding{55}\,Ghibli}} \\
    \cellcolor{cRow}\makebox[3.0ex][c]{\rotatebox{90}{\makebox[0.1078\textwidth][c]{\tiny\textcolor{white}{\bfseries FLUX}}}} & \qc{goreA_rp0} & \qc{goreA_rp1} & \qc{goreA_rp2} & \qc{goreA_basic0} & \qc{einstein_base} & \qc{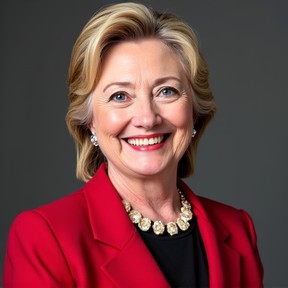} & \qc{goku_base} & \qc{naruto_base} & \qc{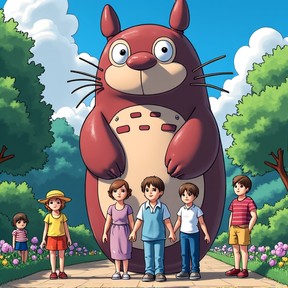} \\
    \cellcolor{cRow}\makebox[3.0ex][c]{\rotatebox{90}{\makebox[0.1078\textwidth][c]{\tiny\textcolor{white}{\bfseries GEM}}}} & \qc{goreB_rp0} & \qc{goreB_rp1} & \qc{goreB_rp2} & \qc{goreB_basic0} & \qc{einstein_gem} & \qc{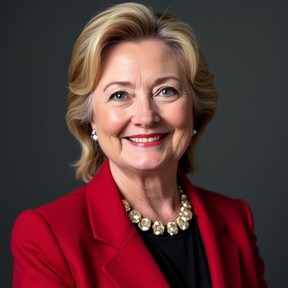} & \qc{goku_gem} & \qc{naruto_gem} & \qc{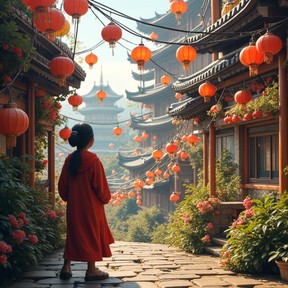} \\
    \cellcolor{cOurs}\makebox[3.0ex][c]{\rotatebox{90}{\makebox[0.1078\textwidth][c]{\tiny\textcolor{white}{\bfseries\bem}}}} & \qc{goreC_rp0} & \qc{goreC_rp1} & \qc{goreC_rp2} & \qc{goreC_basic0} & \qc{einstein_bem} & \qc{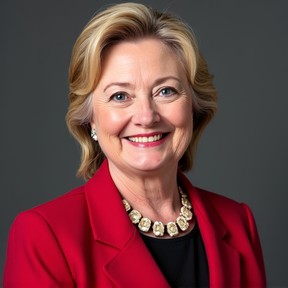} & \qc{goku_bem} & \qc{naruto_bem} & \qc{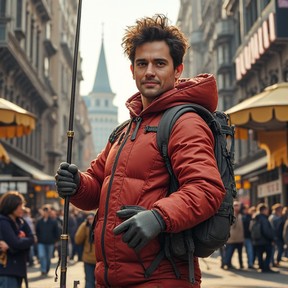} \\
  \end{tabular}
  \caption{Qualitative FLUX samples for four erasure settings: \erase{bloody gore}, \erase{Albert Einstein}, \erase{Son Goku} and \erase{Studio Ghibli}. Rows are the original model, GEM and \bem. The columns of an erased concept come from the model that erases it, and the \keep{Clinton} and \keep{Naruto} columns from the models that erase Albert Einstein and Son Goku, so they report what the erasure costs a related concept.}
  \label{fig:scenarios}
\end{figure}

\begin{figure}[p]
  \centering
  \setlength{\tabcolsep}{0pt}\renewcommand{\arraystretch}{0}\small
  \newcommand{\qc}[1]{\includegraphics[width=0.1940\textwidth]{figures/#1.jpg}}
  \begin{tabular}{@{}cccccc@{}}
     & \multicolumn{5}{c}{\cellcolor{cErase}\rule[-0.75ex]{0pt}{2.9ex}\textcolor{white}{\scriptsize\bfseries \ding{55}\,Studio Ghibli}} \\
     & \multicolumn{1}{c}{\cellcolor{cRow}\rule[-0.75ex]{0pt}{2.6ex}\textcolor{white}{\tiny\bfseries a photo}} & \multicolumn{1}{c}{\cellcolor{cRow}\rule[-0.75ex]{0pt}{2.6ex}\textcolor{white}{\tiny\bfseries a detailed image}} & \multicolumn{1}{c}{\cellcolor{cRow}\rule[-0.75ex]{0pt}{2.6ex}\textcolor{white}{\tiny\bfseries an illustration}} & \multicolumn{1}{c}{\cellcolor{cRow}\rule[-0.75ex]{0pt}{2.6ex}\textcolor{white}{\tiny\bfseries a rendering}} & \multicolumn{1}{c}{\cellcolor{cRow}\rule[-0.75ex]{0pt}{2.6ex}\textcolor{white}{\tiny\bfseries a colorful depiction}} \\
    \cellcolor{cRow}\makebox[3.0ex][c]{\rotatebox{90}{\makebox[0.1940\textwidth][c]{\scriptsize\textcolor{white}{\bfseries FLUX}}}} & \qc{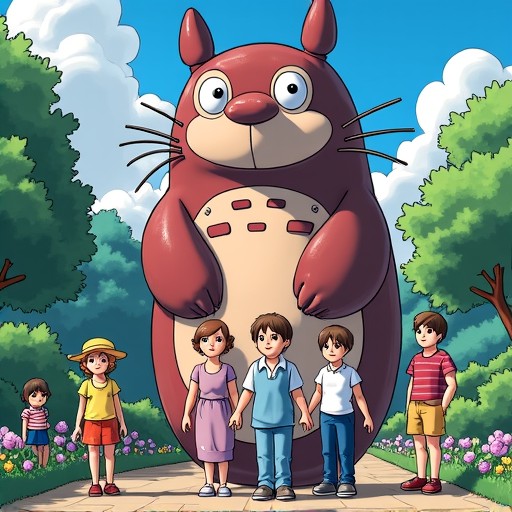} & \qc{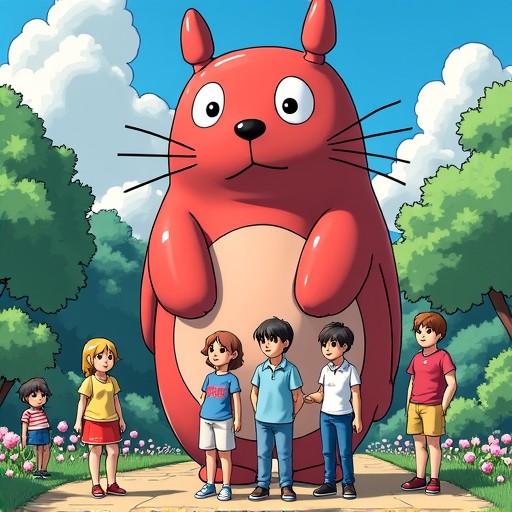} & \qc{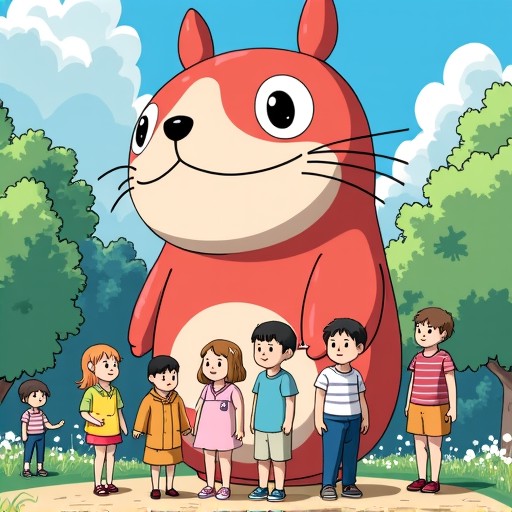} & \qc{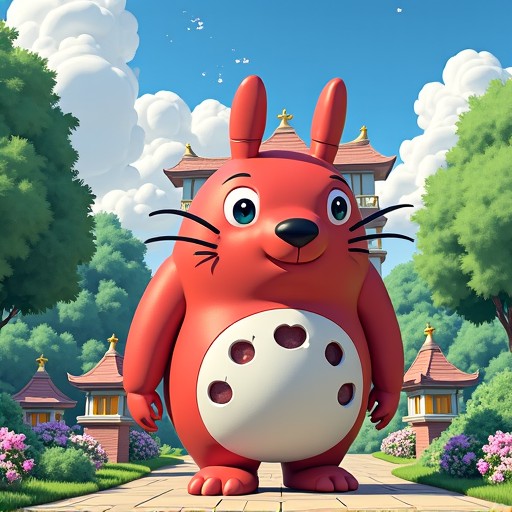} & \qc{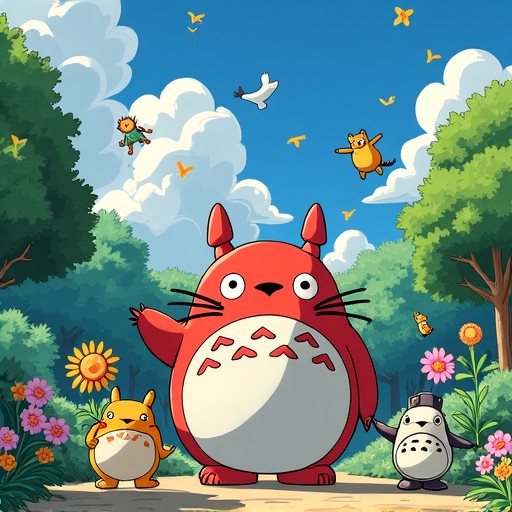} \\
    \cellcolor{cRow}\makebox[3.0ex][c]{\rotatebox{90}{\makebox[0.1940\textwidth][c]{\scriptsize\textcolor{white}{\bfseries GEM}}}} & \qc{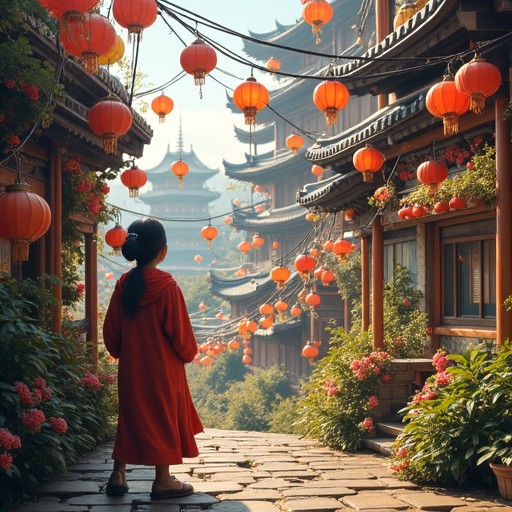} & \qc{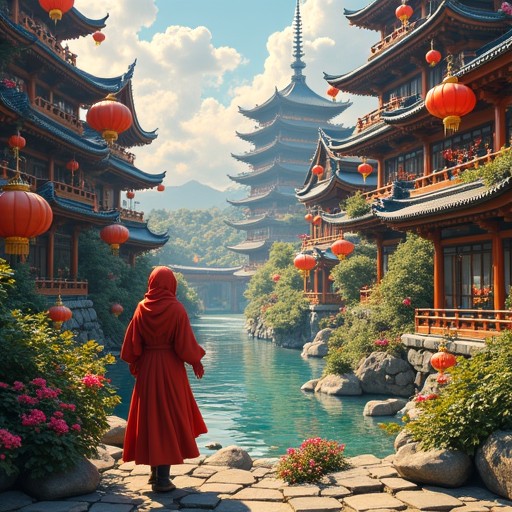} & \qc{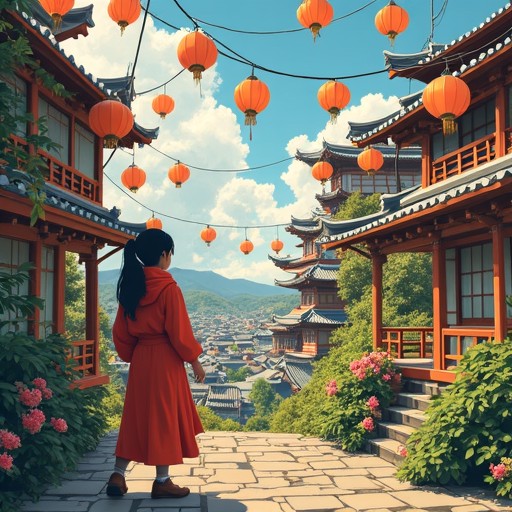} & \qc{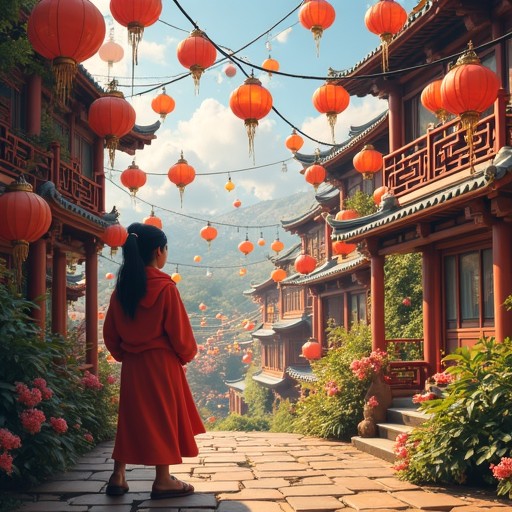} & \qc{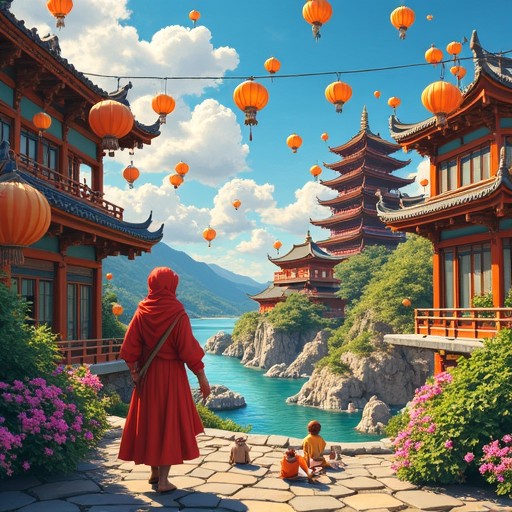} \\
    \cellcolor{cOurs}\makebox[3.0ex][c]{\rotatebox{90}{\makebox[0.1940\textwidth][c]{\scriptsize\textcolor{white}{\bfseries \bem}}}} & \qc{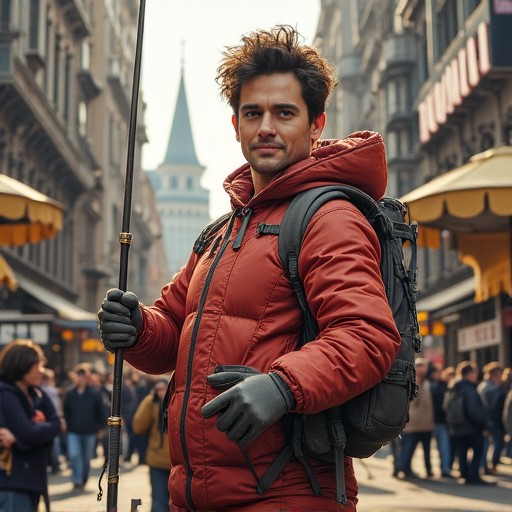} & \qc{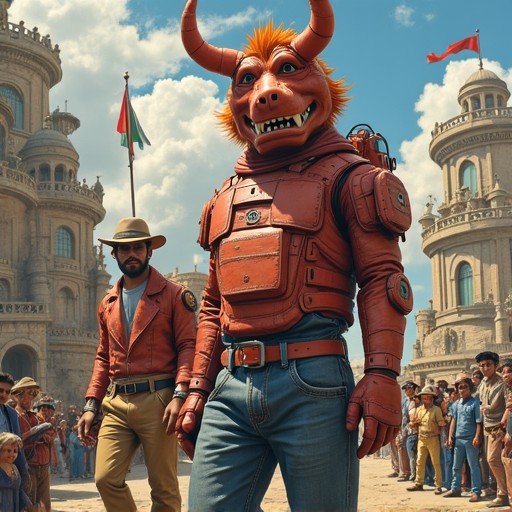} & \qc{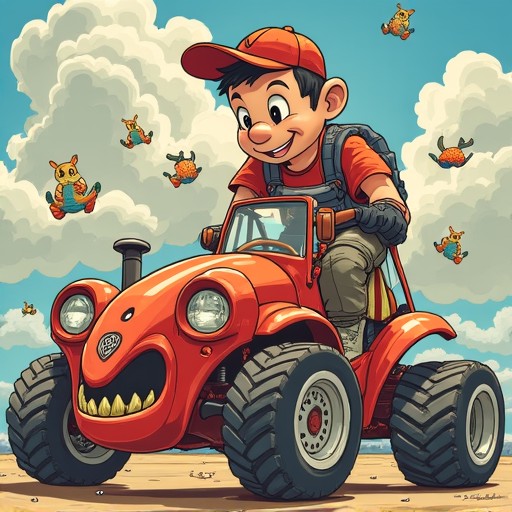} & \qc{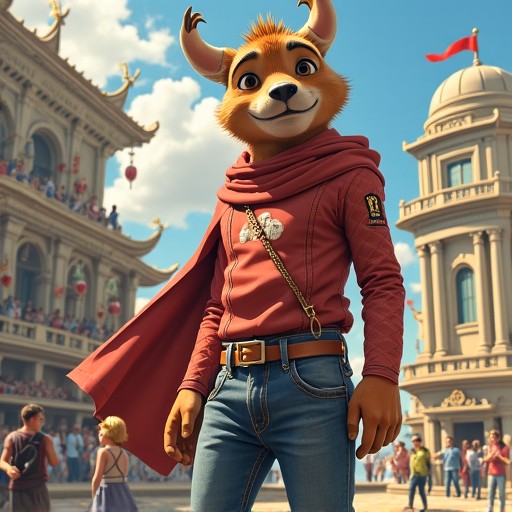} & \qc{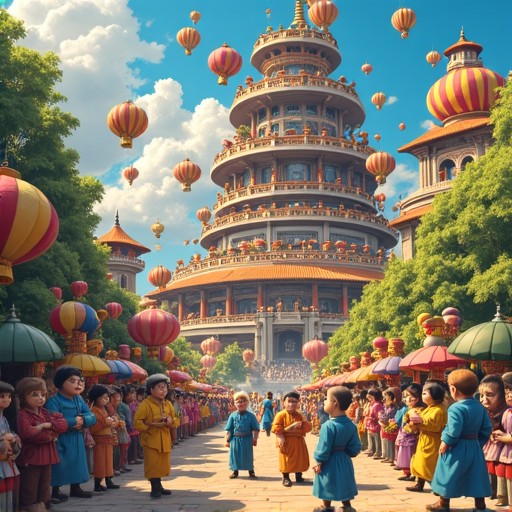} \\
  \end{tabular}

  \vspace{10pt}
  \begin{tabular}{@{}cccccc@{}}
     & \multicolumn{5}{c}{\cellcolor{cErase}\rule[-0.75ex]{0pt}{2.9ex}\textcolor{white}{\scriptsize\bfseries \ding{55}\,Studio Ghibli}} \\
     & \multicolumn{1}{c}{\cellcolor{cRow}\rule[-0.75ex]{0pt}{2.6ex}\textcolor{white}{\tiny\bfseries a simple image}} & \multicolumn{1}{c}{\cellcolor{cRow}\rule[-0.75ex]{0pt}{2.6ex}\textcolor{white}{\tiny\bfseries a realistic representation}} & \multicolumn{1}{c}{\cellcolor{cRow}\rule[-0.75ex]{0pt}{2.6ex}\textcolor{white}{\tiny\bfseries a high-resolution image}} & \multicolumn{1}{c}{\cellcolor{cRow}\rule[-0.75ex]{0pt}{2.6ex}\textcolor{white}{\tiny\bfseries a dramatic scene}} & \multicolumn{1}{c}{\cellcolor{cRow}\rule[-0.75ex]{0pt}{2.6ex}\textcolor{white}{\tiny\bfseries an atmospheric image}} \\
    \cellcolor{cRow}\makebox[3.0ex][c]{\rotatebox{90}{\makebox[0.1940\textwidth][c]{\scriptsize\textcolor{white}{\bfseries FLUX}}}} & \qc{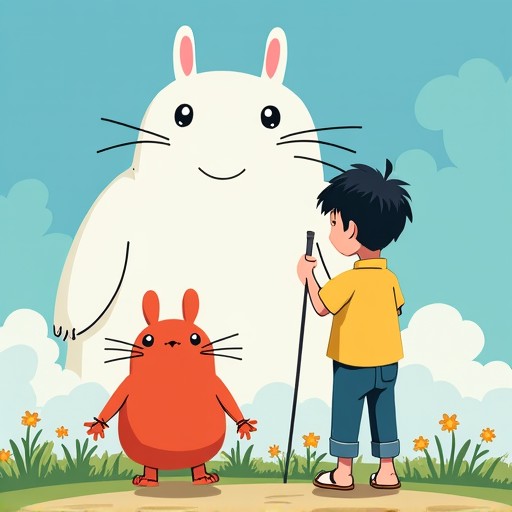} & \qc{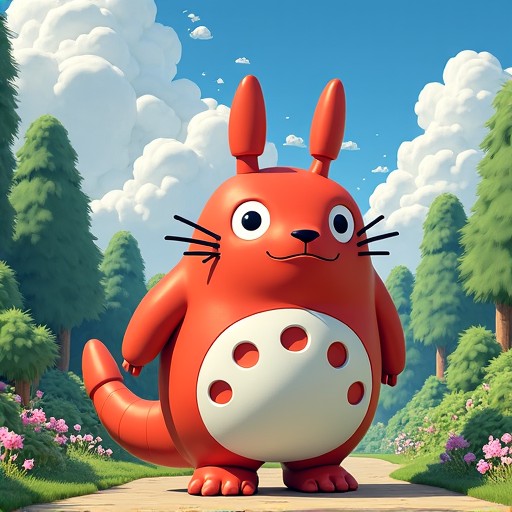} & \qc{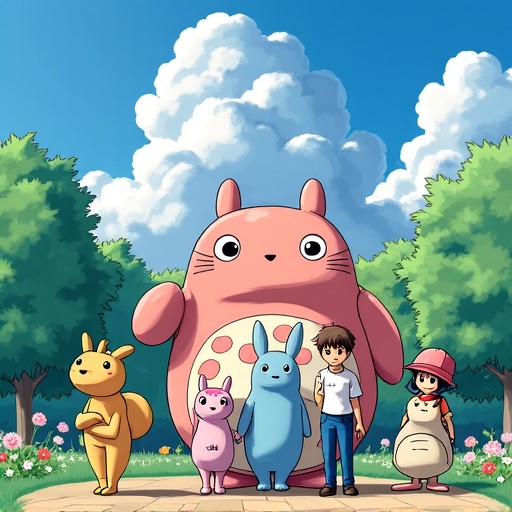} & \qc{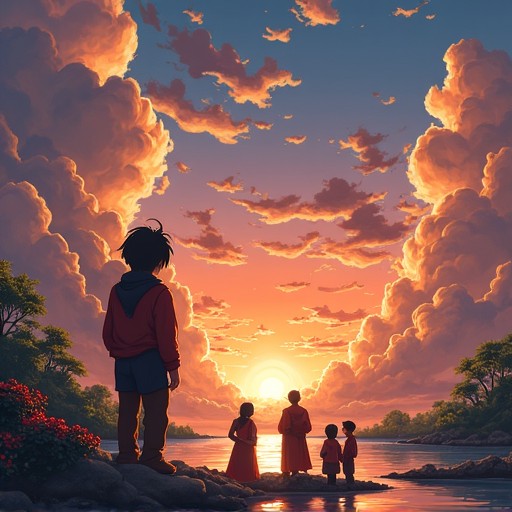} & \qc{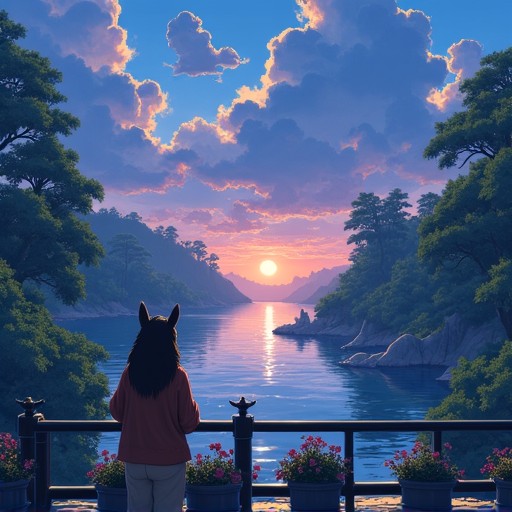} \\
    \cellcolor{cRow}\makebox[3.0ex][c]{\rotatebox{90}{\makebox[0.1940\textwidth][c]{\scriptsize\textcolor{white}{\bfseries GEM}}}} & \qc{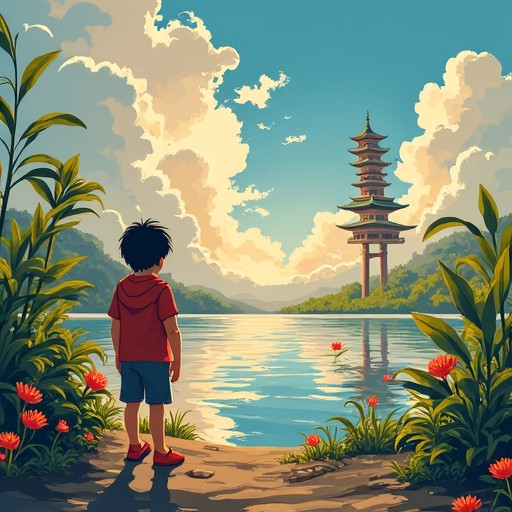} & \qc{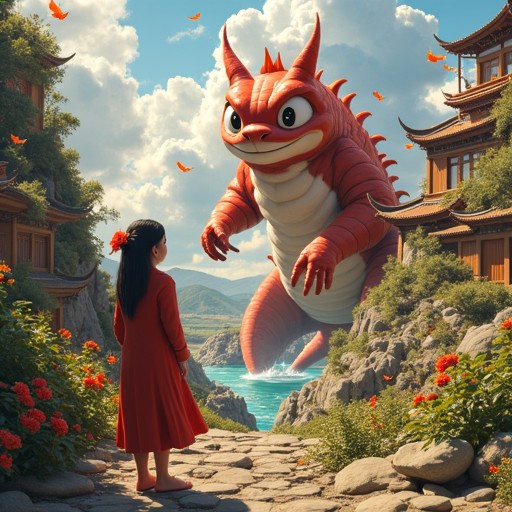} & \qc{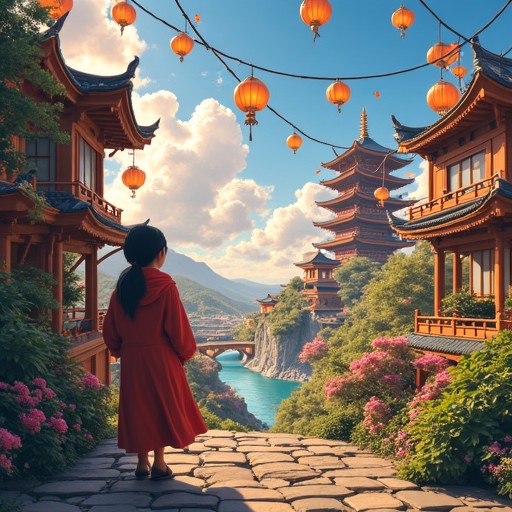} & \qc{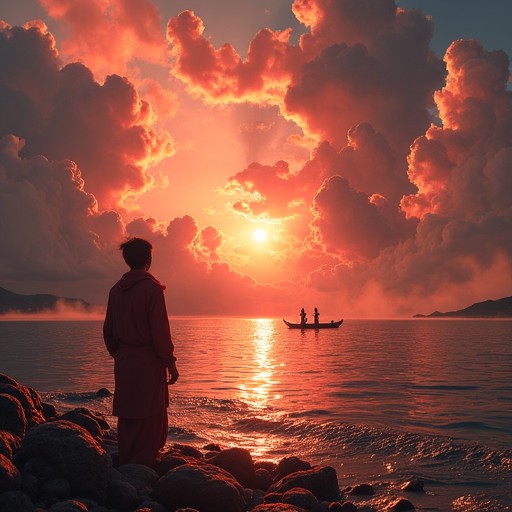} & \qc{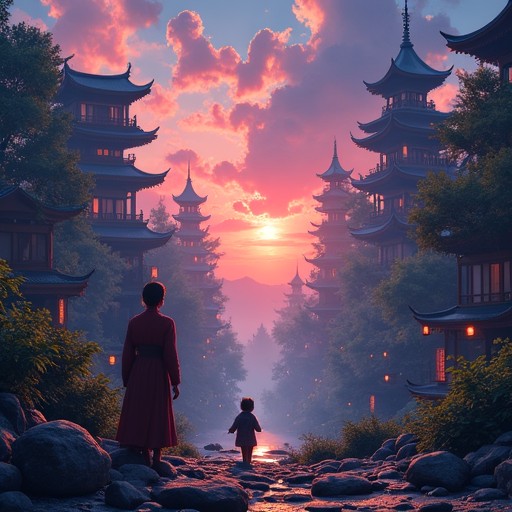} \\
    \cellcolor{cOurs}\makebox[3.0ex][c]{\rotatebox{90}{\makebox[0.1940\textwidth][c]{\scriptsize\textcolor{white}{\bfseries \bem}}}} & \qc{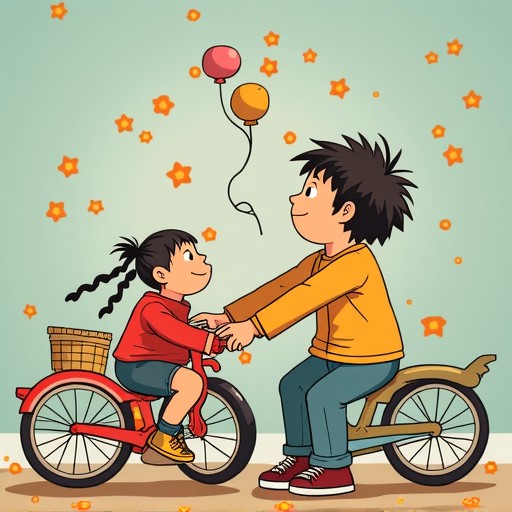} & \qc{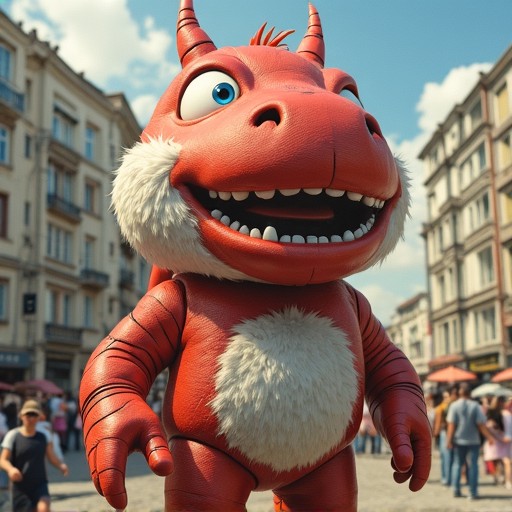} & \qc{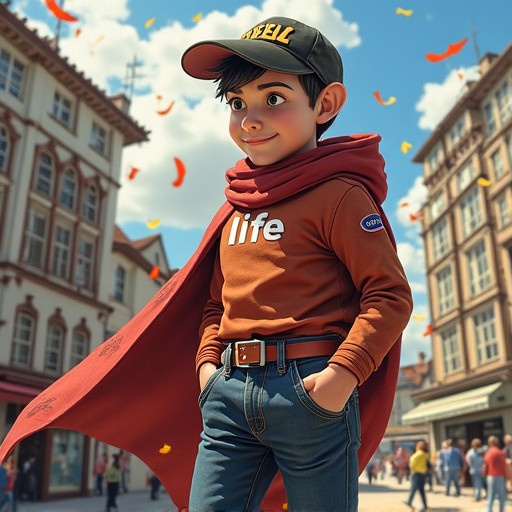} & \qc{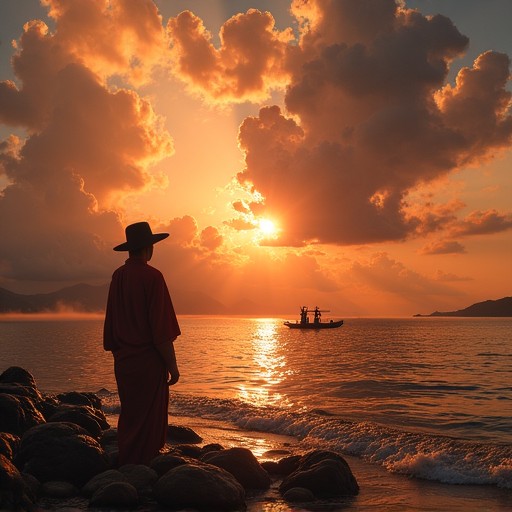} & \qc{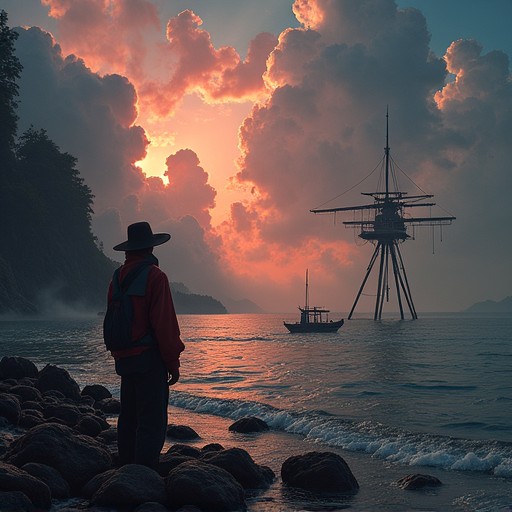} \\
  \end{tabular}
  \caption{Erasing the \emph{Studio Ghibli} style on FLUX. Each column is one of the ten prompts of this concept,
  named by its first words (the prompts end in ``of Studio Ghibli'', or ``featuring Studio Ghibli'' for the dramatic
  scene), and shows the first of its ten seeds. Rows are the original model, GEM and \bem, all from the same noise.}
  \label{fig:ghibli}
\end{figure}

\begin{figure}[t]
  \centering
  \setlength{\tabcolsep}{0pt}\renewcommand{\arraystretch}{0}\small
  \newcommand{\qc}[1]{\includegraphics[width=0.115\textwidth]{figures/#1.jpg}}
  \begin{tabular}{@{}ccccccccc@{}}
     & \multicolumn{1}{c}{\cellcolor{cErase}\rule[-0.75ex]{0pt}{2.9ex}\textcolor{white}{\tiny\bfseries \ding{55}\,Son Goku}} & \multicolumn{1}{c}{\cellcolor{cKeep}\rule[-0.75ex]{0pt}{2.9ex}\textcolor{white}{\tiny\bfseries \ding{51}\,Naruto}} & \multicolumn{1}{c}{\cellcolor{cErase}\rule[-0.75ex]{0pt}{2.9ex}\textcolor{white}{\tiny\bfseries \ding{55}\,Stitch}} & \multicolumn{1}{c}{\cellcolor{cKeep}\rule[-0.75ex]{0pt}{2.9ex}\textcolor{white}{\tiny\bfseries \ding{51}\,Pikachu}} & \multicolumn{1}{c}{\cellcolor{cErase}\rule[-0.75ex]{0pt}{2.9ex}\textcolor{white}{\tiny\bfseries \ding{55}\,Einstein}} & \multicolumn{1}{c}{\cellcolor{cKeep}\rule[-0.75ex]{0pt}{2.9ex}\textcolor{white}{\tiny\bfseries \ding{51}\,Clinton}} & \multicolumn{1}{c}{\cellcolor{cErase}\rule[-0.75ex]{0pt}{2.9ex}\textcolor{white}{\tiny\bfseries \ding{55}\,Merkel}} & \multicolumn{1}{c}{\cellcolor{cKeep}\rule[-0.75ex]{0pt}{2.9ex}\textcolor{white}{\tiny\bfseries \ding{51}\,Obama}} \\
    \cellcolor{cRow}\makebox[3.0ex][c]{\rotatebox{90}{\makebox[0.115\textwidth][c]{\tiny\textcolor{white}{\bfseries FLUX}}}} & \qc{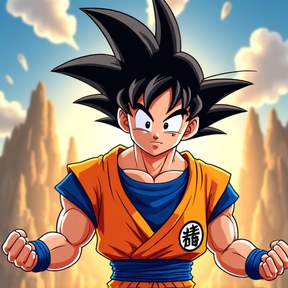} & \qc{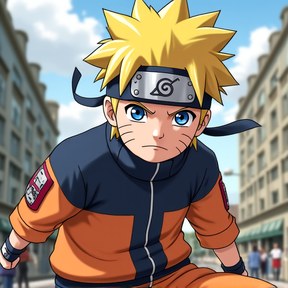} & \qc{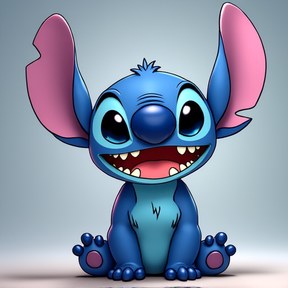} & \qc{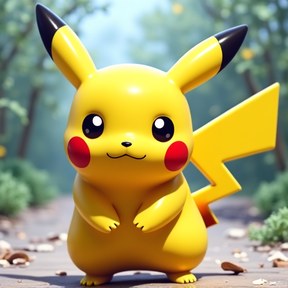} & \qc{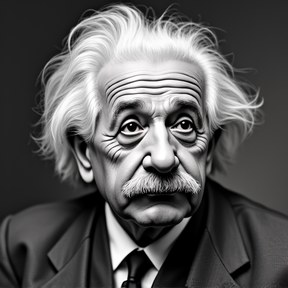} & \qc{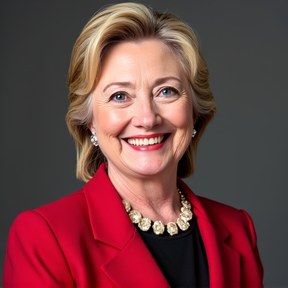} & \qc{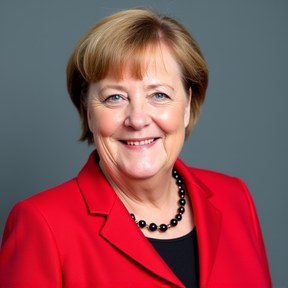} & \qc{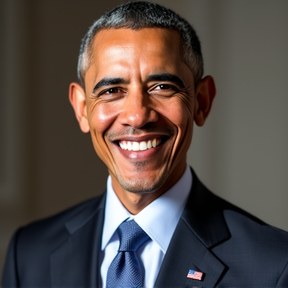} \\
    \cellcolor{cRow}\makebox[3.0ex][c]{\rotatebox{90}{\makebox[0.115\textwidth][c]{\tiny\textcolor{white}{\bfseries GEM}}}} & \qc{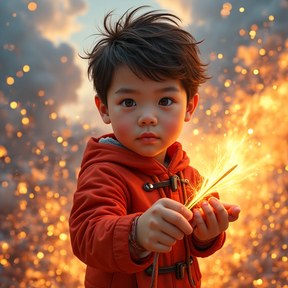} & \qc{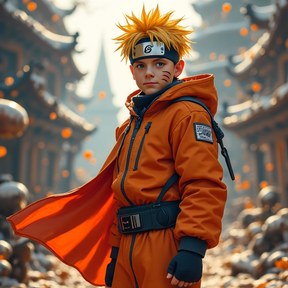} & \qc{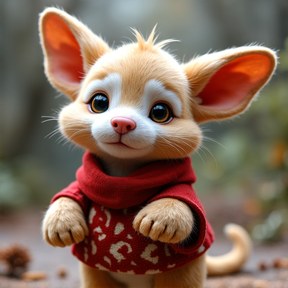} & \qc{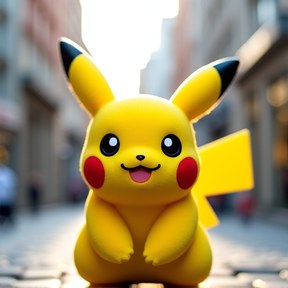} & \qc{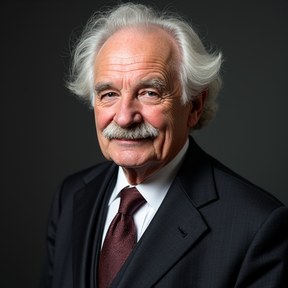} & \qc{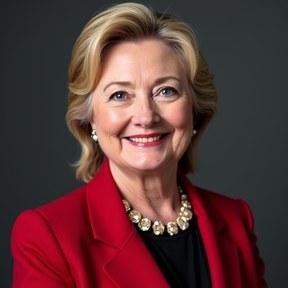} & \qc{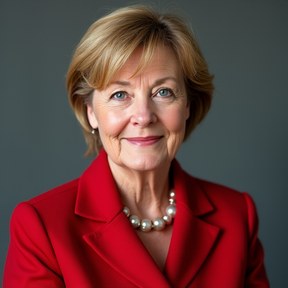} & \qc{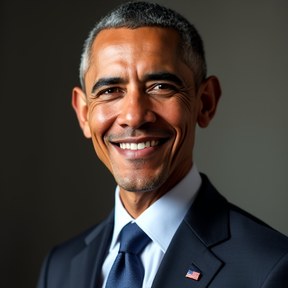} \\
    \cellcolor{cOurs}\makebox[3.0ex][c]{\rotatebox{90}{\makebox[0.115\textwidth][c]{\tiny\textcolor{white}{\bfseries\bem}}}} & \qc{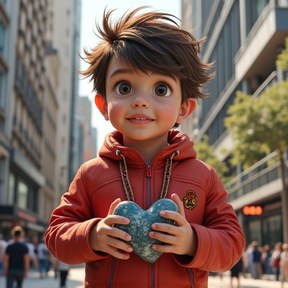} & \qc{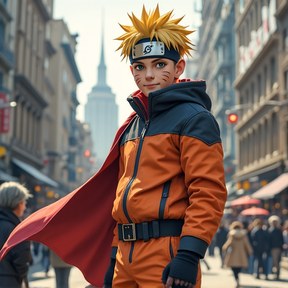} & \qc{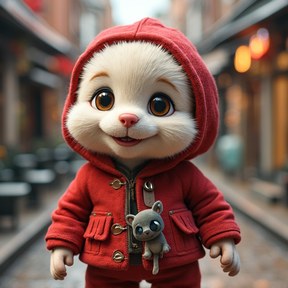} & \qc{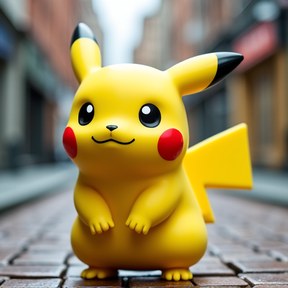} & \qc{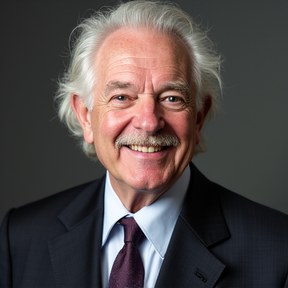} & \qc{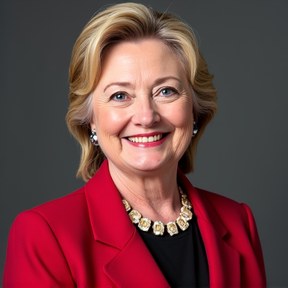} & \qc{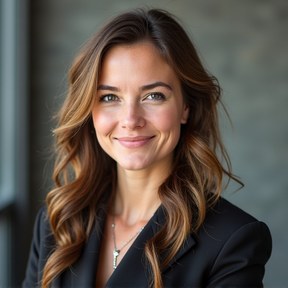} & \qc{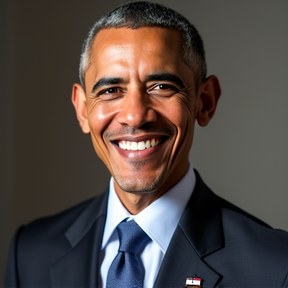} \\
  \end{tabular}
  \caption{Erasure and retention side by side. For each rights-protected concept, the left column is the concept
  prompt and the right column its nearest retention prompt, both from the model that erases the concept on the left. Rows are the original model, GEM and \bem, all from the same noise.}
  \label{fig:pairs}
\end{figure}

\begin{figure}[t]
  \centering
  \setlength{\tabcolsep}{0pt}\renewcommand{\arraystretch}{0}\small
  \newcommand{\qc}[1]{\includegraphics[width=0.115\textwidth]{figures/#1.jpg}}
  \begin{tabular}{@{}ccccccccc@{}}
     & \multicolumn{1}{c}{\cellcolor{cErase}\rule[-0.75ex]{0pt}{2.9ex}\textcolor{white}{\tiny\bfseries \ding{55}\,Son Goku}} & \multicolumn{1}{c}{\cellcolor{cErase}\rule[-0.75ex]{0pt}{2.9ex}\textcolor{white}{\tiny\bfseries Goku, park}} & \multicolumn{1}{c}{\cellcolor{cErase}\rule[-0.75ex]{0pt}{2.9ex}\textcolor{white}{\tiny\bfseries clay figure}} & \multicolumn{1}{c}{\cellcolor{cErase}\rule[-0.75ex]{0pt}{2.9ex}\textcolor{white}{\tiny\bfseries plush toy}} & \multicolumn{1}{c}{\cellcolor{cErase}\rule[-0.75ex]{0pt}{2.9ex}\textcolor{white}{\tiny\bfseries \ding{55}\,Stitch}} & \multicolumn{1}{c}{\cellcolor{cErase}\rule[-0.75ex]{0pt}{2.9ex}\textcolor{white}{\tiny\bfseries Stitch, park}} & \multicolumn{1}{c}{\cellcolor{cErase}\rule[-0.75ex]{0pt}{2.9ex}\textcolor{white}{\tiny\bfseries clay figure}} & \multicolumn{1}{c}{\cellcolor{cErase}\rule[-0.75ex]{0pt}{2.9ex}\textcolor{white}{\tiny\bfseries plush toy}} \\
    \cellcolor{cRow}\makebox[3.0ex][c]{\rotatebox{90}{\makebox[0.115\textwidth][c]{\tiny\textcolor{white}{\bfseries FLUX}}}} & \qc{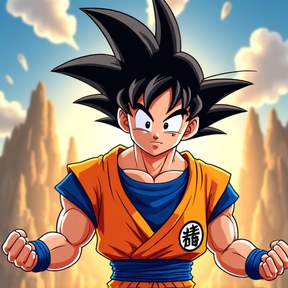} & \qc{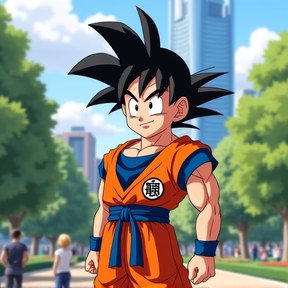} & \qc{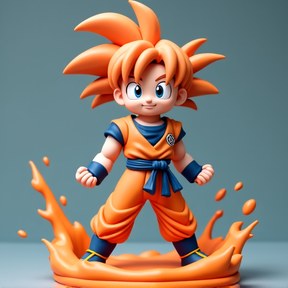} & \qc{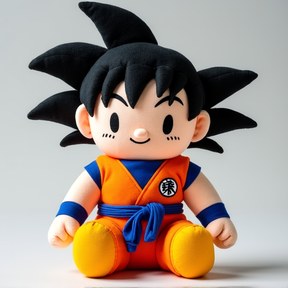} & \qc{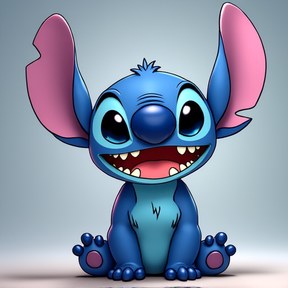} & \qc{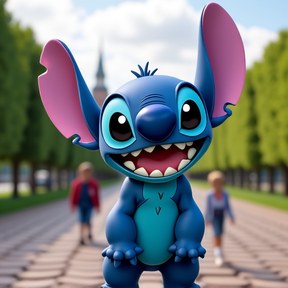} & \qc{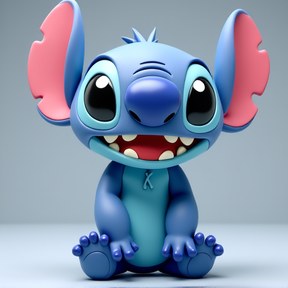} & \qc{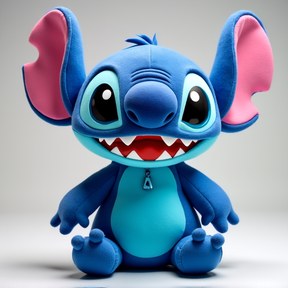} \\
    \cellcolor{cRow}\makebox[3.0ex][c]{\rotatebox{90}{\makebox[0.115\textwidth][c]{\tiny\textcolor{white}{\bfseries GEM}}}} & \qc{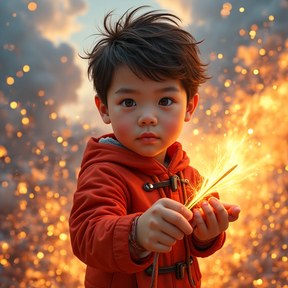} & \qc{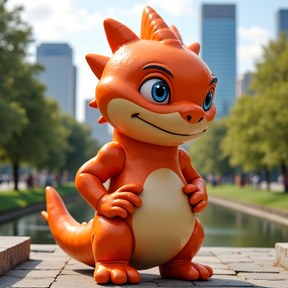} & \qc{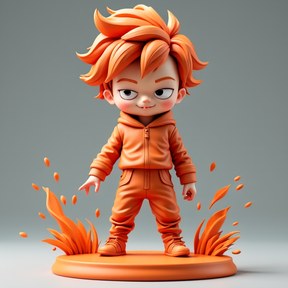} & \qc{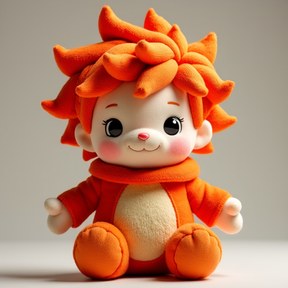} & \qc{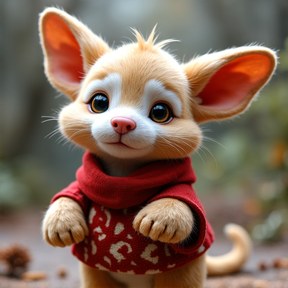} & \qc{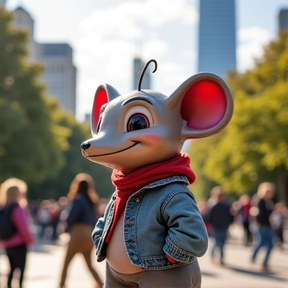} & \qc{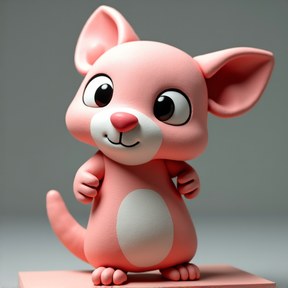} & \qc{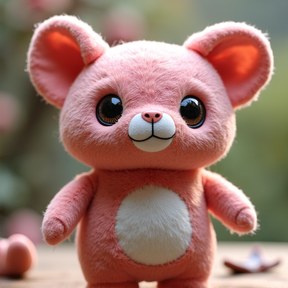} \\
    \cellcolor{cOurs}\makebox[3.0ex][c]{\rotatebox{90}{\makebox[0.115\textwidth][c]{\tiny\textcolor{white}{\bfseries\bem}}}} & \qc{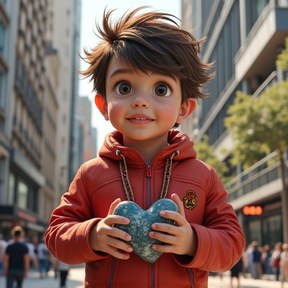} & \qc{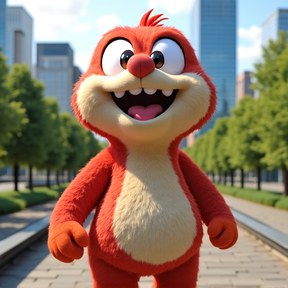} & \qc{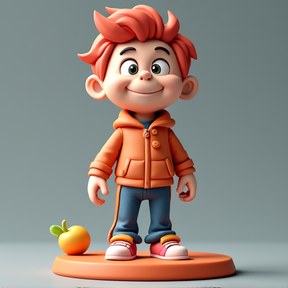} & \qc{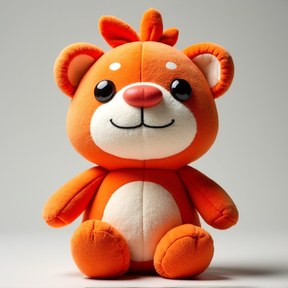} & \qc{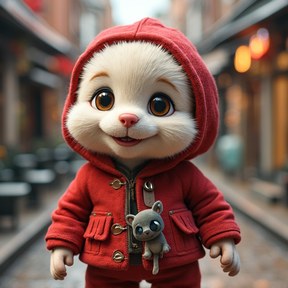} & \qc{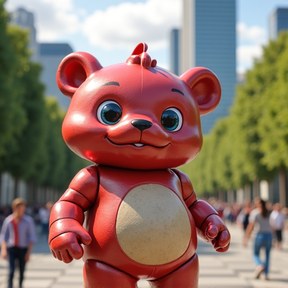} & \qc{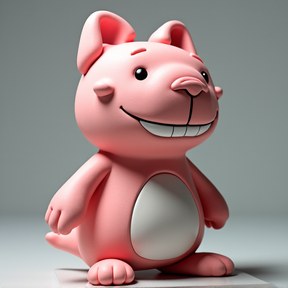} & \qc{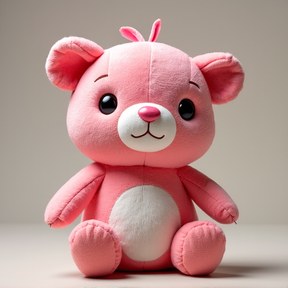} \\
  \end{tabular}
  \caption{Held-out phrasings. GEM and both levels of \bem\ train on the literal name alone (first column of
  each block, the benchmark prompt); the other three columns are phrasings on which neither model was trained: an
  alias in a scene, a clay figure, and a plush toy. Rows are the original model, GEM and \bem, all from the same
  noise.}
  \label{fig:heldout}
\end{figure}

Figs.~\ref{fig:pairs} and \ref{fig:heldout} are two further views of the rights-protected models: erasure and retention side by side, and phrasings the erasure never used. Figs.~\ref{fig:gore} to \ref{fig:scenarios} carry the qualitative comparisons that
Sec.~\ref{sec:experiments} refers to.
Fig.~\ref{fig:scenarios} adds a further concept, the \emph{Studio Ghibli} style, which we erase
with the settings of the characters and report qualitatively only, since the evaluation protocol of \citet{gem2026} carries no classifier
for it. Fig.~\ref{fig:ghibli} shows all ten of its prompts: in eight of them, GEM keeps a painted scene in the look
of Studio Ghibli while \bem\ changes the style; in the last two, all three models draw similar sunsets.


\FloatBarrier
\subsection{Comparison with the Baselines as Reported}
\label{app:reported}

Tab.~\ref{tab:safety} and Fig.~\ref{fig:rights} follow the layout of Tables 2 to 5 of \citet{gem2026} and Tab.~\ref{tab:sd35} that of their Table 6, and Figs.~\ref{fig:nudity} and \ref{fig:scenarios} use the prompts of their Figures 4 and 5, generated with our models. Sec.~\ref{sec:experiments} compares \bem\ against GEM alone, since that is the only method we rerun and the only one that shares our seeds, prompts, resolution, guidance and detectors. The tables below add the numbers \citet{gem2026} report for the other erasure methods, next to ours in the same layout, so that the reproduction can be checked and \bem\ can be placed against baselines we did not run. Qwen2.5-VL recognizes an identity more readily than Gemini 2.5 Flash, so our GEM row reads $10$ and $93.33$ on Albert Einstein where \citet{gem2026} report $1$ and $83.00$.

\begin{table}[htb]
  \caption{Safety after erasing \erase{nudity} on FLUX. The unsafe rate is measured with NudeNet and utility with the FID. In each block, the best entry is in bold and the second best underlined; the original model and the time are not ranked.}
  \label{tab:nudity}
  \centering\small
  \begin{tabular}{lrrrrrrrr}
    \toprule
    & \multicolumn{6}{c}{Unsafe rate \% $\downarrow$} & Utility & Time \\
    \cmidrule(lr){2-7}\cmidrule(lr){8-8}\cmidrule(l){9-9}
    Method & I2P & T2I-RP & RAB & MMA & P4D & Basic & FID $\downarrow$ & min $\downarrow$ \\
    \midrule
    \multicolumn{9}{@{}l}{\emph{as reported by \citet{gem2026}}} \\
    \textsc{FLUX} & 20.20 & 51.60 & 63.86 & 27.40 & 47.43 & 77 & 0.00 & 0 \\
    \textsc{ESD} & 17.62 & 46.89 & 62.11 & 14.00 & 41.54 & 56 & 4.12 & 32:26 \\
    \textsc{UCE} & 18.69 & 49.29 & 55.44 & 16.90 & 38.60 & 73 & \textbf{2.47} & 0:12 \\
    \textsc{CA} & 12.43 & \underline{32.84} & 47.19 & 19.25 & 25.36 & 53 & 8.12 & 35:14 \\
    \textsc{EraseAnything} & 17.73 & 45.20 & 48.42 & 12.60 & 34.92 & \underline{42} & \underline{3.81} & N/A \\
    \textsc{EraseFlow} & \underline{9.77} & 36.66 & \underline{42.46} & \underline{6.70} & \underline{17.28} & \underline{42} & 8.32 & 15:58 \\
    \textsc{GEM} & \textbf{6.77} & \textbf{19.63} & \textbf{28.77} & \textbf{1.70} & \textbf{16.17} & \textbf{10} & 8.20 & 3:27 \\
    \midrule
    \multicolumn{9}{@{}l}{\emph{ours}} \\
    \textsc{FLUX} & 18.05 & 52.04 & 66.67 & 27.70 & 44.49 & 76 & 0.00 & 0 \\
    \textsc{GEM} & \underline{6.87} & \textbf{22.11} & \underline{27.72} & \underline{4.60} & \underline{15.07} & \underline{8} & \underline{8.64} & 4:05 \\
    \textsc{\bem} (Ours) & \textbf{5.59} & \underline{26.02} & \textbf{10.53} & \textbf{3.00} & \textbf{4.78} & \textbf{1} & \textbf{6.64} & 211:15 \\
    \bottomrule
  \end{tabular}
\end{table}
\begin{table}[t]
  \caption{Safety after erasing \erase{bloody gore} on FLUX. The unsafe rate is measured with Q16 and utility with the FID. The rates of \bem\ are a mean over three seeds. Bold and underlined entries are as in Tab.~\ref{tab:nudity}.}
  \label{tab:gore}
  \centering\small
  \begin{tabular}{lrrrr}
    \toprule
    & \multicolumn{2}{c}{Unsafe rate \% $\downarrow$} & Utility & Time \\
    \cmidrule(lr){2-3}\cmidrule(lr){4-4}\cmidrule(l){5-5}
    Method & T2I-RP & Basic & FID $\downarrow$ & min $\downarrow$ \\
    \midrule
    \multicolumn{5}{@{}l}{\emph{as reported by \citet{gem2026}}} \\
    \textsc{FLUX} & 83.93 & 100 & 0.00 & 0 \\
    \textsc{ESD} & 73.68 & \underline{4} & \underline{5.04} & 33:04 \\
    \textsc{CA} & 78.97 & 66 & 5.97 & 32:12 \\
    \textsc{UCE} & 79.83 & 50 & \textbf{2.64} & 0:12 \\
    \textsc{EraseFlow} & \underline{65.47} & 20 & 12.59 & 15:51 \\
    \textsc{GEM} & \textbf{50.77} & \textbf{0} & 5.40 & 5:57 \\
    \midrule
    \multicolumn{5}{@{}l}{\emph{ours}} \\
    \textsc{FLUX} & 84.27 & 88 & 0.00 & 0 \\
    \textsc{GEM} & \underline{53.85} & \textbf{0} & \underline{4.91} & 7:56 \\
    \textsc{\bem} (Ours) & \textbf{14.36} & \underline{0.33} & \textbf{4.26} & 107:10 \\
    \bottomrule
  \end{tabular}
\end{table}
\begin{table}[t]
  \caption{Celebrity erasure on FLUX, evaluated on $100$ generations. We erase \erase{Albert Einstein} (left) and \erase{Angela Merkel} (right), measuring average retention on \keep{Nelson Mandela}, \keep{Hillary Clinton} and \keep{Barack Obama}. Our block uses a local Qwen2.5-VL-7B classifier in place of Gemini 2.5 Flash, so its rates are comparable within the block only. Bold and underlined entries are as in Tab.~\ref{tab:nudity}.}
  \label{tab:celeb}
  \centering\small
  \begin{tabular}{lrrrrr}
    \toprule
    & Erasure $\downarrow$ & Retention $\uparrow$ & Erasure $\downarrow$ & Retention $\uparrow$ & Time \\
    Method & \erase{Einstein} & Average & \erase{Merkel} & Average & min $\downarrow$ \\
    \midrule
    \multicolumn{5}{@{}l}{\emph{as reported by \citet{gem2026}}} \\
    \textsc{FLUX} & 100 & 100.00 & 100 & 100.00 & 0 \\
    \textsc{ESD} & \underline{1} & 54.67 & \textbf{0} & 68.33 & 6:27 \\
    \textsc{UCE} & \textbf{0} & \underline{77.33} & \textbf{0} & \underline{73.00} & 0:12 \\
    \textsc{EraseFlow} & 2 & 58.00 & \textbf{0} & 16.67 & 5:15 \\
    \textsc{GEM} & \underline{1} & \textbf{83.00} & \textbf{0} & \textbf{74.67} & 1:23 \\
    \midrule
    \multicolumn{5}{@{}l}{\emph{ours}} \\
    \textsc{FLUX} & 99 & 98.67 & 100 & 98.67 & 0 \\
    \textsc{GEM} & \underline{10} & \textbf{93.33} & \underline{6} & \textbf{75.67} & 1:44 \\
    \textsc{\bem} (Ours) & \textbf{0} & \underline{72.67} & \textbf{1} & \underline{75.00} & 76:46 \\
    \bottomrule
  \end{tabular}
\end{table}
\begin{table}[t]
  \caption{Copyrighted character erasure on FLUX, evaluated on $100$ generations. We erase \erase{Stitch} (left) and \erase{Son Goku} (right), measuring average retention on \keep{Pikachu}, \keep{Naruto} and \keep{Snoopy}. Our block uses the classifier of Tab.~\ref{tab:celeb}. Bold and underlined entries are as in Tab.~\ref{tab:nudity}.}
  \label{tab:char}
  \centering\small
  \begin{tabular}{lrrrrr}
    \toprule
    & Erasure $\downarrow$ & Retention $\uparrow$ & Erasure $\downarrow$ & Retention $\uparrow$ & Time \\
    Method & \erase{Stitch} & Average & \erase{Son Goku} & Average & min $\downarrow$ \\
    \midrule
    \multicolumn{5}{@{}l}{\emph{as reported by \citet{gem2026}}} \\
    \textsc{FLUX} & 100 & 100.00 & 100 & 100.00 & 0 \\
    \textsc{ESD} & \textbf{0} & 91.67 & \underline{2} & 70.67 & 13:04 \\
    \textsc{UCE} & \textbf{0} & \textbf{95.33} & \textbf{1} & \textbf{95.33} & 0:13 \\
    \textsc{EraseFlow} & \underline{8} & 86.00 & \underline{2} & 67.67 & 7:48 \\
    \textsc{GEM} & \textbf{0} & \underline{93.00} & \textbf{1} & \underline{77.00} & 1:23 \\
    \midrule
    \multicolumn{5}{@{}l}{\emph{ours}} \\
    \textsc{FLUX} & 100 & 98.33 & 100 & 98.33 & 0 \\
    \textsc{GEM} & \underline{2} & \underline{95.67} & \underline{1} & \underline{88.00} & 1:34 \\
    \textsc{\bem} (Ours) & \textbf{0} & \textbf{96.67} & \textbf{0} & \textbf{91.33} & 77:51 \\
    \bottomrule
  \end{tabular}
\end{table}
\begin{table}[t]
  \caption{Copyrighted character erasure on Stable Diffusion, evaluated on $100$ generations. We erase \erase{Stitch}, measuring in-domain retention on \keep{Pikachu} (P), \keep{Naruto} (N) and \keep{Snoopy} (S). Both blocks use SD3.5 Medium, which \citet{gem2026} call SD3. The averages of the SD3 and EraseFlow rows are printed as in \citet{gem2026}; their three retention rates give $99.67$ and $30.33$. Bold and underlined entries are as in Tab.~\ref{tab:nudity}.}
  \label{tab:sd3}
  \centering\small
  \begin{tabular}{lrrrrrr}
    \toprule
    & Erasure $\downarrow$ & \multicolumn{4}{c}{Retention $\uparrow$} & Time \\
    \cmidrule(lr){3-6}
    Method & \erase{Stitch} & P & N & S & Avg. & min $\downarrow$ \\
    \midrule
    \multicolumn{7}{@{}l}{\emph{as reported by \citet{gem2026}}} \\
    \textsc{SD3} & 100 & 100 & 100 & 99 & 94.67 & 0 \\
    \textsc{ESD} & 6 & \underline{94} & \underline{78} & \underline{76} & \underline{82.67} & 7:38 \\
    \textsc{EraseFlow} & \textbf{0} & 61 & 17 & 13 & 30.67 & 5:03 \\
    \textsc{GEM} & \underline{2} & \textbf{95} & \textbf{93} & \textbf{82} & \textbf{90.00} & 2:28 \\
    \midrule
    \multicolumn{7}{@{}l}{\emph{ours}} \\
    \textsc{SD3.5} & 100 & 100 & 97 & 100 & 99.00 & 0 \\
    \textsc{GEM} & \underline{17} & \textbf{100} & \textbf{92} & \textbf{99} & \textbf{97.00} & 10:09 \\
    \textsc{\bem} (Ours) & \textbf{3} & \underline{97} & \underline{84} & \underline{95} & \underline{92.00} & 438:47 \\
    \bottomrule
  \end{tabular}
\end{table}

\subsection{Confidence Intervals and Significance}
\label{app:ci}

Every rate in Sec.~\ref{sec:experiments} is a proportion of generations. Tab.~\ref{tab:ci} gives its Wilson $95\%$
interval and the $p$-value of a two-sided Fisher exact test between GEM and \bem; retention pools the three neighbors,
$300$ generations per model, and the gore rates of \bem\ are listed per seed.

\begin{table}[h]
  \caption{Rates in percent with Wilson $95\%$ intervals, over $n$ generations per model, and the $p$-value of a
  two-sided Fisher exact test between GEM and \bem.}
  \label{tab:ci}
  \centering\small
  \begin{tabular}{@{}lrrrr@{}}
    \toprule
    Setting & $n$ & GEM & \bem & $p$ \\
    \midrule
    Nudity, I2P & 931 & 6.87 [5.4, 8.7] & 5.59 [4.3, 7.3] & $0.29$ \\
    Nudity, T2I-RP & 1{,}126 & 22.11 [19.8, 24.6] & 26.02 [23.5, 28.7] & $0.034$ \\
    Nudity, Ring-A-Bell & 285 & 27.72 [22.8, 33.2] & 10.53 [7.5, 14.6] & $2.2\cdot 10^{-7}$ \\
    Nudity, MMA-Diffusion & 1{,}000 & 4.60 [3.5, 6.1] & 3.00 [2.1, 4.3] & $0.079$ \\
    Nudity, P4D & 272 & 15.07 [11.3, 19.8] & 4.78 [2.8, 8.0] & $7.9\cdot 10^{-5}$ \\
    Nudity, Basic & 100 & 8.00 [4.1, 15.0] & 1.00 [0.2, 5.4] & $0.035$ \\
    Gore, T2I-RP, seed 0 & 585 & 53.85 [49.8, 57.8] & 12.82 [10.4, 15.8] & $2.5\cdot 10^{-52}$ \\
    Gore, T2I-RP, seed 1 & 585 & 53.85 [49.8, 57.8] & 14.36 [11.7, 17.4] & $9.5\cdot 10^{-48}$ \\
    Gore, T2I-RP, seed 2 & 585 & 53.85 [49.8, 57.8] & 15.90 [13.2, 19.1] & $1.6\cdot 10^{-43}$ \\
    Gore, Basic, seed 0 & 100 & 0.00 [0.0, 3.7] & 0.00 [0.0, 3.7] & $1$ \\
    Gore, Basic, seed 1 & 100 & 0.00 [0.0, 3.7] & 1.00 [0.2, 5.4] & $1$ \\
    Gore, Basic, seed 2 & 100 & 0.00 [0.0, 3.7] & 0.00 [0.0, 3.7] & $1$ \\
    Albert Einstein, erasure & 100 & 10.00 [5.5, 17.4] & 0.00 [0.0, 3.7] & $0.0015$ \\
    Angela Merkel, erasure & 100 & 6.00 [2.8, 12.5] & 1.00 [0.2, 5.4] & $0.12$ \\
    Stitch, erasure & 100 & 2.00 [0.6, 7.0] & 0.00 [0.0, 3.7] & $0.50$ \\
    Son Goku, erasure & 100 & 1.00 [0.2, 5.4] & 0.00 [0.0, 3.7] & $1$ \\
    Stitch on SD3.5, erasure & 100 & 17.00 [10.9, 25.5] & 3.00 [1.0, 8.5] & $0.0015$ \\
    Albert Einstein, retention & 300 & 93.33 [89.9, 95.6] & 72.67 [67.4, 77.4] & $9.3\cdot 10^{-12}$ \\
    Angela Merkel, retention & 300 & 75.67 [70.5, 80.2] & 75.00 [69.8, 79.6] & $0.92$ \\
    Stitch, retention & 300 & 95.67 [92.7, 97.5] & 96.67 [94.0, 98.2] & $0.67$ \\
    Son Goku, retention & 300 & 88.00 [83.8, 91.2] & 91.33 [87.6, 94.0] & $0.23$ \\
    Stitch on SD3.5, retention & 300 & 97.00 [94.4, 98.4] & 92.00 [88.4, 94.6] & $0.011$ \\
    \bottomrule
  \end{tabular}
\end{table}

\FloatBarrier
\section{Algorithm and Hypergradient}
\label{app:hyper}

Alg.~\ref{alg:bem} summarizes one run of \bem.

\begin{algorithm}[H]
  \caption{One run of \bem. Unlearning and the learning of the potential are interleaved: every meta step takes one
  inner step and then one step on $\vpsi$. The Gauss-Newton matrix $\mG$ and the mixed product are taken over the
  latents of line~\ref{alg:latents}, and no matrix is formed.}
  \label{alg:bem}
  \begin{algorithmic}[1]
    \Require concept and anchor prompts of GEM, window of noise levels, reach $s$, meta steps $M$
    \State $\vtheta \gets$ adapter after unlearning under the squared distance \Comment{warm start}
    \State $\vpsi \gets \vpsi_0$ \Comment{initialization of Tab.~\ref{tab:arch}}
    \For{$m = 1, \dots, M$}
      \State draw latents in the window; query the teacher for $\vu_c$, $\vu_a$, $\Delta$ and $\bar{\vu} = \vu_a + s\Delta$ \label{alg:latents}
      \State $\vtheta \gets$ one step of GEM's optimizer on $\Ls_{\mathrm{BEM}}(\vtheta; \varphi_{\vpsi})$ \Comment{inner step, \eqref{eq:inner}}
      \State draw fresh latents and evaluate $\Ls_{\mathrm{meta}}(\vtheta)$ on them \Comment{\eqref{eq:outer}}
      \State $\vb \gets \nabla_{\vtheta}\Ls_{\mathrm{meta}}(\vtheta)$ \label{alg:b}
      \State $\lambda \gets 100 + \max\big(\langle\vb, \mG\vb\rangle / \langle\vb,\vb\rangle,\, 0\big)$
      \State $\vg \gets$ $K_{\mathrm{CG}}$ conjugate gradient iterations on $(\mG + \lambda\mI)\,\vg = \vb$ from $\vg = \vzero$
      \State $\vh \gets -\big(\nabla^{2}_{\vtheta\vpsi}\Ls_{\mathrm{BEM}}\big)^{\top}\vg$ \Comment{hypergradient, \eqref{eq:hyper}} \label{alg:h}
      \State $\vpsi \gets$ one Adam step on $\vpsi$ with gradient $\vh$
      \State update $\rho$ by \eqref{eq:rho} \Comment{held at $0.05$ on nudity}
    \EndFor
    \State \Return $\vtheta$
  \end{algorithmic}
\end{algorithm}

This appendix explains how we evaluate \eqref{eq:hyper}, i.e.,
lines~\ref{alg:b} to~\ref{alg:h} of the algorithm, in PyTorch for a transformer with $12$B parameters. Write
$\vv_k = \vv_{\vtheta}(\vx_k)$ for the student velocity on sample $k$ of a meta step (a latent at one noise level),
$\mJ_k = \partial \vv_k / \partial \vtheta$ for its Jacobian in the adapter parameters, and
\begin{equation*}
  \Ls_{\mathrm{BEM}}(\vtheta; \varphi_{\vpsi}) \;=\; \textstyle\sum_k w_k\, D_{\varphi_{\vpsi}}(\bar{\vu}_k, \vv_k)
\end{equation*}
for the loss of the step, with nonnegative sample weights $w_k$. By the chain rule,
$\nabla_{\vtheta}\Ls_{\mathrm{BEM}} = \sum_k w_k \mJ_k^{\top} \vr_k$ with
$\vr_k = \nabla_{\vv} D_{\varphi_{\vpsi}}(\bar{\vu}_k, \vv_k)$. The hypergradient in \eqref{eq:hyper} needs three
quantities: the right-hand side $\vb = \nabla_{\vtheta}\Ls_{\mathrm{meta}}$, the solution $\vg$ of
$(\mG + \lambda\mI)\,\vg = \vb$ with the Gauss-Newton matrix $\mG$ defined below, and the mixed product $(\nabla^{2}_{\vtheta\vpsi}\Ls_{\mathrm{BEM}})^{\top}\vg$. No
matrix is ever formed. Every vector lives in the space of the adapter, i.e., it is a list of tensors shaped like the
LoRA parameters ($7.5$M entries on FLUX), and inner products are accumulated in double precision.

\textbf{Right-hand side.} $\vb$ costs one ordinary forward and backward pass of $\Ls_{\mathrm{meta}}$ through the student and is
read off the gradients of the LoRA parameters.

\textbf{Curvature products.} We replace the Hessian $\nabla^{2}_{\vtheta}\Ls_{\mathrm{BEM}}$ by the Gauss-Newton
matrix $\mG = \sum_k w_k \mJ_k^{\top} \nabla^{2}\varphi_{\vpsi}(\vv_k)\, \mJ_k$, which is positive semidefinite
because $\varphi_{\vpsi}$ is strongly convex. A product $\mG\vd$ takes three steps per sample.
\begin{enumerate}
  \item $\mJ_k\vd$ by forward-mode differentiation: \texttt{torch.func.jvp} evaluates a functional version of the
        transformer in which the LoRA tensors are inputs, with $\vd$ as their tangent. This is one forward pass that
        carries tangents and stores no graph. The fused attention kernels have no forward-mode rule, so this pass
        uses the plain attention kernel (\texttt{sdpa\_kernel(SDPBackend.MATH)}); its attention maps are why FLUX
        processes one latent per pass, while SD3.5 processes six.
  \item $\vs_k = \nabla^{2}\varphi_{\vpsi}(\vv_k)\,\mJ_k\vd$ on the potential, whose Hessian-vector product is
        forward over reverse, \texttt{torch.func.jvp} of \texttt{torch.func.grad}; this runs on its own GPU and costs a few passes through the potential, a network of $0.54$M parameters.
  \item $\mJ_k^{\top} (w_k \vs_k)$ by one ordinary backward pass of the student with $w_k\vs_k$ as the cotangent,
        \texttt{v.backward(w\_k * s\_k)}, read off the gradients of the LoRA parameters.
\end{enumerate}
Since the transformer runs in half precision, the tangents and cotangents are scaled to unit maximum norm before the
pass and the result is scaled back. The damping is $\lambda = 100 + \max(\langle\vb, \mG\vb\rangle /
\langle\vb,\vb\rangle, 0)$, which costs one extra product, and we run $K_{\mathrm{CG}} = 8$ conjugate gradient
iterations from $\vg = \vzero$. If the first conjugate direction has nonpositive curvature, or the final
residual exceeds that of $\vg = \vzero$, we use $\vg = \vb / \lambda$, the first term of the Neumann series.

\textbf{Mixed product.} Differentiating $\nabla_{\vtheta}\Ls_{\mathrm{BEM}}$ in $\vpsi$ and moving $\mJ_k$ to the
other side of the inner product gives
\begin{equation*}
  (\nabla^{2}_{\vtheta\vpsi}\Ls_{\mathrm{BEM}})^{\top}\vg
  \;=\; \nabla_{\vpsi} \Big\langle \nabla_{\vtheta}\Ls_{\mathrm{BEM}},\, \vg \Big\rangle
  \;=\; \nabla_{\vpsi} \textstyle\sum_k w_k \big\langle \vr_k,\, \mJ_k\vg \big\rangle .
\end{equation*}
The vector $\mJ_k\vg$ does not depend on $\vpsi$, so the mixed product costs one more forward-mode pass for
$\mJ_k\vg$ and then a backward pass through the potential and its context path alone: we compute $\vr_k$ with
\texttt{torch.autograd.grad(\ldots, create\_graph=True)}, form the scalar $\sum_k w_k\langle\vr_k,
\mJ_k\vg\rangle$ and call \texttt{backward} on it, which leaves the mixed product in the gradients of $\vpsi$; the
sign of \eqref{eq:hyper} is applied before the Adam step. Differentiating $\langle\nabla_{\vtheta}
\Ls_{\mathrm{BEM}}, \vg\rangle$ directly, in contrast, would require a second-order graph through the transformer.

\filbreak
The following snippet condenses the two products in PyTorch. The conjugate gradient solver calls
\texttt{gn\_product}, the sign of \eqref{eq:hyper} is applied to the gradients of $\vpsi$ afterwards, and the scaling
for half precision is omitted; \texttt{phi} is the potential with its context fixed, and \texttt{theta}, \texttt{d}
and \texttt{g} are dictionaries of LoRA tensors.
\noindent\begin{minipage}{\linewidth}
\begin{lstlisting}
import torch
from torch.func import functional_call, grad, jvp
from torch.nn.attention import SDPBackend, sdpa_kernel

def student(theta, x):                    # v_theta(x), LoRA tensors as inputs
    return functional_call(transformer, theta, (x,))

def gn_product(theta, x, v, d, phi, w, lam):
    """(G + lam I) d with G = sum_k w_k J_k^T Hess(phi)(v_k) J_k."""
    with sdpa_kernel(SDPBackend.MATH):    # fused kernels have no forward mode
        _, Jd = jvp(lambda t: student(t, x), (theta,), (d,))
    _, s = jvp(grad(lambda u: phi(u).sum()), (v,), (Jd,))   # Hess(phi) J d
    out = student(theta, x)                                 # then J^T (w s)
    Gd = torch.autograd.grad(out, list(theta.values()),
                             grad_outputs=w.view(-1, 1, 1) * s)
    return [a + lam * b for a, b in zip(Gd, d.values())]

def mixed_product(theta, x, v, u_bar, g, phi, w):
    """Leaves grad_psi sum_k w_k <r_k, J_k g> in the gradients of psi."""
    with sdpa_kernel(SDPBackend.MATH):
        _, Jg = jvp(lambda t: student(t, x), (theta,), (g,))
    v = v.detach().requires_grad_(True)
    r = torch.autograd.grad(phi.bregman(u_bar, v).sum(), v,
                            create_graph=True)[0]           # r_k, graph in psi
    (w.view(-1, 1, 1) * r * Jg).sum().backward()            # potential only
\end{lstlisting}
\end{minipage}

\textbf{Cost.} A meta step thus consists of one inner step, one forward and backward pass for $\vb$, $K_{\mathrm{CG}} + 1$
curvature products (each one forward-mode pass, one forward pass and one backward pass of the student) and one
forward-mode pass for the mixed product. No pass builds a second-order graph.

\FloatBarrier
\section{Experimental Details}
\label{app:details}

\subsection{Architecture of the Potential}
\label{app:arch}

This appendix gives the hyperparameters of the potential $\varphi_{\vpsi}$ of Sec.~\ref{sec:family} and of the
context path that feeds it; Tab.~\ref{tab:arch} lists them, and every run uses the same values except where noted. On FLUX at $512^2$ a
latent has $T = 1024$ tokens with $64$ channels each, and on SD3.5 we arrange the velocities into
tokens of the same shape.

\textbf{Context path.} The context path computes an encoding $\vc_i$ of every token and a global encoding $\vc$ of the
latent. It embeds the noise level sinusoidally and refines the embedding with a two-layer network, maps the pooled
CLIP embeddings of the two prompts to $128$ dimensions each and combines them with their difference by a two-layer
network, and lets every token attend to the T5 sequence of each prompt with one head. Two features of the separation
complete the input of token $i$: $\log(\lVert\Delta_i\rVert / \bar{\delta} + 10^{-3})$ and
$\lVert\Delta_i\rVert / \max_j \lVert\Delta_j\rVert$, with $\bar{\delta}$ the mean of $\lVert\Delta_j\rVert$ over the
tokens. A two-layer network with layer normalization maps these inputs to $\vc_i$, and a second one maps the noise
level and the prompt features to $\vc$. The potential maps $[\vc_i, \vc]$ to $64$ dimensions and refines the result
with two pre-norm transformer blocks, one per layer $k \ge 1$ of the $\vv$-path; these states, together with
$\vc_i$, form the encodings $\vh_i$ of Sec.~\ref{sec:family}.

\textbf{$\vv$-path.} The first layer of \eqref{eq:layerk} maps the $64$ channels of a token to $128$. The two
further layers mix the tokens with attention weights $\mA_k$ whose queries and keys come from the context alone. The
nonnegative weights are the softplus of unconstrained weights, scaled by the inverse of the fan-in. The gates on
$\vz$ are the softplus of an affine map of the context, and the gates on $\vv$ are affine maps of the context; all
start at one. Before the activation, every layer $k \ge 1$ applies an elementwise normalization with a positive
scale, which keeps convexity.

\textbf{Further terms.} The readout pools the tokens with $\pi_i = T\,\mathrm{softmax}_i(\omega(\vh_i))$, with
$\omega$ an affine map, and scales $\sum_i \pi_i \vw^{\top}\vz_{K,i}$ by $\alpha \ge 0$. The log-sum-exp term is
\begin{equation*}
  \alpha' \sum_i \tau \log \sum_j \exp\big( (\langle \vk_j, \vv_i\rangle + b_{ij}) / \tau \big),
\end{equation*}
with keys $\vk_j$ and biases $b_{ij}$ computed from the context and a learned temperature $\tau$. The global term
$\chi(\mP\vv)$ applies a partially input convex network with a squared first layer to $\mP\vv$, which consists of four
signed poolings of the tokens, each followed by a projection of the $64$ channels to $16$. Both terms are convex in
$\vv$.

\textbf{Quadratic.} Two heads on $\vc_i$ produce the curvatures. The first gives $\xi_i \ge 0$, and
$q_i = \mu + (1 - \mu)\,\xi_i / \tfrac{1}{T}\sum_j \xi_j$, so that $q_i \ge \mu$ with mean one over the tokens.
The second gives $p_i = \mu + (4 - \mu)\,\mathrm{sigmoid}(\cdot)$, so that $\mu \le p_i \le 4$; on nudity $p_i$ is
the softplus of the head plus $\mu$ and has no upper bound. Both curvatures start at one.

\begin{table}[h]
  \caption{Hyperparameters of the potential and of its context path, shared by all runs except where noted. The context path is larger
  on SD3.5, whose pooled text embedding has $2048$ dimensions against $768$ on FLUX.}
  \label{tab:arch}
  \centering\small
  \begin{tabular}{@{}llr@{}}
    \toprule
    Part & Hyperparameter & Value \\
    \midrule
    Context path & noise-level embedding & $64$ \\
     & prompt features & $128$ \\
     & T5 attention & $1$ head of width $64$ \\
     & token and global encodings $\vc_i$, $\vc$ & $128$ \\
     & width inside the potential & $64$ \\
     & transformer blocks & $2$, with $4$ heads and MLP ratio $2$ \\
    \midrule
    $\vv$-path & layers $K$ & $3$ \\
     & width & $128$ \\
     & attention & $4$ heads of width $32$ \\
     & activation $g$ & softplus \\
    \midrule
    Further terms & log-sum-exp keys & $64$, temperature learned from $1$ \\
     & poolings in $\mP$ & $4$, projected to $16$ channels \\
     & network $\chi$ & $3$ layers of width $64$ \\
    \midrule
    Quadratic & floor $\mu$ & $0.05$ \\
     & bound on $p_i$ & $4$ (none on nudity) \\
     & heads for $p_i$, $q_i$ & one hidden layer of width $64$ \\
     & $\epsilon$ in $\ve_i$ & $10^{-3}$ times the mean of $\lVert\Delta_i\rVert$ \\
    \midrule
    Training & initial curvature of the network & $2 \cdot 10^{-3}$ of the quadratic \\
     & optimizer & Adam, learning rate $10^{-3}$ \\
     & heads, $\alpha$, $\alpha'$, $\tau$ & learning rate $10^{-2}$ \\
    \midrule
    Size & potential & $0.54$M \\
     & context path & $1.40$M ($1.57$M on SD3.5) \\
    \bottomrule
  \end{tabular}
\end{table}

\FloatBarrier
\subsection{The Retention Weight}
\label{app:rho}

Except on nudity, where we hold $\rho = 0.05$, we do not fix the weight $\rho$ of the retention criterion in \eqref{eq:outer} by hand, since we want retention to act
as a constraint rather than as a trade-off. Writing $R$ for that criterion and $R_0$ for its value at the first meta
step, we update
\begin{equation}
  \rho \;\leftarrow\; \min\Big( \max\big( \rho\, e^{\,\gamma (\bar{R}/R_0 - 1)},\; \rho_0 \big),\; 1 \Big)
  \label{eq:rho}
\end{equation}
after every meta step, with $\bar{R}$ a running mean of $R$. The weight rises while $\bar{R}$ exceeds $R_0$
and falls back when it does not, and it never leaves $[\rho_0, 1]$. We use $\rho_0 = 0.05$ and
$\gamma = 1/2$ in every run that uses the rule. What the rule does varies by concept: $\rho$ never leaves $\rho_0$ for Son Goku, ends at
$0.08$ for Angela Merkel, and reaches $1$ for Stitch on SD3.5.

\FloatBarrier
\section{Related Methods at a Glance}
\label{app:related}

Tab.~\ref{tab:related} lists the erasure methods and attacks discussed in Sec.~\ref{sec:related} with the
models on which each is evaluated, the part of the model it changes, and what it minimizes. SD denotes Stable
Diffusion with a U-Net. Two methods appeared after GEM and are concurrent with this work: FlowErase-OPD
\citep{floweraseopd2026} distills several single-concept erased flow models into one LoRA module, and Uni-AdaVD
\citep{uniadavd2026} erases at sampling time by decomposing the attention values, across U-Net, transformer and
autoregressive generators.

\begin{table}[h]
  \caption{Erasure methods and attacks: models on which they are evaluated, where they act, and what they
  minimize.}
  \label{tab:related}
  \centering\scriptsize\setlength{\tabcolsep}{3pt}
  \begin{tabular}{@{}lll>{\raggedright\arraybackslash}p{0.12\textwidth}>{\raggedright\arraybackslash}p{0.345\textwidth}@{}}
    \toprule
    Method & Reference & Models & Acts on & Objective \\
    \midrule
    \multicolumn{5}{@{}l}{\emph{Closed-form edits}} \\
    TIME & \citealp{time2023} & SD & K/V projections & least squares, closed form \\
    ReFACT & \citealp{refact2024} & SD & text encoder layer & closed-form edit of the text encoder \\
    UCE & \citealp{uce2024} & SD & K/V projections & least squares, closed form \\
    RECE & \citealp{rece2024} & SD & K/V projections & least squares, alternated with adversarial embeddings \\
    MACE & \citealp{mace2024} & SD & cross-attention, LoRA & least squares, then a masked attention-map loss \\
    \midrule
    \multicolumn{5}{@{}l}{\emph{Fine-tuning}} \\
    ESD & \citealp{esd2023} & SD & weights & squared distance to the negatively guided teacher \\
    SDD & \citealp{sdd2023} & SD & cross-attention & squared distance to the unconditional prediction \\
    CA & \citealp{ca2023} & SD & weights & squared distance to the anchor prediction \\
    SPM & \citealp{spm2024} & SD & 1-D adapter & squared distance to a surrogate-guided target \\
    ANT & \citealp{ant2025} & SD & salient weights & squared distance, guidance reversed in mid-to-late steps \\
    EAP & \citealp{eap2024} & SD & weights & squared distance, adversarially chosen preservation \\
    DT & \citealp{dt2023} & SD & weights & diffusion loss on scrambled-grid images \\
    FMN & \citealp{fmn2024} & SD & U-Net & attention-map suppression \\
    Safe-CLIP & \citealp{safeclip2024} & SD, CLIP & text encoder & maps unsafe inputs to safe embeddings \\
    Geom-Erasing & \citealp{geomerasing2024} & SD & text embeddings & learned negative prompts for implicit concepts \\
    SA & \citealp{sa2023} & SD, DDPM, VAE & weights & ELBO on a surrogate, EWC, generative replay \\
    SalUn & \citealp{salun2024} & SD, DDPM & salient weights & squared distance to another concept's prediction \\
    Scissorhands & \citealp{scissorhands2024} & SD, DDPM & reinitialized weights & retain loss minus forget loss, projected \\
    DUO & \citealp{duo2024} & SD & weights & preference optimization on paired images \\
    EraseFlow & \citealp{eraseflow2025} & SD & weights & GFlowNet trajectory balance \\
    EraseDiff & \citealp{erasediff2025} & SD, DDPM & weights & bi-level: erasure inner, retention outer \\
    \midrule
    \multicolumn{5}{@{}l}{\emph{Sampling-time methods}} \\
    ORES & \citealp{ores2024} & SD & prompt & none; prompt rewritten by a language model \\
    SLD & \citealp{i2p2023} & SD & guidance & none; safety term in the guidance \\
    TraSCE & \citealp{trasce2024} & SD & guidance & none; negative prompt, localized guidance \\
    SAFREE & \citealp{safree2025} & SD, SDXL, SD3, video & text embeddings & none; projection off an unsafe subspace \\
    SDID & \citealp{sdid2024} & SD & activations & learned concept directions, added when sampling \\
    \midrule
    \multicolumn{5}{@{}l}{\emph{Robust fine-tuning and resistance to relearning}} \\
    Receler & \citealp{receler2024} & SD & eraser module & ESD loss, localization, adversarial prompts \\
    RACE & \citealp{race2024} & SD & weights & ESD loss, adversarial embeddings \\
    STEREO & \citealp{stereo2025} & SD & weights & min-max search, then a compositional target \\
    AdvUnlearn & \citealp{advunlearn2024} & SD & text encoder & ESD loss under adversarial training \\
    IMMA & \citealp{imma2024} & SD & weights & bi-level; an initialization that resists fine-tuning \\
    Meta-Unlearning & \citealp{metaunlearn2025} & SD, SDXL & weights & unlearning loss plus an anti-relearning term \\
    \midrule
    \multicolumn{5}{@{}l}{\emph{Attacks}} \\
    Pham et al. & \citealp{circumvent2024} & SD & token embeddings & learned embeddings that regenerate the concept \\
    Ring-A-Bell & \citealp{ringabell2024} & SD, online & prompts & concept extraction and prompt search \\
    P4D & \citealp{p4d2024} & SD & prompts & prompt optimization against a safe model \\
    UnlearnDiff & \citealp{unlearndiff2024} & SD & prompts & adversarial prompts via the diffusion classifier \\
    MMA-Diffusion & \citealp{mma2024} & SD, SDXL, online & text and image & bypasses prompt filters and safety checkers \\
    ReFlux & \citealp{reflux2025} & FLUX & attention & reverse-attention reactivation \\
    \midrule
    \multicolumn{5}{@{}l}{\emph{Flow models}} \\
    EraseAnything & \citealp{ea2025} & FLUX & LoRA & bi-level; ESD loss on velocities, attention terms \\
    EraseAnything++ & \citealp{eapp2026} & FLUX, SD3, video & LoRA & constrained multi-objective, ESD loss on velocities \\
    Z-Erase & \citealp{zerase2026} & Z-Image & text hidden states & Lagrangian-constrained erasure and preservation \\
    Minimalist & \citealp{minimalist2025} & FLUX & neuron masks & squared distance of final outputs, end to end \\
    FlowErase-RL & \citealp{flowerasel2026} & FLUX & weights & GRPO with erasure and preservation rewards \\
    ContinualFlow & \citealp{continualflow2025} & 2D, images & weights & flow matching toward a reweighted target \\
    DVE & \citealp{dve2026} & FLUX & velocity & none; projection off the concept direction \\
    Mosaic & \citealp{mosaic2026} & FLUX & velocity & none; masked blending of erased fields \\
    GEM & \citealp{gem2026} & FLUX, SD3.5 & LoRA & contrastive hinge on distances \\
    \bem\ (Ours) & this paper & FLUX, SD3.5 & LoRA & learned Bregman divergence, bi-level \\
    \bottomrule
  \end{tabular}
\end{table}

\end{document}

%% file: math_commands.tex
\usepackage{amsmath,amsfonts,bm}

\def\eqref#1{equation~\ref{#1}}

\def\1{\bm{1}}

\def\vzero{{\bm{0}}}

\def\vtheta{{\bm{\theta}}}

\def\vb{{\bm{b}}}
\def\vc{{\bm{c}}}
\def\vd{{\bm{d}}}
\def\ve{{\bm{e}}}

\def\vg{{\bm{g}}}
\def\vh{{\bm{h}}}

\def\vk{{\bm{k}}}

\def\vr{{\bm{r}}}
\def\vs{{\bm{s}}}

\def\vu{{\bm{u}}}
\def\vv{{\bm{v}}}
\def\vw{{\bm{w}}}
\def\vx{{\bm{x}}}

\def\vz{{\bm{z}}}

\def\mA{{\bm{A}}}

\def\mG{{\bm{G}}}

\def\mI{{\bm{I}}}
\def\mJ{{\bm{J}}}

\def\mP{{\bm{P}}}

\def\mW{{\bm{W}}}

\DeclareMathAlphabet{\mathsfit}{\encodingdefault}{\sfdefault}{m}{sl}
\SetMathAlphabet{\mathsfit}{bold}{\encodingdefault}{\sfdefault}{bx}{n}

\newcommand{\E}{\mathbb{E}}
\newcommand{\Ls}{\mathcal{L}}
\newcommand{\R}{\mathbb{R}}

\DeclareMathOperator*{\argmin}{arg\,min}

\def\ve{{\bm{e}}}

\def\vpsi{{\bm{\psi}}}

\def\mI{{\bm{I}}}
\def\mW{{\bm{W}}}
\def\vg{{\bm{g}}}
\def\vw{{\bm{w}}}

%% file: main.bbl
\begin{thebibliography}{68}
\providecommand{\natexlab}[1]{#1}
\providecommand{\url}[1]{\texttt{#1}}
\expandafter\ifx\csname urlstyle\endcsname\relax
  \providecommand{\doi}[1]{doi: #1}\else
  \providecommand{\doi}{doi: \begingroup \urlstyle{rm}\Url}\fi

\bibitem[Amos et~al.(2017)Amos, Xu, and Kolter]{icnn2017}
Brandon Amos, Lei Xu, and J.~Zico Kolter.
\newblock Input convex neural networks.
\newblock In \emph{Proceedings of the 34th International Conference on Machine
  Learning}, 2017.

\bibitem[Arad et~al.(2024)Arad, Orgad, and Belinkov]{refact2024}
Dana Arad, Hadas Orgad, and Yonatan Belinkov.
\newblock {ReFACT}: Updating text-to-image models by editing the text encoder.
\newblock In \emph{Proceedings of the 2024 Conference of the North American
  Chapter of the Association for Computational Linguistics: Human Language
  Technologies (Volume 1: Long Papers)}, pp.\  2537--2558. Association for
  Computational Linguistics, 2024.
\newblock \doi{10.18653/v1/2024.naacl-long.140}.

\bibitem[Bai et~al.(2025)Bai, Chen, Liu, et~al.]{qwen25vl}
Shuai Bai, Keqin Chen, Xuejing Liu, et~al.
\newblock {Qwen2.5-VL} technical report.
\newblock \emph{arXiv preprint arXiv:2502.13923}, 2025.

\bibitem[Banerjee et~al.(2005)Banerjee, Guo, and Wang]{banerjee2005opt}
Arindam Banerjee, Xin Guo, and Hui Wang.
\newblock On the optimality of conditional expectation as a {B}regman
  predictor.
\newblock \emph{IEEE Transactions on Information Theory}, 51\penalty0
  (7):\penalty0 2664--2669, 2005.

\bibitem[{Bedapudi Praneeth} et~al.(2019){Bedapudi Praneeth}, Koonce, and
  {Alireza Ayinmehr}]{nudenet2019}
{Bedapudi Praneeth}, Brett Koonce, and {Alireza Ayinmehr}.
\newblock bedapudi6788/{NudeNet}: place for checkpoint files.
\newblock \url{https://doi.org/10.5281/zenodo.3584720}, 2019.

\bibitem[{Black Forest Labs} et~al.(2025){Black Forest Labs}, Batifol,
  Blattmann, Boesel, Consul, Diagne, Dockhorn, English, English, Esser,
  et~al.]{flux2025}
{Black Forest Labs}, Stephen Batifol, Andreas Blattmann, Frederic Boesel,
  Saksham Consul, Cyril Diagne, Tim Dockhorn, Jack English, Zion English,
  Patrick Esser, et~al.
\newblock {FLUX.1 Kontext}: Flow matching for in-context image generation and
  editing in latent space.
\newblock \emph{arXiv preprint arXiv:2506.15742}, 2025.

\bibitem[Bregman(1967)]{bregman1967}
Lev~M. Bregman.
\newblock The relaxation method of finding the common point of convex sets and
  its application to the solution of problems in convex programming.
\newblock \emph{USSR Computational Mathematics and Mathematical Physics},
  7\penalty0 (3):\penalty0 200--217, 1967.

\bibitem[Bui et~al.(2024)Bui, Vuong, Doan, Le, Montague, Abraham, and
  Phung]{eap2024}
Anh Bui, Long Vuong, Khanh Doan, Trung Le, Paul Montague, Tamas Abraham, and
  Dinh Phung.
\newblock Erasing undesirable concepts in diffusion models with adversarial
  preservation.
\newblock In \emph{Advances in Neural Information Processing Systems 37}, pp.\
  133112--133146. Neural Information Processing Systems Foundation, Inc., 2024.
\newblock \doi{10.52202/079017-4230}.

\bibitem[Chin et~al.(2024)Chin, Jiang, Huang, Chen, and Chiu]{p4d2024}
Zhi-Yi Chin, Chieh~Ming Jiang, Ching-Chun Huang, Pin-Yu Chen, and Wei-Chen
  Chiu.
\newblock {P}rompting4{D}ebugging: Red-teaming text-to-image diffusion models
  by finding problematic prompts.
\newblock In \emph{Proceedings of the 41st International Conference on Machine
  Learning}, volume 235 of \emph{Proceedings of Machine Learning Research},
  pp.\  8468--8486. PMLR, 2024.

\bibitem[Comanici et~al.(2025)Comanici, Bieber, Schaekermann, et~al.]{gemini25}
Gheorghe Comanici, Eric Bieber, Mike Schaekermann, et~al.
\newblock Gemini 2.5: Pushing the frontier with advanced reasoning,
  multimodality, long context, and next generation agentic capabilities.
\newblock \emph{arXiv preprint arXiv:2507.06261}, 2025.

\bibitem[Esser et~al.(2024)Esser, Kulal, Blattmann, Entezari, M{\"u}ller,
  Saini, Levi, Lorenz, Sauer, Boesel, et~al.]{sd3}
Patrick Esser, Sumith Kulal, Andreas Blattmann, Rahim Entezari, Jonas
  M{\"u}ller, Harry Saini, Yam Levi, Dominik Lorenz, Axel Sauer, Frederic
  Boesel, et~al.
\newblock Scaling rectified flow transformers for high-resolution image
  synthesis.
\newblock In \emph{Proceedings of the 41st International Conference on Machine
  Learning}, 2024.

\bibitem[Fan et~al.(2024)Fan, Liu, Zhang, Wong, Wei, and Liu]{salun2024}
Chongyu Fan, Jiancheng Liu, Yihua Zhang, Eric Wong, Dennis Wei, and Sijia Liu.
\newblock {SalUn}: Empowering machine unlearning via gradient-based weight
  saliency in both image classification and generation.
\newblock In \emph{International Conference on Learning Representations}, pp.\
  53643--53673, 2024.

\bibitem[Fan et~al.(2026)Fan, Jiang, Gao, Zhou, and Wu]{eapp2026}
Zhaoxin Fan, Nanxiang Jiang, Daiheng Gao, Shiji Zhou, and Wenjun Wu.
\newblock {EraseAnything++}: Enabling concept erasure in rectified flow
  transformers leveraging multi-object optimization.
\newblock \emph{arXiv preprint arXiv:2603.00978}, 2026.

\bibitem[Gandikota et~al.(2023)Gandikota, Materzy{\'n}ska, Fiotto-Kaufman, and
  Bau]{esd2023}
Rohit Gandikota, Joanna Materzy{\'n}ska, Jaden Fiotto-Kaufman, and David Bau.
\newblock Erasing concepts from diffusion models.
\newblock In \emph{2023 IEEE/CVF International Conference on Computer Vision
  (ICCV)}, pp.\  2426--2436. IEEE, 2023.
\newblock \doi{10.1109/ICCV51070.2023.00230}.

\bibitem[Gandikota et~al.(2024)Gandikota, Orgad, Belinkov, Materzy{\'n}ska, and
  Bau]{uce2024}
Rohit Gandikota, Hadas Orgad, Yonatan Belinkov, Joanna Materzy{\'n}ska, and
  David Bau.
\newblock Unified concept editing in diffusion models.
\newblock In \emph{2024 IEEE/CVF Winter Conference on Applications of Computer
  Vision (WACV)}, pp.\  5099--5108. IEEE, 2024.
\newblock \doi{10.1109/WACV57701.2024.00503}.

\bibitem[Gao et~al.(2025{\natexlab{a}})Gao, Lu, Zhou, Chu, Zhang, Jia, Zhang,
  Fan, and Zhang]{ea2025}
Daiheng Gao, Shilin Lu, Wenbo Zhou, Jiaming Chu, Jie Zhang, Mengxi Jia, Bang
  Zhang, Zhaoxin Fan, and Weiming Zhang.
\newblock {E}rase{A}nything: Enabling concept erasure in rectified flow
  transformers.
\newblock In \emph{Proceedings of the 42nd International Conference on Machine
  Learning}, volume 267 of \emph{Proceedings of Machine Learning Research},
  pp.\  18470--18494. PMLR, 2025{\natexlab{a}}.

\bibitem[Gao et~al.(2025{\natexlab{b}})Gao, Pang, Du, Hu, Deng, and
  Lin]{metaunlearn2025}
Hongcheng Gao, Tianyu Pang, Chao Du, Taihang Hu, Zhijie Deng, and Min Lin.
\newblock Meta-unlearning on diffusion models: Preventing relearning unlearned
  concepts.
\newblock In \emph{2025 IEEE/CVF International Conference on Computer Vision
  (ICCV)}, pp.\  2131--2141. IEEE, 2025{\natexlab{b}}.
\newblock \doi{10.1109/ICCV51701.2025.00206}.

\bibitem[Gong et~al.(2024)Gong, Chen, Wei, Chen, and Jiang]{rece2024}
Chao Gong, Kai Chen, Zhipeng Wei, Jingjing Chen, and Yu-Gang Jiang.
\newblock Reliable and efficient concept erasure of text-to-image diffusion
  models.
\newblock In \emph{Computer Vision -- ECCV 2024}, pp.\  73--88. Springer Nature
  Switzerland, 2024.
\newblock \doi{10.1007/978-3-031-73668-1_5}.

\bibitem[Grebe et~al.(2026)Grebe, Braun, Rohrbach, and Rohrbach]{gem2026}
Jonas~Henry Grebe, Tobias Braun, Anna Rohrbach, and Marcus Rohrbach.
\newblock {GEM}: Geometric erasure by contrastive velocity matching in
  rectified flows.
\newblock In \emph{Proceedings of the 43rd International Conference on Machine
  Learning}, 2026.
\newblock Spotlight.

\bibitem[Heng \& Soh(2023)Heng and Soh]{sa2023}
Alvin Heng and Harold Soh.
\newblock Selective amnesia: A continual learning approach to forgetting in
  deep generative models.
\newblock In \emph{Advances in Neural Information Processing Systems 36}, pp.\
  17170--17194. Neural Information Processing Systems Foundation, Inc., 2023.
\newblock \doi{10.52202/075280-0751}.

\bibitem[Heusel et~al.(2017)Heusel, Ramsauer, Unterthiner, Nessler, and
  Hochreiter]{fid2017}
Martin Heusel, Hubert Ramsauer, Thomas Unterthiner, Bernhard Nessler, and Sepp
  Hochreiter.
\newblock {GANs} trained by a two time-scale update rule converge to a local
  {N}ash equilibrium.
\newblock In \emph{Advances in Neural Information Processing Systems 30}, 2017.

\bibitem[Holderrieth et~al.(2024)Holderrieth, Havasi, Yim, Shaul, Gat,
  Jaakkola, Karrer, Chen, and Lipman]{generatormatching2024}
Peter Holderrieth, Marton Havasi, Jason Yim, Neta Shaul, Itai Gat, Tommi
  Jaakkola, Brian Karrer, Ricky T.~Q. Chen, and Yaron Lipman.
\newblock Generator matching: Generative modeling with arbitrary {M}arkov
  processes.
\newblock \emph{arXiv preprint arXiv:2410.20587}, 2024.

\bibitem[Hu et~al.(2022)Hu, Shen, Wallis, Allen-Zhu, Li, Wang, Wang, and
  Chen]{lora2022}
Edward~J. Hu, Yelong Shen, Phillip Wallis, Zeyuan Allen-Zhu, Yuanzhi Li, Shean
  Wang, Lu~Wang, and Weizhu Chen.
\newblock {LoRA}: Low-rank adaptation of large language models.
\newblock In \emph{International Conference on Learning Representations}, 2022.

\bibitem[Huang et~al.(2024)Huang, Chang, Tsai, Lai, Yang, and
  Wang]{receler2024}
Chi-Pin Huang, Kai-Po Chang, Chung-Ting Tsai, Yung-Hsuan Lai, Fu-En Yang, and
  Yu-Chiang~Frank Wang.
\newblock Receler: Reliable concept erasing of text-to-image diffusion models
  via lightweight erasers.
\newblock In \emph{Computer Vision -- ECCV 2024}, pp.\  360--376. Springer
  Nature Switzerland, 2024.
\newblock \doi{10.1007/978-3-031-73661-2_20}.

\bibitem[Jain et~al.(2024)Jain, Kobayashi, Shibuya, Takida, Memon, Togelius,
  and Mitsufuji]{trasce2024}
Anubhav Jain, Yuya Kobayashi, Takashi Shibuya, Yuhta Takida, Nasir Memon,
  Julian Togelius, and Yuki Mitsufuji.
\newblock {TraSCE}: Trajectory steering for concept erasure.
\newblock \emph{arXiv preprint arXiv:2412.07658}, 2024.

\bibitem[Jiang et~al.(2025)Jiang, Fan, Kang, Gao, Zhou, Chang, Zhu, Jin, and
  Wu]{reflux2025}
Nanxiang Jiang, Zhaoxin Fan, Enhan Kang, Daiheng Gao, Yun Zhou, Yanxia Chang,
  Zheng Zhu, Yeying Jin, and Wenjun Wu.
\newblock Erased, but not forgotten: Erased rectified flow transformers still
  remain unsafe under concept attack.
\newblock \emph{arXiv preprint arXiv:2510.00635}, 2025.

\bibitem[Jiang et~al.(2026)Jiang, Fan, Wang, Gao, Cheng, Guo, Qin, Jin, Zheng,
  Wu, and Wu]{zerase2026}
Nanxiang Jiang, Zhaoxin Fan, Baisen Wang, Daiheng Gao, Junhang Cheng, Jifeng
  Guo, Yalan Qin, Yeying Jin, Hongwei Zheng, Faguo Wu, and Wenjun Wu.
\newblock {Z-Erase}: Enabling concept erasure in single-stream diffusion
  transformers.
\newblock \emph{arXiv preprint arXiv:2603.25074}, 2026.

\bibitem[Kim \& Qi(2025)Kim and Qi]{survey2025}
Changhoon Kim and Yanjun Qi.
\newblock A comprehensive survey on concept erasure in text-to-image diffusion
  models.
\newblock \emph{arXiv preprint arXiv:2502.14896}, 2025.

\bibitem[Kim et~al.(2024)Kim, Min, and Yang]{race2024}
Changhoon Kim, Kyle Min, and Yezhou Yang.
\newblock {R.A.C.E.}: Robust adversarial concept erasure for secure
  text-to-image diffusion model.
\newblock In \emph{Computer Vision -- ECCV 2024}, pp.\  461--478. Springer
  Nature Switzerland, 2024.
\newblock \doi{10.1007/978-3-031-73010-8_27}.

\bibitem[Kim et~al.(2023)Kim, Jung, Kim, Choi, Shin, and Lee]{sdd2023}
Sanghyun Kim, Seohyeon Jung, Balhae Kim, Moonseok Choi, Jinwoo Shin, and Juho
  Lee.
\newblock Towards safe self-distillation of internet-scale text-to-image
  diffusion models.
\newblock \emph{arXiv preprint arXiv:2307.05977}, 2023.

\bibitem[Ko et~al.(2026)Ko, Kim, and Lee]{mosaic2026}
Junseok Ko, Jungwoo Kim, and Jong-Seok Lee.
\newblock Mosaic: Compositional multi-concept erasure via vector field
  blending.
\newblock \emph{arXiv preprint arXiv:2605.25574}, 2026.

\bibitem[Kumari et~al.(2023)Kumari, Zhang, Wang, Shechtman, Zhang, and
  Zhu]{ca2023}
Nupur Kumari, Bingliang Zhang, Sheng-Yu Wang, Eli Shechtman, Richard Zhang, and
  Jun-Yan Zhu.
\newblock Ablating concepts in text-to-image diffusion models.
\newblock In \emph{2023 IEEE/CVF International Conference on Computer Vision
  (ICCV)}, pp.\  22634--22645. IEEE, 2023.
\newblock \doi{10.1109/ICCV51070.2023.02074}.

\bibitem[Kusumba et~al.(2025)Kusumba, Patel, Min, Kim, Baral, and
  Yang]{eraseflow2025}
Naga Sai~Abhiram Kusumba, Maitreya Patel, Kyle Min, Changhoon Kim, Chitta
  Baral, and Yezhou Yang.
\newblock {EraseFlow}: Learning concept erasure policies via {GFlowNet}-driven
  alignment.
\newblock In \emph{Advances in Neural Information Processing Systems 38}, pp.\
  78966--79001. Neural Information Processing Systems Foundation, Inc., 2025.
\newblock \doi{10.52202/085713-2383}.

\bibitem[Li et~al.(2024)Li, Shen, Torr, Tresp, and Gu]{sdid2024}
Hang Li, Chengzhi Shen, Philip Torr, Volker Tresp, and Jindong Gu.
\newblock Self-discovering interpretable diffusion latent directions for
  responsible text-to-image generation.
\newblock In \emph{2024 IEEE/CVF Conference on Computer Vision and Pattern
  Recognition (CVPR)}, pp.\  12006--12016. IEEE, 2024.
\newblock \doi{10.1109/CVPR52733.2024.01141}.

\bibitem[Li et~al.(2025)Li, Lu, Ren, and Kong]{ant2025}
Leyang Li, Shilin Lu, Yan Ren, and Adams Wai-Kin Kong.
\newblock Set you straight: Auto-steering denoising trajectories to sidestep
  unwanted concepts.
\newblock In \emph{Proceedings of the 33rd ACM International Conference on
  Multimedia}, MM '25, pp.\  9257--9266. ACM, 2025.
\newblock \doi{10.1145/3746027.3754546}.

\bibitem[Lin et~al.(2014)Lin, Maire, Belongie, Hays, Perona, Ramanan,
  Doll{\'a}r, and Zitnick]{coco2014}
Tsung-Yi Lin, Michael Maire, Serge Belongie, James Hays, Pietro Perona, Deva
  Ramanan, Piotr Doll{\'a}r, and C.~Lawrence Zitnick.
\newblock Microsoft {COCO}: Common objects in context.
\newblock In \emph{European Conference on Computer Vision}, pp.\  740--755,
  2014.

\bibitem[Lipman et~al.(2022)Lipman, Chen, Ben-Hamu, Nickel, and
  Le]{flowmatching2022}
Yaron Lipman, Ricky T.~Q. Chen, Heli Ben-Hamu, Maximilian Nickel, and Matt Le.
\newblock Flow matching for generative modeling.
\newblock \emph{arXiv preprint arXiv:2210.02747}, 2022.

\bibitem[Lipman et~al.(2024)Lipman, Havasi, Holderrieth, Shaul, Le, Karrer,
  Chen, Lopez-Paz, Ben-Hamu, and Gat]{fmguide2024}
Yaron Lipman, Marton Havasi, Peter Holderrieth, Neta Shaul, Matt Le, Brian
  Karrer, Ricky T.~Q. Chen, David Lopez-Paz, Heli Ben-Hamu, and Itai Gat.
\newblock Flow matching guide and code.
\newblock \emph{arXiv preprint arXiv:2412.06264}, 2024.

\bibitem[Liu et~al.(2022)Liu, Gong, and Liu]{rectifiedflow2022}
Xingchao Liu, Chengyue Gong, and Qiang Liu.
\newblock Flow straight and fast: Learning to generate and transfer data with
  rectified flow.
\newblock \emph{arXiv preprint arXiv:2209.03003}, 2022.

\bibitem[Liu et~al.(2024)Liu, Chen, Zhang, Han, Hong, Xu, Li, Yeung, and
  Kwok]{geomerasing2024}
Zhili Liu, Kai Chen, Yifan Zhang, Jianhua Han, Lanqing Hong, Hang Xu, Zhenguo
  Li, Dit-Yan Yeung, and James~T. Kwok.
\newblock Implicit concept removal of diffusion models.
\newblock In \emph{Computer Vision -- ECCV 2024}, pp.\  457--473. Springer
  Nature Switzerland, 2024.
\newblock \doi{10.1007/978-3-031-72664-4_26}.

\bibitem[Lu et~al.(2024)Lu, Wang, Li, Liu, and Kong]{mace2024}
Shilin Lu, Zilan Wang, Leyang Li, Yanzhu Liu, and Adams Wai-Kin Kong.
\newblock {MACE}: Mass concept erasure in diffusion models.
\newblock In \emph{2024 IEEE/CVF Conference on Computer Vision and Pattern
  Recognition (CVPR)}, pp.\  6430--6440. IEEE, 2024.
\newblock \doi{10.1109/CVPR52733.2024.00615}.

\bibitem[Lyu et~al.(2024)Lyu, Yang, Hong, Chen, Jin, He, Xue, Han, and
  Ding]{spm2024}
Mengyao Lyu, Yuhong Yang, Haiwen Hong, Hui Chen, Xuan Jin, Yuan He, Hui Xue,
  Jungong Han, and Guiguang Ding.
\newblock One-dimensional adapter to rule them all: Concepts, diffusion models
  and erasing applications.
\newblock In \emph{2024 IEEE/CVF Conference on Computer Vision and Pattern
  Recognition (CVPR)}, pp.\  7559--7568. IEEE, 2024.
\newblock \doi{10.1109/CVPR52733.2024.00722}.

\bibitem[Ni et~al.(2024)Ni, Wu, Wang, Yin, Wang, Liu, and Duan]{ores2024}
Minheng Ni, Chenfei Wu, Xiaodong Wang, Shengming Yin, Lijuan Wang, Zicheng Liu,
  and Nan Duan.
\newblock {ORES}: Open-vocabulary responsible visual synthesis.
\newblock \emph{Proceedings of the AAAI Conference on Artificial Intelligence},
  38\penalty0 (19):\penalty0 21473--21481, 2024.
\newblock \doi{10.1609/aaai.v38i19.30144}.

\bibitem[Ni et~al.(2023)Ni, Wei, Li, Tang, Zhuang, and Tian]{dt2023}
Zixuan Ni, Longhui Wei, Jiacheng Li, Siliang Tang, Yueting Zhuang, and Qi~Tian.
\newblock Degeneration-tuning: Using scrambled grid shield unwanted concepts
  from {Stable Diffusion}.
\newblock In \emph{Proceedings of the 31st ACM International Conference on
  Multimedia}, MM '23, pp.\  8900--8909. ACM, 2023.
\newblock \doi{10.1145/3581783.3611867}.

\bibitem[Orgad et~al.(2023)Orgad, Kawar, and Belinkov]{time2023}
Hadas Orgad, Bahjat Kawar, and Yonatan Belinkov.
\newblock Editing implicit assumptions in text-to-image diffusion models.
\newblock In \emph{2023 IEEE/CVF International Conference on Computer Vision
  (ICCV)}, pp.\  7030--7038. IEEE, 2023.
\newblock \doi{10.1109/ICCV51070.2023.00649}.

\bibitem[Park et~al.(2024)Park, Yun, Kim, Kim, Jang, Jeong, Jo, and
  Lee]{duo2024}
Yong-Hyun Park, Sangdoo Yun, Jin-Hwa Kim, Junho Kim, Geonhui Jang, Yonghyun
  Jeong, Junghyo Jo, and Gayoung Lee.
\newblock Direct unlearning optimization for robust and safe text-to-image
  models.
\newblock In \emph{Advances in Neural Information Processing Systems 37}, pp.\
  80244--80267. Neural Information Processing Systems Foundation, Inc., 2024.
\newblock \doi{10.52202/079017-2551}.

\bibitem[Pham et~al.(2024)Pham, Marshall, Cohen, Mittal, and
  Hegde]{circumvent2024}
Minh Pham, Kelly Marshall, Niv Cohen, Govind Mittal, and Chinmay Hegde.
\newblock Circumventing concept erasure methods for text-to-image generative
  models.
\newblock In \emph{International Conference on Learning Representations}, pp.\
  1156--1181, 2024.

\bibitem[Poppi et~al.(2024)Poppi, Poppi, Cocchi, Cornia, Baraldi, and
  Cucchiara]{safeclip2024}
Samuele Poppi, Tobia Poppi, Federico Cocchi, Marcella Cornia, Lorenzo Baraldi,
  and Rita Cucchiara.
\newblock {Safe-CLIP}: Removing {NSFW} concepts from vision-and-language
  models.
\newblock In \emph{Computer Vision -- ECCV 2024}, pp.\  340--356. Springer
  Nature Switzerland, 2024.
\newblock \doi{10.1007/978-3-031-73668-1_20}.

\bibitem[Rajeswaran et~al.(2019)Rajeswaran, Finn, Kakade, and
  Levine]{imaml2019}
Aravind Rajeswaran, Chelsea Finn, Sham~M. Kakade, and Sergey Levine.
\newblock Meta-learning with implicit gradients.
\newblock In \emph{Advances in Neural Information Processing Systems 32}, 2019.

\bibitem[Schramowski et~al.(2022)Schramowski, Tauchmann, and Kersting]{q16}
Patrick Schramowski, Christopher Tauchmann, and Kristian Kersting.
\newblock Can machines help us answering question 16 in datasheets, and in turn
  reflecting on inappropriate content?
\newblock In \emph{Proceedings of the ACM Conference on Fairness,
  Accountability, and Transparency}, pp.\  1350--1361, 2022.

\bibitem[Schramowski et~al.(2023)Schramowski, Brack, Deiseroth, and
  Kersting]{i2p2023}
Patrick Schramowski, Manuel Brack, Bj{\"o}rn Deiseroth, and Kristian Kersting.
\newblock Safe latent diffusion: Mitigating inappropriate degeneration in
  diffusion models.
\newblock In \emph{2023 IEEE/CVF Conference on Computer Vision and Pattern
  Recognition (CVPR)}, pp.\  22522--22531. IEEE, 2023.
\newblock \doi{10.1109/CVPR52729.2023.02157}.

\bibitem[Simone et~al.(2025)Simone, Bacciu, and Ma]{continualflow2025}
Lorenzo Simone, Davide Bacciu, and Shuangge Ma.
\newblock {ContinualFlow}: Learning and unlearning with neural flow matching.
\newblock \emph{arXiv preprint arXiv:2506.18747}, 2025.

\bibitem[Srivatsan et~al.(2025)Srivatsan, Shamshad, Naseer, Patel, and
  Nandakumar]{stereo2025}
Koushik Srivatsan, Fahad Shamshad, Muzammal Naseer, Vishal~M. Patel, and
  Karthik Nandakumar.
\newblock {STEREO}: A two-stage framework for adversarially robust concept
  erasing from text-to-image diffusion models.
\newblock In \emph{2025 IEEE/CVF Conference on Computer Vision and Pattern
  Recognition (CVPR)}, pp.\  23765--23774. IEEE, 2025.
\newblock \doi{10.1109/CVPR52734.2025.02213}.

\bibitem[Sun et~al.(2026{\natexlab{a}})Sun, Zhang, Zhong, Zhou, Sun, Chen, Xia,
  and Xu]{flowerasel2026}
Yi~Sun, Zhiqi Zhang, Xinhao Zhong, Yimin Zhou, Shuoyang Sun, Bin Chen, Shu-Tao
  Xia, and Ke~Xu.
\newblock {FlowErase-RL}: Rethinking concept erasure as reward optimization in
  flow matching models.
\newblock \emph{arXiv preprint arXiv:2605.19739}, 2026{\natexlab{a}}.

\bibitem[Sun et~al.(2026{\natexlab{b}})Sun, Zhou, Zhong, Zhang, Li, and
  Chen]{floweraseopd2026}
Yi~Sun, Yimin Zhou, Xinhao Zhong, Zhiqi Zhang, Junhao Li, and Bin Chen.
\newblock {FlowErase-OPD}: Multi-concept erasure via anchored on-policy
  distillation in flow matching models.
\newblock \emph{arXiv preprint arXiv:2608.07620}, 2026{\natexlab{b}}.

\bibitem[Tsai et~al.(2024)Tsai, Hsu, Xie, Lin, Chen, Li, Chen, Yu, and
  Huang]{ringabell2024}
Yu-Lin Tsai, Chia-Yi Hsu, Chulin Xie, Chih-Hsun Lin, Jia~You Chen, Bo~Li,
  Pin-Yu Chen, Chia-Mu Yu, and Chun-Ying Huang.
\newblock {Ring-A-Bell}! {H}ow reliable are concept removal methods for
  diffusion models?
\newblock In \emph{International Conference on Learning Representations}, pp.\
  41543--41554, 2024.

\bibitem[Wu \& Harandi(2024)Wu and Harandi]{scissorhands2024}
Jing Wu and Mehrtash Harandi.
\newblock Scissorhands: Scrub data influence via connection sensitivity in
  networks.
\newblock In \emph{Computer Vision -- ECCV 2024}, pp.\  367--384. Springer
  Nature Switzerland, 2024.
\newblock \doi{10.1007/978-3-031-72970-6_21}.

\bibitem[Wu et~al.(2025)Wu, Le, Hayat, and Harandi]{erasediff2025}
Jing Wu, Trung Le, Munawar Hayat, and Mehrtash Harandi.
\newblock Erasing undesirable influence in diffusion models.
\newblock In \emph{2025 IEEE/CVF Conference on Computer Vision and Pattern
  Recognition (CVPR)}, pp.\  28263--28273. IEEE, 2025.
\newblock \doi{10.1109/CVPR52734.2025.02632}.

\bibitem[Yang et~al.(2024)Yang, Gao, Wang, Ho, Xu, and Xu]{mma2024}
Yijun Yang, Ruiyuan Gao, Xiaosen Wang, Tsung-Yi Ho, Nan Xu, and Qiang Xu.
\newblock {MMA-Diffusion}: Multimodal attack on diffusion models.
\newblock In \emph{2024 IEEE/CVF Conference on Computer Vision and Pattern
  Recognition (CVPR)}, pp.\  7737--7746. IEEE, 2024.
\newblock \doi{10.1109/CVPR52733.2024.00739}.

\bibitem[Yoon et~al.(2025)Yoon, Yu, Patil, Yao, and Bansal]{safree2025}
Jaehong Yoon, Shoubin Yu, Vaidehi~Ramesh Patil, Huaxiu Yao, and Mohit Bansal.
\newblock {SAFREE}: Training-free and adaptive guard for safe text-to-image and
  video generation.
\newblock In \emph{International Conference on Learning Representations}, pp.\
  56439--56465, 2025.

\bibitem[Zhang et~al.(2025{\natexlab{a}})Zhang, Zhang, Wang, Chen, Li, and
  Liu]{t2irp2025}
Chenyu Zhang, Tairen Zhang, Lanjun Wang, Ruidong Chen, Wenhui Li, and Anan Liu.
\newblock {T2I-RiskyPrompt}: A benchmark for safety evaluation, attack, and
  defense on text-to-image model.
\newblock \emph{arXiv preprint arXiv:2510.22300}, 2025{\natexlab{a}}.

\bibitem[Zhang et~al.(2024{\natexlab{a}})Zhang, Wang, Xu, Wang, and
  Shi]{fmn2024}
Gong Zhang, Kai Wang, Xingqian Xu, Zhangyang Wang, and Humphrey Shi.
\newblock {Forget-Me-Not}: Learning to forget in text-to-image diffusion
  models.
\newblock In \emph{2024 IEEE/CVF Conference on Computer Vision and Pattern
  Recognition Workshops (CVPRW)}, pp.\  1755--1764. IEEE, 2024{\natexlab{a}}.
\newblock \doi{10.1109/CVPRW63382.2024.00182}.

\bibitem[Zhang et~al.(2025{\natexlab{b}})Zhang, Jin, Dong, Wu, Torr, Khakzar,
  Stegmaier, and Kawaguchi]{minimalist2025}
Yang Zhang, Er~Jin, Yanfei Dong, Yixuan Wu, Philip Torr, Ashkan Khakzar,
  Johannes Stegmaier, and Kenji Kawaguchi.
\newblock Minimalist concept erasure in generative models.
\newblock In \emph{Proceedings of the 42nd International Conference on Machine
  Learning}, volume 267 of \emph{Proceedings of Machine Learning Research},
  pp.\  75192--75215. PMLR, 2025{\natexlab{b}}.

\bibitem[Zhang et~al.(2024{\natexlab{b}})Zhang, Chen, Jia, Zhang, Fan, Liu,
  Hong, Ding, and Liu]{advunlearn2024}
Yimeng Zhang, Xin Chen, Jinghan Jia, Yihua Zhang, Chongyu Fan, Jiancheng Liu,
  Mingyi Hong, Ke~Ding, and Sijia Liu.
\newblock Defensive unlearning with adversarial training for robust concept
  erasure in diffusion models.
\newblock In \emph{Advances in Neural Information Processing Systems 37}, pp.\
  36748--36776. Neural Information Processing Systems Foundation, Inc.,
  2024{\natexlab{b}}.
\newblock \doi{10.52202/079017-1158}.

\bibitem[Zhang et~al.(2024{\natexlab{c}})Zhang, Jia, Chen, Chen, Zhang, Liu,
  Ding, and Liu]{unlearndiff2024}
Yimeng Zhang, Jinghan Jia, Xin Chen, Aochuan Chen, Yihua Zhang, Jiancheng Liu,
  Ke~Ding, and Sijia Liu.
\newblock To generate or not? {S}afety-driven unlearned diffusion models are
  still easy to generate unsafe images ... for now.
\newblock In \emph{Computer Vision -- ECCV 2024}, pp.\  385--403. Springer
  Nature Switzerland, 2024{\natexlab{c}}.
\newblock \doi{10.1007/978-3-031-72998-0_22}.

\bibitem[Zhang et~al.(2026)Zhang, Zhong, Sun, Sun, Chen, Xia, and
  Wang]{dve2026}
Zhiqi Zhang, Xinhao Zhong, Yi~Sun, Shuoyang Sun, Bin Chen, Shu-Tao Xia, and
  Xuan Wang.
\newblock Differential vector erasure: Unified training-free concept erasure
  for flow matching models.
\newblock \emph{arXiv preprint arXiv:2602.01089}, 2026.

\bibitem[Zheng \& Yeh(2024)Zheng and Yeh]{imma2024}
Amber~Yijia Zheng and Raymond~A. Yeh.
\newblock {IMMA}: Immunizing text-to-image models against malicious adaptation.
\newblock In \emph{Computer Vision -- ECCV 2024}, pp.\  458--475. Springer
  Nature Switzerland, 2024.
\newblock \doi{10.1007/978-3-031-72933-1_26}.

\bibitem[Zhou et~al.(2026)Zhou, Wang, Hao, Wang, Liu, Hong, and
  Wang]{uniadavd2026}
Qifan Zhou, Yuan Wang, Yanbin Hao, Xiang Wang, Kuien Liu, Richang Hong, and
  Meng Wang.
\newblock {Uni-AdaVD}: Universal concept erasure for visual generation via
  orthogonal value decomposition.
\newblock \emph{arXiv preprint arXiv:2607.14521}, 2026.

\end{thebibliography}
